\documentclass{article}
\PassOptionsToPackage{table}{xcolor}
\let\origaddcontentsline\addcontentsline
\usepackage{iclr2027_conference,times}
\let\addcontentsline\origaddcontentsline
\usepackage{hyperref}
\usepackage{amsmath}
\usepackage{enumitem}
\usepackage{xcolor}
\usepackage{caption}
\iclrfinalcopy

\usepackage{array}
\usepackage{booktabs}
\usepackage{tabularx}
\usepackage{multirow}
\usepackage{xspace}
\usepackage[most]{tcolorbox}
\usepackage{placeins}
\usepackage{fvextra}
\fvset{fontsize=\footnotesize, frame=single, framesep=5pt, rulecolor=\color{black!35}, breaklines=true, breakanywhere=true, breaksymbolleft={}, breakindent=1.5em, baselinestretch=1.08}
\hypersetup{colorlinks=true,linkcolor=blue!45!black,citecolor=blue!45!black,urlcolor=blue!45!black}
\graphicspath{{}{appendix/}}
\definecolor{astrablue}{HTML}{1764A0}
\definecolor{astralight}{HTML}{EAF2F9}
\colorlet{tabband}{black!8}
\colorlet{tabzebra}{black!4}
\newcommand{\best}[1]{\textbf{#1}}
\newcommand{\second}[1]{\underline{#1}}
\newcommand{\hd}[1]{\textbf{\boldmath #1}}
\newcommand{\sysname}[1]{\textbf{\boldmath #1}}
\AtBeginEnvironment{table}{\renewcommand{\arraystretch}{1.15}}
\newcommand{\astra}{GPT-6-Astra\xspace}
\newtcolorbox{promptcase}[1]{enhanced,breakable,
  colback=astralight!45!white,colframe=astrablue!65!black,
  colbacktitle=astrablue!85!black,coltitle=white,
  fonttitle=\bfseries\small,title={#1},boxrule=0.45pt,arc=1.5pt,
  left=6pt,right=6pt,top=5pt,bottom=5pt,
  before skip=8pt,after skip=8pt,fontupper=\small}

\title{Profiling a Frontier Agent with EBench:\\ From Direct Control to Harness with Policy}

\author{\hspace*{-\tabcolsep}\parbox[t]{\textwidth}{\centering
{\normalfont Yifei Yao$^{1}$\thanks{Equal contribution. \quad $^{\dagger}$Corresponding author: Xing Gao (\texttt{gaoxing@pjlab.org.cn}).} \quad Yude Zou$^{1,2}$\footnotemark[1] \quad Guanbo Jiao$^{1,3}$\footnotemark[1] \quad Xing Gao$^{1,\dagger}$ \quad Ning Gao$^{1,2}$}\\
{\normalfont Yukai Wang$^{1}$ \quad Jiantong Chen$^{1}$ \quad Zihou Zhu$^{1}$ \quad Xinyu Li$^{1}$ \quad Ningyu Yan$^{1}$ \quad Shuai Wang$^{1}$}\\
{\normalfont Jianming Ma$^{1,2}$ \quad Haoyu Guo$^{1}$  \quad Weinan Zhang$^{1,2}$ \quad Chunhua Shen$^{1,4}$}\\[4pt]
{\normalfont $^1$Shanghai Artificial Intelligence Laboratory \quad $^2$Shanghai Jiao Tong University}\\
{\normalfont $^3$Fudan University \quad $^4$Zhejiang University}}
}

\begin{document}
\addtocontents{toc}{\protect\setcounter{tocdepth}{-1}}

\fancypagestyle{firstpage}{%
  \fancyhf{}%
  \setlength{\headheight}{16pt}%
  \renewcommand{\headrulewidth}{0.4pt}%
  \lhead{\raisebox{-2pt}{\includegraphics[height=13pt]{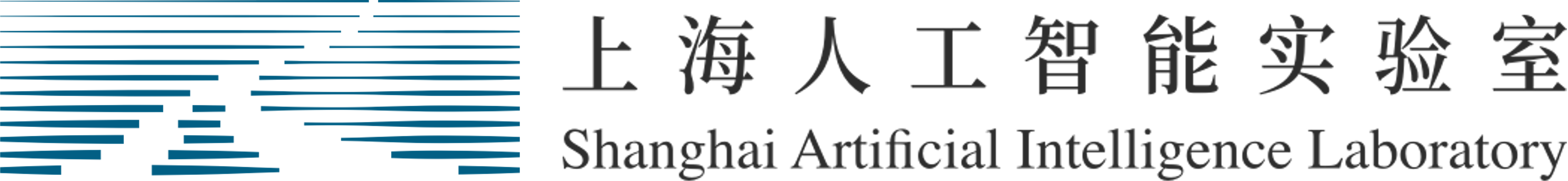}}}%
  \rhead{\raisebox{-2pt}{\includegraphics[height=12pt]{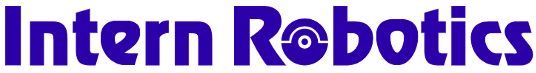}}}%
}
\maketitle
\thispagestyle{firstpage}
\lhead{Preprint}
\vspace{-10pt}
\begin{abstract}
Frontier multimodal large language models (MLLMs) can control robots without robot-specific training, but how their capabilities compare and interact with post-trained vision-language-action (VLA) and world action model (WAM) policies across manipulation demands remains to be thoroughly evaluated. In this report, we evaluate GPT-6-Astra on EBench as the core of an agent that controls a robot through an agent harness under the same visual and proprioceptive observation protocol as post-trained policies. We study two agent configurations: direct control, in which the agent generates waypoint commands directly, and harness with policy, in which it can also invoke a post-trained VLA or WAM policy for action execution. Under direct control, the agent achieves competitive overall performance relative to post-trained policies. It leads on short-horizon tasks and exhibits the least performance variation across generalization dimensions, whereas it underperforms post-trained policies on long-horizon tasks, tabletop manipulation, and high-precision contact-rich tasks. In the harness with policy, invoking a post-trained VLA or WAM combines the complementary strengths of both components, raising overall task success to the level of the strongest compared post-trained policies while substantially reducing token cost. However, performance remains below the per-task upper bound of the standalone components due to suboptimal policy invocation, fixed-instruction execution without subgoals, and grasp disruption when the agent resumes control. Overall, our study characterizes the capability boundary of a frontier MLLM agent and suggests directions for improving future agent harness design. Project page: \url{https://internrobotics.github.io/EBench-Astra-Tech-Report/}.
\end{abstract}
\vspace{-10pt}

\section{Introduction}
\label{sec:intro}

Transferring knowledge from large pretrained models to robotics has become a primary approach to general-purpose robot control~\citep{firoozi2023foundation, ma2024vlasurvey, shen2026wamsurvey}. Vision-language-action (VLA) models adapt vision-language backbones for end-to-end visuomotor control~\citep{kim2024openvla, black2025pi0, pi05, ma2026internvla, gigabrainteam2026gigabrain, yuan2026qwenrobotmanip}, whereas world action models (WAMs) jointly model future visual dynamics and actions to utilize the physical and temporal priors of video generative models~\citep{dreamzero, kim2026cosmos, li2026lingbotva, yuan2026fastwam, wang2026openwam, cai2026internw0}. Both approaches rely on post-training with robot manipulation trajectories: although effective within their training distributions, their generalization is bounded by the scale and diversity of collected robot data. By contrast, frontier multimodal large language models (MLLMs) acquire stronger emergent capabilities from massive-scale pretraining data, and this training scale remains costly to replicate with embodied data alone.

Without robot-specific post-training, frontier MLLMs provide an alternative approach through their spatial understanding and embodied reasoning capabilities~\citep{zhang2026robodojo, google2026argonrobotics}. \astra~\citep{openai2026astra}, for example, can generate end-effector (EEF) commands directly from camera observations and language instructions~\citep{zhang2026robodojo, liu2026roboicl, jia2026agent, lou2026knowbody}. Combined with an agent harness that supplies its prompt, observation interface, and tools, the model serves as the core of an agent that controls the robot in closed loop, as formalized in Section~\ref{sec:framework}. Beyond direct control, recent studies explore a harness with policy in which the agent invokes a post-trained policy for action execution~\citep{zhang2026harnessvla, baton, wang2026harnesspai}. Such a harness combines the agent's task planning, progress monitoring, and error recovery with the closed-loop visuomotor control that the policy learns from robot demonstrations.

Although prior studies have evaluated both direct control and a harness with policy, they have not comprehensively profiled these agent configurations against post-trained VLA and WAM policies across diverse manipulation demands. Specifically, two central questions remain to be addressed. First, under direct control, on which tasks does the agent outperform post-trained VLAs and WAMs, where does it fall short, and what limits its performance? Second, in a harness-with-policy setting, how does invoking a post-trained policy complement the agent's direct control, which limitations of direct control can it alleviate, and what new failure modes arise when the agent invokes a policy for execution and subsequently resumes decision-making?

To further explore these questions, we evaluate \astra as the core of an agent on EBench~\citep{gao2026ebench}. This diagnostic manipulation benchmark categorizes tasks by mobility, manipulation precision, and task horizon and evaluates robustness along the generalization dimensions of unseen objects, unseen backgrounds, paraphrased instructions, and their mixture. For a fair comparison, the agent operates under the same observation protocol as post-trained policies, receiving only RGB camera images and proprioceptive robot states without depth, camera calibration, or simulator ground truth. For each task, the agent receives a task description annotated by \astra from a single training episode as single-shot in-context learning (ICL). Across all 510 test episodes, we evaluate both direct control and a harness with policy invoking either the VLA $\pi_{0.5}$ or the WAM InternW0-$\Delta$, compare each configuration against the post-trained policies on the EBench leaderboard, and analyze execution trajectories across paired episodes.

Under direct control, the \astra-based agent achieves a 46.7\% overall success rate and a 0.654 stage-completion score, outperforming post-trained policies such as Qwen-RobotManip and $\pi_{0.5}$ trained on 6{,}600 in-domain demonstrations. Among the compared post-trained policies, the agent ranks first on short-horizon tasks with a 73.2\% success rate, shows the smallest success-rate variation across the four generalization dimensions, and completes a novel composite instruction zero-shot on which both compared post-trained policies fail. Conversely, its success rate drops to 27.3\% on long-horizon tasks, 20.0\% on tabletop manipulation, and 11.3\% on high-precision contact-rich tasks. Trajectory analysis shows that while the agent performs well at visual grounding and in-episode error recovery, its physical execution is limited by long action chunks between observations and by contact geometry that it must infer explicitly from RGB views.

Invoking a post-trained policy through the agent harness increases the overall success rate to 53.0\% with $\pi_{0.5}$ and 56.0\% with InternW0-$\Delta$ and reduces the agent's token cost per success by more than half. With InternW0-$\Delta$, the harness surpasses all post-trained policies compared in Table~\ref{tab:overall}. The harness with policy retains the agent's high success rate on short-horizon tasks and improves performance on long-horizon and bimanual tasks where the post-trained policy is competent. At the same time, case analysis identifies two limitations of the current harness with policy. First, neither component resolves high-precision contact-rich manipulation. Second, agent and policy coordinate loosely in three respects: the agent invokes the policy for most execution steps even on tasks where that policy alone rarely succeeds, the policy follows the fixed task instruction instead of the agent's subgoal, and the agent often leaves a held grasp unmaintained when it resumes control after a policy invocation, which makes object drops more frequent.

The main contributions of this report are as follows:
\begin{itemize}[leftmargin=*,itemsep=2pt,topsep=3pt]
  \item \textbf{Paired evaluation across agent configurations and post-trained policies.} Under the same observation protocol as post-trained policies, we evaluate an \astra-based agent under direct control and in a harness with policy across all 510 EBench test episodes. We compare both agent configurations against standalone post-trained VLA and WAM policies across all task groups and generalization dimensions.
  \item \textbf{Characterization of agent capabilities under direct control.} We locate where the \astra-based agent leads or trails eight post-trained policies while using only a single annotated task demonstration and no parameter updates, trace its unsuccessful episodes to their failure mechanisms, and examine its in-episode adaptation, the safety of its execution, and its zero-shot composition of skills on a novel instruction. Together, these analyses show where the agent and the policies complement each other.
  \item \textbf{Analysis of harness with policy and implications for agent harness design.} We evaluate the harness with either policy against each of its components on every task, trace the episodes that newly fail under the harness and those that fail with both policies to their decisive incidents, and analyze the agent's policy invocation and the resulting token cost. These analyses locate the remaining gap to the better component in policy invocation, subgoal passing, and grasp maintenance when the agent resumes control, which informs the design of future agent harnesses.
\end{itemize}

\FloatBarrier

\section{Related Work}
\label{sec:related}

\paragraph{Robot manipulation with foundation models.} Two lines of work apply foundation models to robot manipulation. \textit{Post-trained policies} typically adapt foundation models on robot manipulation trajectories, including VLA models that fine-tune vision-language backbones for action prediction~\citep{kim2024openvla, black2025pi0, pi05} and WAMs that jointly model future visual dynamics and actions~\citep{dreamzero, kim2026cosmos, yuan2026fastwam, miao2026internw0deltaworldactionmodel}. \textit{Direct control with frontier MLLMs}, in contrast, uses frontier MLLMs as the cores of agents that control robots without parameter updates. Early approaches translate model predictions into robot motion through generated programs that call perception and control APIs~\citep{codeaspolicies} or through embodiment-specific semantic primitives~\citep{chen2026showharness}, whereas more recent studies prompt MLLMs to output EEF or joint commands directly~\citep{jia2026agent, lou2026knowbody}. On the RoboDojo benchmark, for example, \astra generates robot actions directly from observations~\citep{zhang2026robodojo}, and providing in-context demonstration segments together with in-episode interaction history further improves task success~\citep{liu2026roboicl, sensenova2026astra}. Rather than evaluating direct control in isolation, our work systematically compares \astra under direct control against post-trained VLA and WAM policies on EBench~\citep{gao2026ebench} by task group and generalization dimension, linking dimension-level performance to step-by-step execution trajectories.

\paragraph{Harness with policy.} Combining both lines, recent frameworks build hierarchical harnesses in which an agent with an MLLM core invokes post-trained policies for action execution~\citep{qin2026towards, sensenova2026astra}. For instance, Harness VLA invokes a frozen VLA alongside scripted or analytical primitives~\citep{zhang2026harnessvla}, BATON coordinates long-horizon VLA execution through a transition-aware memory that verifies subtask entry and exit conditions~\citep{baton}, and HarnessPAI synthesizes task-specific control programs that invoke post-trained policies without online deliberation~\citep{wang2026harnesspai}. Related frameworks also adapt harness coordination logic, policy interfaces, or recovery tools during deployment~\citep{tang2026adahvla, li2026racap, ding2026zetta}. However, existing studies primarily evaluate each proposed harness as a single integrated system. In contrast, we evaluate the agent both under direct control and in a harness with policy, and compare both agent configurations against standalone VLA and WAM policies across the same test scenes under a fixed in-context protocol. This paired evaluation quantifies how the agent's policy invocation decisions align with the standalone competence of the post-trained policy, and traces episode-level gains and losses to the agent's direct control, policy execution, or the transitions between them.

\FloatBarrier
\section{Problem Formulation}
\label{sec:framework}

To compare a frontier MLLM with post-trained policies on a common benchmark, we first define the agent and its harness, and then specify the two agent configurations evaluated in this study: direct control and harness with policy.

\subsection{Agent and Agent Harness}
\label{sec:framework:definition}

An \emph{agent} $(F, H)$ consists of an MLLM core $F$ and an \emph{agent harness} $H = (p_H, \mathcal{O}_H, \mathcal{T}_H)$ through which it controls a robot. In this formulation, the prompt $p_H$ specifies the robot, the rules of interaction, and the use of each tool. The observation interface $\mathcal{O}_H$ determines which sensor readings and robot states reach the core, and the tool set $\mathcal{T}_H$ lists the calls available to the core together with how each is executed. Because the harness runs the decision loop, executing each call and returning its result to the core, the agent defines a closed-loop policy $\pi_A$ without robot-specific post-training of its core. For a task $\tau$ on an episode scene $s$ with language instruction $\ell_s$ and task context $c_\tau$, at each decision step $t$ the agent issues a tool call $a_t$ conditioned on the episode interaction history $h_t = (o_1, a_1, \dots, o_t)$, where $o_1$ is the initial observation and each later $o_t$ is the result returned for $a_{t-1}$:
\begin{equation}
  a_t \sim \pi_A\big(\cdot \mid h_t,\, \ell_s,\, c_\tau\big), \qquad a_t \in \mathcal{A}_H \cup \{\texttt{stop}\}, \qquad \mathcal{A}_H = \mathcal{A}_H^{\mathrm{act}} \cup \mathcal{A}_H^{\mathrm{info}},
  \label{eq:agent_policy}
\end{equation}
where $\mathcal{A}_H$ contains the calls of the tools in $\mathcal{T}_H$ with their admissible arguments. A call in $\mathcal{A}_H^{\mathrm{act}}$ advances the physics, either by commanding the robot through joint or EEF targets, gripper commands, motion primitives, or programs that compose them, or by invoking a controller such as a post-trained policy for a requested number of physics steps, and returns an updated observation through $\mathcal{O}_H$. A call in $\mathcal{A}_H^{\mathrm{info}}$ returns information without advancing the physics, such as the current camera views and robot state or a stored demonstration. Agent control ends when the agent issues $\texttt{stop}$ or reaches the physics-step limit. Unlike post-trained VLA and WAM policies that learn visuomotor control through parameter updates on robot trajectories, the agent adapts online solely through in-context conditioning, visual and proprioceptive observations, and tool calls.

\subsection{Direct Control and Harness with Policy}
\label{sec:framework:execution}

The two agent configurations evaluated in this study share the observation interface $\mathcal{O}_H$ and differ in the tool calls $\mathcal{A}_H$ exposed by the agent harness. Under \emph{direct control}, $\mathcal{A}_H^{\mathrm{act}} = \mathcal{A}_{\mathrm{EEF}}$, which consists of EEF, base, and gripper commands, and $\mathcal{A}_H^{\mathrm{info}}$ contains only a call that returns the current views and robot state, so the agent generates all motion commands itself. Under \emph{harness with policy}, following Harness VLA~\citep{zhang2026harnessvla}, the harness expands the action set to $\mathcal{A}_H^{\mathrm{act}} = \mathcal{A}_{\mathrm{EEF}} \cup \mathcal{A}_{\mathrm{pol}}$, where $\mathcal{A}_{\mathrm{pol}}$ denotes invocations of a post-trained policy, and adds to $\mathcal{A}_H^{\mathrm{info}}$ a call that returns the annotated task demonstration on request. In each policy invocation, the agent specifies a step limit within which the post-trained policy executes autonomously under the fixed task instruction before returning an updated observation through $\mathcal{O}_H$. In this configuration, the agent alternates between issuing direct waypoint commands and invoking the post-trained policy for closed-loop trajectory execution, while retaining task planning, progress monitoring, and error recovery. Section~\ref{sec:capability} characterizes the capability boundary under direct control, and Section~\ref{sec:tool} evaluates how the harness with policy affects performance across capability dimensions.

\section{Experimental Setup}
\label{sec:setup}

\begin{figure*}[t]
  \centering
  \includegraphics[width=\linewidth]{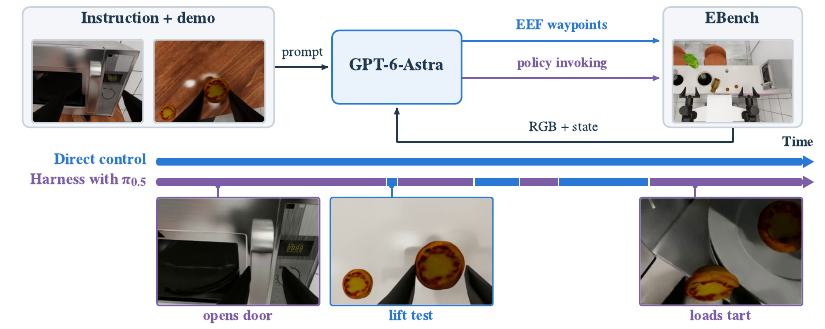}
  \caption{Overview of the \astra control loop on EBench and executor timelines on a Microwave test scene. The input prompt combines the task instruction with a detailed task description annotated from a single training episode. The timeline bars compare active executors across an episode under direct control and under the harness with $\pi_{0.5}$, with blue segments executed by the agent and purple segments by $\pi_{0.5}$, and with keyframes showing right-wrist camera views from that episode at the marked steps.}
  \label{fig:overview}
\end{figure*}

\subsection{Benchmark}
\label{sec:setup:benchmark}

EBench~\citep{gao2026ebench} is a diagnostic simulation benchmark for generalist robot manipulation, built on GenManip~\citep{gao2025genmanip} and Isaac Sim with the ARX Lift~2 dual-arm mobile robot. Its 26 tasks span 9 scene categories across household, retail, and industrial settings, testing functional contexts such as food preparation, household cleaning, retail checkout, and component assembly. Completing these tasks requires relating object functions and spatial layouts to the task goal and sequencing the necessary manipulation steps. EBench annotates each task along five capability dimensions: scene, atomic skill, operating mode, manipulation precision, and task horizon. Our analyses mainly group tasks by the last three, and Appendix~\ref{app:skills} reports the results of Section~\ref{sec:tool} by atomic skill. Operating mode, which we refer to as mobility, separates 19 mobile from 7 tabletop tasks, manipulation precision separates 14 low-, 8 medium-, and 4 high-precision tasks, and task horizon separates 11 short- from 15 long-horizon tasks. The 510 test episodes span four generalization dimensions: Background with unseen backgrounds, Object with unseen object instances, Instruction with paraphrased instructions, and Mix with their joint perturbation.

The evaluation covers all 26 test tasks across 510 episodes, with 20 episodes for each of 24 tasks and 15 each for Sandwich and Microwave, spanning 468 distinct scenes. Performance is evaluated using two official metrics: task success rate (SR) and stage-completion score, which assigns partial credit for intermediate stages completed by the end of an episode according to task-specific rules. Both metrics are first averaged across episodes within each task and then averaged equally across all 26 tasks. Each episode is bounded by a task-specific physics-step limit ranging from 1{,}500 to 5{,}000 steps at 30\,Hz, which is not disclosed to the agent.

\subsection{Agent and Control Interface}
\label{sec:setup:agent}

We use \astra~\citep{openai2026astra} with high reasoning effort as the core of the agent and initiate an independent session thread for each episode. At each decision step, based on the current visual and proprioceptive observations, the agent outputs one or more EEF and base waypoints, including target gripper states and execution durations. Meanwhile, the agent provides a brief text explaining the rationale for the command, which we refer to as its \emph{stated reason}. These waypoints are interpolated into an action chunk and executed in the simulator, after which the updated observations are returned. The shared task prompt includes empirical operational routines, such as grasp verification: closing the gripper while stationary, lifting by 1\,cm, and confirming that the object moves with the gripper, as detailed in Appendix~\ref{app:prompts}. Token usage, tool calls, and inference time are unconstrained. If the agent terminates an episode before the physics-step limit, the robot holds its final pose until the evaluation server computes the official SR and score.

\paragraph{Observation protocol.}
\label{sec:setup:protocol}
Across both direct control and the harness with policy, the agent is evaluated under the same observation protocol as post-trained policies, receiving only RGB camera observations and proprioceptive robot states through pre-defined tools. Specifically, the interface supplies the views of three RGB cameras: the overview camera, termed \texttt{overlook} in the prompts, and the left and right wrist cameras. All three views are provided at the first step and can be requested selectively thereafter, together with the proprioceptive state comprising both EEF poses, gripper widths, the mobile base pose, and elapsed simulator steps. It provides no depth maps, camera calibration parameters, ground-truth object poses, segmentation masks, intermediate reward signals, subgoal indicators, or remaining step budgets. Official SR and score values are computed only after an episode terminates, and no state or experience is retained across episodes.

\paragraph{Annotated task demonstrations.} Each task is accompanied by a detailed task description derived from a single demonstration in the EBench training set as single-shot ICL. Specifically, prior to evaluation, a separate \astra instance with high reasoning effort inspects one training trajectory, selects 10--18 keyframes, summarizes the operation sequence, and pairs the selected keyframes with the recorded robot states and actions, as shown in Appendix~\ref{app:demos}. During evaluation, the agent is instructed to adapt the demonstrated procedure and spatial relations to current visual observations rather than replaying recorded coordinates.

\subsection{Configurations}
\label{sec:setup:configs}

We evaluate the \astra-based agent under two agent configurations, direct control and harness with policy instantiated with either $\pi_{0.5}$ or InternW0-$\Delta$, as illustrated in Figure~\ref{fig:overview}.

\paragraph{Direct control.} The agent harness exposes only the EEF waypoint interface described above, and the agent invokes no external policy. This configuration isolates the standalone manipulation capability of the MLLM core.

\paragraph{Harness with policy.} In this configuration, the agent retains the EEF waypoint interface and can additionally invoke one of two frozen post-trained policies as a tool: the official leaderboard checkpoint of the VLA $\pi_{0.5}$~\citep{pi05} or the WAM InternW0-$\Delta$~\citep{miao2026internw0deltaworldactionmodel}, both post-trained on EBench demonstrations. For each policy invocation, the agent specifies a step limit of 1--300 physics steps together with a stated reason. Within this step limit, the post-trained policy executes autonomously under the fixed task instruction, after which the updated visual and proprioceptive observations are returned to the agent. The harness also lets the agent query the annotated task demonstration again at any point. At each decision step, the agent chooses between issuing direct EEF waypoints and invoking the post-trained policy. In other words, the agent controls when and for how many steps the policy executes, but cannot pass agent-modified task prompts to the underlying policy, as discussed in Section~\ref{sec:tool:cases}.

\subsection{Compared Policies}

\begin{table}[t]
  \centering
  \fontsize{8}{9}\selectfont
  \setlength{\tabcolsep}{1.5pt}
  \caption{SR and score on the EBench test set, overall and by mobility group, of the eight post-trained policies with per-task results on the EBench leaderboard and of the \astra-based agent under direct control and in the harness with each policy. Best and second-best values in each column are in bold and underlined. MLLM denotes an agent whose motion commands all come from its MLLM core, and Hybrid an agent that also invokes a post-trained policy.}
  \label{tab:overall}
  \begin{tabularx}{\linewidth}{>{\raggedright\arraybackslash}p{3.05cm}>{\centering\arraybackslash}p{1.05cm}*{6}{>{\centering\arraybackslash}X}}
\toprule
 & & \multicolumn{2}{c}{\hd{Overall}} & \multicolumn{2}{c}{\hd{Mobile}} & \multicolumn{2}{c}{\hd{Tabletop}} \\
\cmidrule(lr){3-4}\cmidrule(lr){5-6}\cmidrule(lr){7-8}
\hd{Policy} & \hd{Type} & \hd{SR (\%)} & \hd{Score} & \hd{SR (\%)} & \hd{Score} & \hd{SR (\%)} & \hd{Score} \\
\specialrule{\lightrulewidth}{\aboverulesep}{0pt}
\rowcolor{tabband} \multicolumn{8}{c}{\textit{\textbf{Post-trained policies}}} \\
\specialrule{\lightrulewidth}{0pt}{\belowrulesep}
OpenWAM-$\alpha$ & WAM & \second{55.3} & \second{0.701} & 60.2 & 0.740 & \second{42.1} & \second{0.594} \\
InternW0-$\Delta$ & WAM & 49.2 & 0.660 & 55.0 & 0.700 & 33.6 & 0.552 \\
Qwen-RobotManip & VLA & 45.6 & 0.608 & 43.9 & 0.576 & \best{50.0} & \best{0.696} \\
$\pi_{0.5}$ & VLA & 41.4 & 0.544 & 45.9 & 0.578 & 29.3 & 0.452 \\
InternVLA-A1.5 & VLA & 34.2 & 0.468 & 42.3 & 0.562 & 12.1 & 0.215 \\
$\pi_0$ & VLA & 33.7 & 0.475 & 34.0 & 0.469 & 32.9 & 0.491 \\
GigaBrain-0.7 & VLA & 33.3 & 0.460 & 31.6 & 0.413 & 37.9 & 0.588 \\
Fast-WAM & WAM & 25.6 & 0.371 & 34.0 & 0.460 & 2.9 & 0.129 \\
\specialrule{\lightrulewidth}{\aboverulesep}{0pt}
\rowcolor{tabband} \multicolumn{8}{c}{\textit{\textbf{\astra-based agent configurations}}} \\
\specialrule{\lightrulewidth}{0pt}{\belowrulesep}
\rowcolor{astralight} Direct control & MLLM & 46.7 & 0.654 & 56.6 & 0.730 & 20.0 & 0.447 \\
\rowcolor{astralight} Harness with $\pi_{0.5}$ & Hybrid & 53.0 & 0.692 & \second{61.0} & \second{0.771} & 31.4 & 0.480 \\
\rowcolor{astralight} Harness with InternW0-$\Delta$ & Hybrid & \best{56.0} & \best{0.744} & \best{64.3} & \best{0.813} & 33.6 & 0.556 \\
\bottomrule
\end{tabularx}

\end{table}

We compare the \astra-based agent under direct control and in the harness with $\pi_{0.5}$ or InternW0-$\Delta$ against the post-trained VLA and WAM policies on the EBench leaderboard, as summarized in Table~\ref{tab:overall}. Dimension-level and per-task analyses in Section~\ref{sec:capability} compare the agent against eight post-trained policies with per-task results on the 510 test episodes: OpenWAM-$\alpha$~\citep{wang2026openwam}, InternW0-$\Delta$~\citep{miao2026internw0deltaworldactionmodel}, Qwen-RobotManip~\citep{yuan2026qwenrobotmanip}, $\pi_{0.5}$~\citep{pi05}, InternVLA-A1.5~\citep{ma2026internvla}, $\pi_0$~\citep{black2025pi0}, GigaBrain-0.7~\citep{gigabrainteam2026gigabrain}, and Fast-WAM~\citep{yuan2026fastwam}, all post-trained on the public EBench training set of 6{,}600 episodes across the 26 tasks. All comparisons across configurations of the agent are paired by test episode, with statistical procedures detailed in Appendix~\ref{app:stats}.

\section{Capability Boundary Under Direct Control}
\label{sec:capability}

This section evaluates the \astra-based agent under direct control, in which the harness exposes observations and EEF waypoint commands but no post-trained policy. Every motion of the robot is therefore commanded by \astra itself, so the results isolate the manipulation capabilities of the MLLM. We first compare overall performance, task groups, and generalization dimensions with those of the eight post-trained policies, and then trace the failure patterns of unsuccessful episodes, the agent's adaptation within an episode, and the safety risks of its execution. Through case studies and a pilot experiment, we further examine the role of the in-context demonstration and zero-shot compositional generalization, and finally summarize the resulting capability boundary.

\subsection{Overall Performance}
\label{sec:capability:overall}

The \astra-based agent under direct control receives a single annotated demonstration per task and no parameter updates, yet it is competitive with VLA and WAM policies post-trained on 6{,}600 demonstrations of the same tasks. Compared with the eight post-trained policies in Table~\ref{tab:overall}, the direct-control agent ranks third on both metrics with 46.7\% SR and a 0.654 score. It trails only two WAMs, as OpenWAM-$\alpha$ reaches 55.3\% SR and a 0.701 score and InternW0-$\Delta$ reaches 49.2\% SR and a 0.660 score. It outperforms the remaining six policies, including the VLAs Qwen-RobotManip with 45.6\% SR and a 0.608 score and $\pi_{0.5}$ with 41.4\% SR and a 0.544 score. Appendix Tables~\ref{tab:per_task_sr} and~\ref{tab:per_task_score} report the per-task SR and score.

\begin{figure}[ht]
  \centering
  \includegraphics[width=\linewidth]{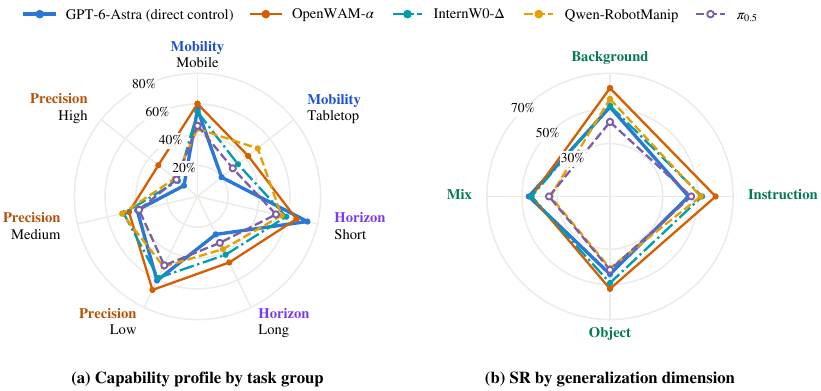}
  \caption{SR of the direct-control agent and of OpenWAM-$\alpha$, InternW0-$\Delta$, Qwen-RobotManip, and $\pi_{0.5}$ (a) across mobility, horizon, and precision task groups and (b) across the four generalization dimensions of the EBench test set. The radial axis of each panel ends at its largest SR, rounded up to a multiple of 10\%.}
  \label{fig:radar}
\end{figure}

The gap between score and SR shows that the direct-control agent frequently earns stage credit in episodes that it does not finish. SR counts only complete successes, whereas the score also credits the stages completed before an unsuccessful episode ends. The SR of the direct-control agent trails that of InternW0-$\Delta$ by 2.5 percentage points and exceeds that of Qwen-RobotManip by only 1.1 points, whereas its score nearly matches that of InternW0-$\Delta$ and clearly exceeds those of Qwen-RobotManip and $\pi_{0.5}$. More generally, its score exceeds its SR by a larger margin than that of any post-trained policy. Section~\ref{sec:capability:profile} therefore first uses Figure~\ref{fig:radar} to locate where the agent leads and trails, and Section~\ref{sec:capability:failures} then examines the stages at which its episodes fail.

\subsection{Capability Profile Across Task Groups and Generalization Dimensions}
\label{sec:capability:profile}
\label{sec:capability:strengths}

\begin{figure}[t]
  \centering
  \includegraphics[width=0.84\linewidth]{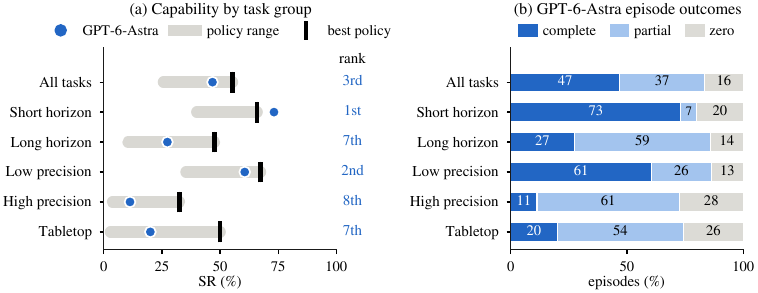}
  \caption{The direct-control agent across the 26 EBench test tasks by task group. (a) SR of the direct-control agent, range across the eight post-trained policies, and best post-trained policy, with its rank among the agent and the eight post-trained policies in the column on the right of the panel. (b) Shares of its episodes ending in complete success, partial credit, or a score of zero, averaged equally across tasks within each group and rounded to whole percent so that each bar sums to 100.}
  \label{fig:summary}
\end{figure}

Across task groups, the performance of the direct-control agent splits by the type of demand rather than declining uniformly with task difficulty, as shown in Figures~\ref{fig:radar}a and~\ref{fig:summary}a and Appendix Figure~\ref{fig:profile}. It ranks first on short-horizon tasks and second on mobile and low-precision tasks, where only OpenWAM-$\alpha$ leads it. It drops to seventh on long-horizon and tabletop tasks and to eighth on high-precision tasks. The following paragraphs profile the agent along the horizon, mobility, and precision of the tasks and along the generalization dimensions, and Section~\ref{sec:capability:failures} traces the failures behind the trailing ranks to their mechanisms.

\paragraph{Task horizon.} The horizon of a task separates the agent's strongest group from one of its weakest. On the 11 short-horizon tasks, it achieves the highest SR of 73.2\% and score of 0.766, surpassing all eight post-trained policies including OpenWAM-$\alpha$ at 65.9\% SR and a 0.700 score. Eight of these 11 tasks require only low precision, and most of them consist of locating an object and its target, repositioning the base, and carrying the object between them, which the agent plans from its current views and executes with base and EEF waypoints. When a task reduces to such coarse transport between located targets, a single annotated demonstration and the priors of the MLLM therefore suffice for the agent to exceed policies post-trained on the full training set across this group. On the 15 long-horizon tasks, by contrast, the agent reaches only 27.3\% SR, whereas OpenWAM-$\alpha$ reaches 47.6\%. On these tasks, the agent often completes the early stages but not the full task. On Detergent, for example, it reaches a mean score of 0.80 but completes the full task in only 55\% of episodes.

\paragraph{Mobility.} The mobility groups order task difficulty differently for the agent than for several post-trained policies. Because 10 of the 11 short-horizon tasks are mobile, the agent ranks second on the 19 mobile tasks with 56.6\% SR. On the seven tabletop tasks, its SR falls to 20.0\%, whereas Qwen-RobotManip reaches 50.0\%. Qwen-RobotManip and GigaBrain-0.7, by contrast, achieve a higher SR on tabletop tasks than on mobile tasks. The tabletop gap extends beyond the four high-precision tabletop tasks to the other three: Frame, Flip cup \& cookies, and Glass into box. On these three tasks, the agent reaches a mean SR of 31.7\%, whereas Qwen-RobotManip reaches 91.7\%. On the mobile tasks, in turn, the agent's leading results rest on visual grounding. The Bookmark task requires locating a thin, flat bookmark on a desk crowded with visually similar stationery and everyday objects and placing it onto an open book. The agent succeeds in 90\% of these episodes, whereas $\pi_{0.5}$ reaches 55\% and the remaining seven post-trained policies reach 25\%--45\%. In the Teacup \& teapot task, the agent has to tell the two objects apart and place the teacup on the saucer and the teapot on the tray. It reaches 85\% SR and a 0.93 score, whereas InternW0-$\Delta$ and Qwen-RobotManip lead the post-trained policies with 45\% SR. In the Remote task, the agent must place the remote into a designated compartment of a divided holder. It succeeds in every episode, whereas InternW0-$\Delta$ succeeds in 85\%. Perfume also targets a compartment of a divided organizer, and on this task the agent ties OpenWAM-$\alpha$ for the highest SR at 75\% with a higher score of 0.88 against 0.78. Bookmark, Teacup \& teapot, and Remote are the three tasks on which the agent leads all eight post-trained policies. These three tasks and Perfume all require locating the specified object among distractors or the specified compartment among several before placing the object there. These leads therefore indicate that the agent grounds objects and targets reliably, as Appendix Figure~\ref{fig:successful} illustrates for Bookmark and Remote from the grasp approach to the placement.

\paragraph{Manipulation precision.} Success declines most steeply with the precision that a task demands. The SR of the agent falls from 60.6\% on low-precision tasks to 40.2\% on medium-precision tasks and 11.3\% on high-precision tasks. Its rank drops accordingly from second to fourth and eighth. On high-precision tasks, OpenWAM-$\alpha$ and $\pi_{0.5}$ retain 32.5\% and 17.5\%, respectively. On Peg in hole and Tighten nut, the agent reaches SRs of only 20\% and 10\% yet mean scores of 0.60 and 0.55. Every unsuccessful episode on these tasks earns 0.50 for positioning the part at the target location and then fails at the final insertion or threading step, as shown in Appendix Figure~\ref{fig:failure_modes}. The agent's failures on these two tasks thus concentrate on the final contact step.

\paragraph{Generalization dimensions.} Across the generalization dimensions, the agent's strength holds even with minimal task context. The post-trained policies are trained on 6{,}600 episodes, whereas the direct-control agent receives for each task only one annotated demonstration derived from a single training episode. This demonstration cannot fully illustrate the shifts that the test episodes apply along the four dimensions of Object, Background, Instruction, and Mix. Even so, the SR of the direct-control agent stays between 44.2\% and 50.8\% across the four dimensions, as shown in Figure~\ref{fig:radar}b and Appendix Table~\ref{tab:generalization}. This 6.6-point range is the smallest among the agent and the eight post-trained policies. The four policies in Figure~\ref{fig:radar}b vary more widely: 7.7 points for InternW0-$\Delta$, 11.5 points for $\pi_{0.5}$, 16.9 points for OpenWAM-$\alpha$, and 21.5 points for Qwen-RobotManip. What the agent takes from the demonstration therefore transfers across these shifts. Moreover, the advantage grows under the Mix dimension, where the object instance, background texture, and instruction phrasing change jointly. Every post-trained policy reaches its lowest SR under Mix. OpenWAM-$\alpha$, for example, falls from 61.5\% under Background to 44.6\% under Mix, and Qwen-RobotManip falls from 55.4\% to 33.8\%. The direct-control agent, by contrast, maintains 46.2\% SR and a 0.643 score under Mix and ranks first on both metrics. In conclusion, the generalization shifts limit the post-trained policies more than the direct-control agent. The limits of the agent lie instead in the manipulation demands of the long-horizon, tabletop, and high-precision groups, which the next subsection examines.

\subsection{Failure Patterns and Token Cost}
\label{sec:capability:failures}
\label{sec:capability:deficits}

Most unsuccessful episodes of the direct-control agent end with partial progress rather than failing initially, which explains the score margin of Section~\ref{sec:capability:overall}. Of the 510 episodes, 237 end in complete success, and 188 of the 273 unsuccessful episodes achieve a nonzero score, so only 85 end at zero, as shown in Figure~\ref{fig:summary}b. Zero scores concentrate on Gear, Utensils, and Glass into box, where the agent fails before any scored progress in 75\%, 65\%, and 50\% of episodes, respectively. The trajectory analysis below traces the unsuccessful episodes to four mechanisms: imprecise contact geometry, delayed feedback over long horizons, uncoordinated bimanual motion, and unconfirmed completion claims. A final paragraph measures the token cost of the resulting failures.

\paragraph{Contact geometry at the final step.} Precise contact fails where the agent has to infer geometry from images. The agent does not infer the surface height, part clearance, and contact geometry precisely enough for the final insertion and threading from RGB views and the base-relative EEF pose alone. Post-trained policies receive the same RGB views without depth but map them to contact motions learned from demonstration trajectories. The same limit appears on tasks with moderate nominal precision. On Glass into box, closing the case requires folding both temples inward, but the agent's corrective pushes leave a temple protruding and prevent the lid from closing, as shown in Appendix Figure~\ref{fig:policies}b.

\paragraph{Delayed feedback and cascading disturbances over long horizons.} The rollouts of long-horizon tasks point to two causes of failure. First, the agent detects disturbances late and recovers from them slowly. It receives an observation only after each commanded waypoint chunk finishes. Although 84\% of its motion commands consist of a single waypoint, each command runs for a median of 48 physics steps, whereas $\pi_{0.5}$ and InternW0-$\Delta$ replan every 30 and 10 physics steps, as Section~\ref{sec:tool:design} describes. A grasp loss or contact change within a chunk therefore persists until the chunk returns, and the retries that follow consume the remaining physics steps. As a result, the step limit ends 229 of all 273 unsuccessful episodes and is by far their most common ending. Second, the agent plans its approaches and grasps without anticipating how contact will move the object, so the scene becomes unstable over a long task. On Detergent, a bottle tips over or falls during the agent's own approach, grasp, or transport in 19 of the 20 episodes, and on Bottle this happens in 11 of the 20. The agent must then grasp the fallen bottle again from the table. In one Detergent episode, for example, the agent places the first of three bottles within 11 commands, but the second bottle falls during the next command, which withdraws and approaches over 100 physics steps. The agent spends the remaining 92 commands grasping that bottle again and then the third bottle, which slips during transport. The episode reaches the step limit with two bottles placed. Such cascades accumulate most on Bottle, which is one of the two longest tasks. No evaluated policy or configuration completes Bottle, as analyzed in Section~\ref{sec:tool:shared}.

\paragraph{Uncoordinated bimanual motion.} On bimanual tasks, the agent coordinates its two arms only between waypoint chunks. Each chunk executes open loop, so the agent cannot adjust one arm to the contact that the other arm makes during execution, and one hand can withdraw or move before the other has secured the shared object. On Flip cup \& cookies, where the cup passes from one hand to the other, the agent fails in 90\% of episodes. Appendix Figure~\ref{fig:failure_modes}c shows one such episode, in which the supporting hand withdraws during the second handover and the cup falls off the table. On Frame, the agent takes the division of labor between the arms from the demonstration, as analyzed in Section~\ref{sec:capability:cases}. Its SR of 65\% on this task exceeds the 50\% of OpenWAM-$\alpha$ but stays below the 95\% of Qwen-RobotManip.

\paragraph{Unconfirmed completion claims.} Finally, the direct-control agent sometimes ends an episode early because it misjudges completion. The agent ends 43 unsuccessful episodes before the step limit, and 21 of them end with a claim of task success while a subgoal remains unmet. Seven of these claims come from Dishwasher, and the agent makes all seven with the door still open, as shown in Appendix Figure~\ref{fig:self_ended}. The instruction of this task asks to put every bowl into the dishwasher, and the last keyframe of the demonstration shows the door closed, yet the agent treats the placement of both bowls as completion. It states an intent to close the door after loading in only one of the 20 episodes, which is also its only success.

\paragraph{Token cost of failures and long horizons.} These failures also dominate the token cost of direct control. The agent resends its growing episode context with every call, so the tokens of an episode grow with its number of calls, as detailed in Appendix~\ref{app:tool:cost}. Across the 510 episodes, the Pearson correlation between the two is $r=0.88$. An unsuccessful episode keeps issuing corrections until it reaches the step limit or ends early. On average, it consumes 5.45M tokens over 55.9 calls, roughly three times the 1.84M tokens and twice the 27.4 calls of a successful episode. Long horizons compound this cost, as an episode of a long-horizon task consumes 5.40M tokens against 1.63M on a short-horizon task while succeeding less often. One success on a long-horizon task therefore costs 20.6M tokens when the tokens spent on unsuccessful episodes within each group are included. This cost is about nine times the 2.23M tokens per success on a short-horizon task.

\subsection{Adaptive Capabilities}
\label{sec:capability:adaptation}

\begin{figure}[t]
  \centering
  \includegraphics[width=\linewidth]{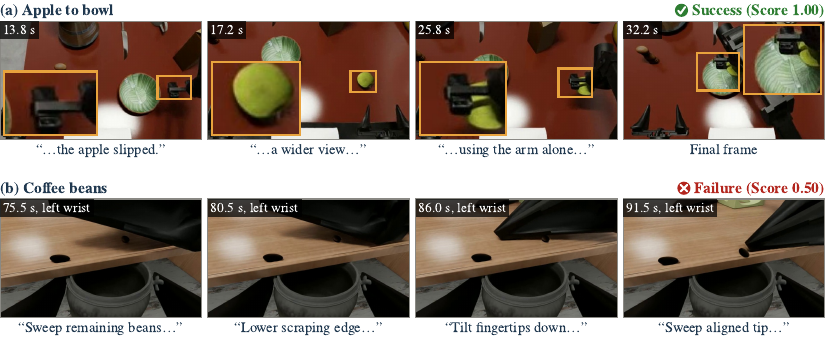}
  \caption{Adaptation in two episodes of the direct-control agent, with the agent's stated reasons below the frames. Orange boxes show enlarged views of the contact and manipulation details. (a) Apple to bowl: after a slip, the agent takes a wider view and carries the apple with the arm alone, reaching a 1.00 score. (b) Coffee beans: after a sweep leaves the beans in place, the agent lowers the edge and tilts the fingertips, reaching a 0.50 score.}
  \label{fig:adaptive}
\end{figure}

The direct-control agent observes the scene only between its own action chunks, yet the returned observations give it in-episode feedback that informs its next decision. After a failed action, the agent records a possible explanation, explores the workspace, and revises its approach. Such in-episode diagnosis is frequent across the 510 episodes: the agent's stated reasons name a slip, a drop, or a displaced object in 302 of them, as detailed in Appendix~\ref{app:events}. The two examples in Figure~\ref{fig:adaptive} show two forms of this adaptation.

First, the agent diagnoses failures that occur while a chunk executes and changes its motion strategy accordingly. For example, in the Apple to bowl task of Figure~\ref{fig:adaptive}a, the interface reports no failure when the apple slips, but the agent infers a possible slip from the closed finger gap and withdraws to obtain a wider view. After regrasping the apple, it avoids the base motion that preceded the slip, transports the apple with the arm alone, and completes the episode. Second, the agent adjusts the contact geometry with the physical guidance supplied in its prompt. In the Coffee beans task of Figure~\ref{fig:adaptive}b, an initial sweep leaves the beans in place. The agent then applies a supplied caveat that the measured local EEF $z$-coordinate is not the height above the tabletop, repeats the sweep 25\,mm lower, and tilts the fingertips downward to establish contact with the surface.

The two episodes thus separate the agent's diagnosis and adaptation from the precision of its physical execution. In both, the agent states a likely cause of the failure and revises its strategy, and the outcome then depends on how precisely the revised motion makes contact. In the Apple to bowl episode, the revised transport completes the task. In the Coffee beans episode, the lowered and tilted sweep follows the supplied caveat quoted in Appendix~\ref{app:cases}, yet the episode ends at a 0.50 score with the collection incomplete. In-episode adaptation therefore reaches the level of diagnosis and strategy, while fine contact marks the boundary that Section~\ref{sec:capability:boundary} locates.

\subsection{Safety in Execution}
\label{sec:capability:safety}

Aggregate task scores do not capture how safely the agent executes its commands, and here the direct-control agent shows clear weaknesses. They stem mainly from unconstrained execution, as the agent drives each command toward its target without accounting for the other arm, the workspace bounds, contact with the scene, or the speed of the motion. Grip losses are common, and more than half of them occur outside grasp tests. Across the 510 episodes under direct control, the gripper-state criteria of Appendix~\ref{app:events} identify 811 grip losses in 333 episodes. Of these losses, 469 in 265 episodes occur in calls whose stated reasons describe no test of a grasp. Recurring patterns in the recorded rollouts illustrate four resulting hazards, and Appendix Figure~\ref{fig:safety} shows the first three. First, independent single-arm commands can destabilize held objects during bimanual tasks. In the Dishwasher task, for example, a bowl slips from the left gripper while the right hand opens the door, as shown in Appendix Figure~\ref{fig:safety}a. Second, recovery attempts can drive the robot beyond safe workspace bounds. After an apple falls outside the workspace, for example, the agent keeps requesting EEF targets beneath the table and behind the robot base, as in Appendix Figure~\ref{fig:safety}b. Third, height and contact with the scene are hard to judge from RGB images without depth. On Coffee beans, a held jar and later the spoon strike the table edge in an awkward pose, while sweeping motions elsewhere pass above the surface, as in Appendix Figure~\ref{fig:safety}c.

The fourth hazard concerns the speed of wrist and joint motion. The agent commands target EEF poses directly and relies on the inverse-kinematics solver to compute joint trajectories, so its commands can turn the wrist through large angles within a few physics steps. In 31 of the 510 episodes, a wrist turns through 90$^\circ$ or more within 24 physics steps, and a joint exceeds 0.08\,rad per physics step in 169 episodes. By comparison, the segments that $\pi_{0.5}$ executes in the harness with policy of Section~\ref{sec:tool} contain no wrist turn this large within 24 physics steps, and they exceed the joint threshold in only 59 episodes, as detailed in Appendix~\ref{app:speed}. On physical hardware, the forces and torques that these fast rotations require risk motor damage or collisions. Together, these cases highlight the need to preserve grasp stability, enforce workspace and motion constraints, and account for contact throughout manipulation before real-world deployment.

\subsection{Case Studies: In-Context Learning, Adaptation, and Precision}
\label{sec:capability:cases}
\label{sec:capability:demonstration}

\begin{figure*}[t]
  \centering
  \includegraphics[width=\linewidth]{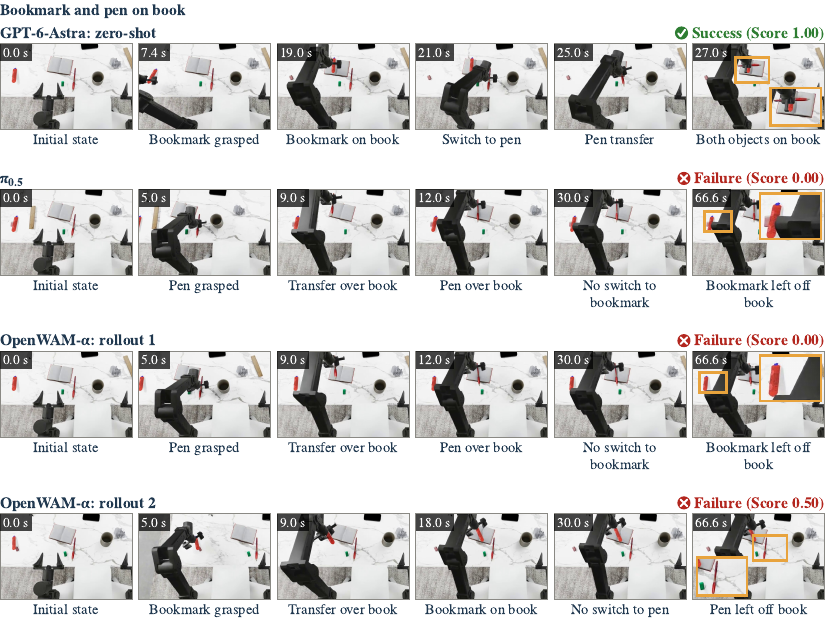}
  \caption{Rollouts of the direct-control agent evaluated zero-shot, $\pi_{0.5}$, and OpenWAM-$\alpha$ on a composite instruction outside the standard EBench tasks: placing a bookmark and then a pen onto the same open book. Keyframes are shown in chronological order with video time.}
  \label{fig:composition}
\end{figure*}

\paragraph{In-context learning.} On Frame and Gear, the annotated demonstration conveys task-level structure to the direct-control agent but not the precise action of a final contact step. Appendix~\ref{app:icl} compares runs with and without the demonstration on all 20 test episodes of each task. On Frame, the demonstration raises the number of successful episodes from 1/20 to 13/20 and the mean score from 0.283 to 0.800, because it supplies the two-arm division of labor that zero-shot execution does not find. On Gear, by contrast, both conditions reach 3/20 successes and a mean score of 0.200, although the failures differ. Without the demonstration, the agent leaves the gear outside the assembly region. With it, the agent positions the gear between the mounted gears and lowers it, yet 15 of the 20 episodes still end without a scored stage, so the mean score does not rise above the zero-shot level.

\paragraph{Adaptation and precision.} On two contrasting tasks, the direct-control agent adjusts its grasps and revisits disrupted goals, whereas the post-trained policies execute precise manipulation more accurately. Appendix~\ref{app:policy_rollouts} sets the agent beside $\pi_{0.5}$ and OpenWAM-$\alpha$ on Teacup \& teapot, where it leads, and on Glass into box, where it trails. On Teacup \& teapot, the agent targets the handle region and adjusts its wrist and gripper configuration across attempts. When later manipulation displaces the cup from the saucer, it revisits the cup and carries it back toward the saucer. In contrast, $\pi_{0.5}$ repeats approach and retraction motions without completing the cup transfer, and OpenWAM-$\alpha$ moves the teapot onto the tray while the cup stays off the saucer. On Glass into box, however, both post-trained policies fold the temples and OpenWAM-$\alpha$ also closes the lid, whereas the agent's corrective contacts leave a temple protruding as described in Section~\ref{sec:capability:failures}.

\subsection{Compositional Generalization}
\label{sec:capability:composition}

To evaluate whether the direct-control agent and post-trained policies can compose skills into a new multi-step sequence, we conduct a pilot experiment with an instruction outside the standard EBench task set: placing a bookmark and then a pen onto the same open book, as shown in Figure~\ref{fig:composition}. The EBench training set contains placing a bookmark onto a book and placing a pen into a holder or onto a ruler, but not placing a pen onto a book. This setting therefore tests more than the sequencing of two familiar tasks, as the combined objective includes a subtask absent from the robot-policy training data. Completing the first placement in the prescribed order awards a partial score of 0.50. The agent and both post-trained policies each run 10 rollouts, and the agent runs zero-shot without an in-context demonstration. It succeeds in all 10 episodes with a 1.00 mean score by placing the bookmark and then the pen onto the book in the required order. By contrast, $\pi_{0.5}$ and OpenWAM-$\alpha$ achieve 0/10 successes with mean scores of 0.40 and 0.35, respectively. In their rollouts, both post-trained policies either place the bookmark and stop without initiating the second step or directly manipulate the pen out of order. In this pilot, generalization to individual atomic skills does not yield multi-step compositional execution, whereas the agent's zero-shot instruction following completes the sequence.

\subsection{The Capability Boundary}
\label{sec:capability:boundary}

Taken together, the results of this section place the capability boundary of the direct-control agent at the transition from task-level reasoning to physical execution, and the way in which the agent acts explains why the boundary lies there. On the reasoning side, the semantic and spatial priors of multimodal pretraining supply the task-level knowledge, so that one annotated demonstration per task suffices to direct them. On the execution side, the agent acts through EEF waypoint chunks and observes the scene only after each chunk, so it lacks the continuous physical feedback that precise contact, coordinated two-arm motion, and long horizons require. The post-trained policies learn their control from demonstration trajectories and act on fresh observations throughout execution, so they receive this feedback by design.

\begin{tcolorbox}[enhanced,colback=astralight!45!white,colframe=astrablue!75!black,boxrule=0pt,leftrule=3pt,arc=1pt,
  left=8pt,right=8pt,top=6pt,bottom=6pt,before skip=8pt,after skip=8pt]
The two sides are complementary: the direct-control agent is stronger in \textbf{visual grounding, robustness across generalization dimensions, in-episode adaptation, and zero-shot skill composition}, and the post-trained policies in \textbf{closed-loop physical execution}. Section~\ref{sec:tool} therefore examines whether this boundary and its token cost shift in a harness with policy, where the MLLM core retains task-level reasoning and invokes a post-trained policy for closed-loop execution.
\end{tcolorbox}

\FloatBarrier
\section{Harness with Policy}
\label{sec:tool}
\suppressfloats[t]

This section tests whether combining the agent with post-trained policies can overcome the execution bottlenecks of direct control. We build the harness-with-policy setting following the Harness VLA framework~\citep{zhang2026harnessvla} and evaluate it across all 510 EBench test episodes with two representative policies: the VLA $\pi_{0.5}$~\citep{pi05} and the WAM InternW0-$\Delta$~\citep{miao2026internw0deltaworldactionmodel}. We first describe the harness design and compare the two harness configurations with direct control and the standalone policies in overall performance, across task groups, and across generalization dimensions. We then diagnose how failure patterns shift once the agent invokes a policy, analyze its policy invocation behavior, and finally examine the failure modes shared across both policies to identify the sources of the remaining gap. The final subsection also compares the token cost of each configuration.

\subsection{Harness Design}
\label{sec:tool:harness}
\label{sec:tool:design}

Following Harness VLA~\citep{zhang2026harnessvla}, our harness with policy exposes a frozen post-trained policy as a callable tool alongside the direct EEF interface described in Section~\ref{sec:setup:configs}, together with a call that returns the annotated task demonstration on request. In each invocation, the agent assigns 1--300 physics steps. Within these steps, the policy acts without intervention from the agent and then returns new visual and proprioceptive observations to the agent. Both harness configurations keep the system prompt, annotated task demonstration, and test scenes fixed and differ in the policy and its execution and history settings. The two policies contrast in temporal conditioning and in standalone strength. For the VLA $\pi_{0.5}$~\citep{pi05}, the harness uses the official EBench leaderboard checkpoint. This policy acts on the current observation alone, predicts 50 joint and gripper targets per inference, and replans every 30 physics steps. Evaluated alone, it trails the direct-control agent on both metrics in Table~\ref{tab:overall}. The WAM InternW0-$\Delta$~\citep{miao2026internw0deltaworldactionmodel} is likewise post-trained on EBench demonstrations but additionally conditions on two earlier frames of the episode. It predicts 32 joint, gripper, and base targets per inference and replans every 10 physics steps, which is three times as often as $\pi_{0.5}$, as detailed in Appendix~\ref{app:tool}. Alone, it exceeds direct control on both metrics. Pairing the same agent with both policies therefore asks \textbf{whether the agent still gains from a policy weaker than itself and whether a policy stronger than the agent lifts the combination beyond both components}.

\subsection{Overall Performance}
\label{sec:tool:overall}

\begin{table}[t]
  \centering
  \footnotesize
  \setlength{\tabcolsep}{4.2pt}
  \caption{Comparison of the \astra-based agent under direct control, each post-trained policy evaluated alone, and the harness with $\pi_{0.5}$ or InternW0-$\Delta$ across all 510 test episodes of EBench. Best and second-best values in each row are shown in bold and underlined.}
  \label{tab:tool}
  \begin{tabularx}{\linewidth}{l*{5}{>{\centering\arraybackslash}X}}
\toprule
\multirow{2}{*}{\hd{Tasks}} & \multirow{2}{*}{\hd{Direct control}} & \multicolumn{2}{c}{\hd{Policy alone}} & \multicolumn{2}{c}{\hd{Harness with}} \\
\cmidrule(lr){3-4}\cmidrule(lr){5-6}
 &  & \hd{$\pi_{0.5}$} & \hd{InternW0-$\Delta$} & \hd{$\pi_{0.5}$} & \hd{InternW0-$\Delta$} \\
\specialrule{\lightrulewidth}{\aboverulesep}{0pt}
\rowcolor{tabband} \multicolumn{6}{c}{\textit{\textbf{All tasks (26)}}} \\
\specialrule{\lightrulewidth}{0pt}{\belowrulesep}
SR (\%) & 46.7 & 41.4 & 49.2 & \second{53.0} & \best{56.0} \\
Score & 0.654 & 0.544 & 0.660 & \second{0.692} & \best{0.744} \\
\specialrule{\lightrulewidth}{\aboverulesep}{0pt}
\rowcolor{tabband} \multicolumn{6}{c}{\textit{\textbf{Precision}}} \\
\specialrule{\lightrulewidth}{0pt}{\belowrulesep}
Low (14) & 60.6 & 49.6 & 58.7 & \best{68.9} & \second{68.2} \\
Medium (8) & 40.2 & 39.0 & \second{49.2} & 47.9 & \best{56.5} \\
High (4) & 11.3 & \best{17.5} & \second{16.3} & 7.5 & 12.5 \\
\specialrule{\lightrulewidth}{\aboverulesep}{0pt}
\rowcolor{tabband} \multicolumn{6}{c}{\textit{\textbf{Mobility}}} \\
\specialrule{\lightrulewidth}{0pt}{\belowrulesep}
Mobile (19) & 56.6 & 45.9 & 55.0 & \second{61.0} & \best{64.3} \\
Tabletop (7) & 20.0 & 29.3 & \best{33.6} & \second{31.4} & \best{33.6} \\
\specialrule{\lightrulewidth}{\aboverulesep}{0pt}
\rowcolor{tabband} \multicolumn{6}{c}{\textit{\textbf{Horizon}}} \\
\specialrule{\lightrulewidth}{0pt}{\belowrulesep}
Short (11) & \second{73.2} & 52.3 & 59.1 & 70.5 & \best{77.3} \\
Long (15) & 27.3 & 33.4 & \best{42.0} & 40.2 & \second{40.4} \\
\bottomrule
\end{tabularx}

\end{table}

Each harness configuration outperforms both of its standalone components in Table~\ref{tab:tool}, though the gains are asymmetric: pairing with the stronger policy yields the higher overall performance, whereas the weaker policy gains more from the agent. Relative to the agent under direct control, overall SR rises by 6.3 percentage points to 53.0\% with $\pi_{0.5}$ and by 9.3 points to 56.0\% with InternW0-$\Delta$. The score rises from 0.654 to 0.692 and 0.744, respectively. OpenWAM-$\alpha$ is the strongest of the eight post-trained policies, and the harness with InternW0-$\Delta$ surpasses it on both metrics and ranks first in Table~\ref{tab:overall}. The harness with $\pi_{0.5}$ trails only OpenWAM-$\alpha$ among these eight. Relative to each policy alone, by contrast, SR rises by 11.6 points with $\pi_{0.5}$ but only by 6.8 points with InternW0-$\Delta$. Appendix Table~\ref{tab:tool_per_task} reports the per-task results.

\begin{figure}[t]
  \centering
  \includegraphics[width=\linewidth]{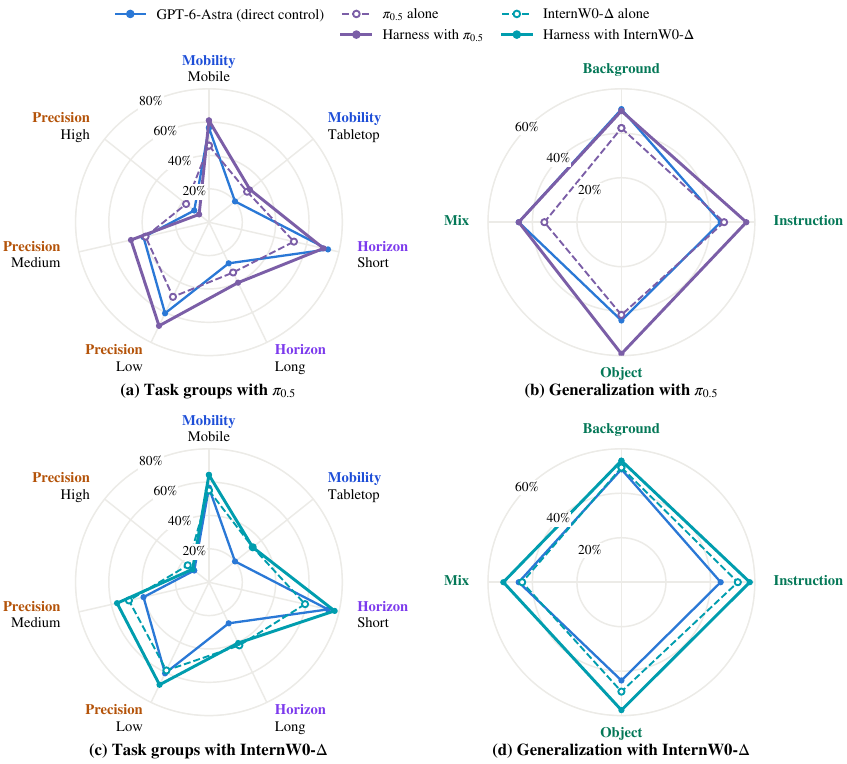}
  \caption{SR of the \astra-based agent under direct control, of each policy alone, and of the harness with that policy, for $\pi_{0.5}$ in (a, b) and InternW0-$\Delta$ in (c, d), across mobility, horizon, and precision task groups in (a, c) and across the four generalization dimensions of the EBench test set in (b, d). Panels in the same column share a radial axis that ends at the largest SR of the column, rounded up to a multiple of 10\%.}
  \label{fig:harness_radar}
\end{figure}

These net improvements combine resolved failures with newly failed episodes. Across the 510 paired episodes, the harness with $\pi_{0.5}$ succeeds on 91 episodes that fail under direct control and fails on 59 that succeed under it. The harness with InternW0-$\Delta$ gains 96 episodes and loses 49, as Appendix Table~\ref{tab:tool_transitions} details. Figure~\ref{fig:harness_radar} shows where this combination holds across task demands. On six of the seven task groups, each harness configuration comes within three percentage points of the stronger of its two components or exceeds it, whereas on high-precision tasks both configurations remain below the standalone policies. Section~\ref{sec:tool:groups} traces these group results to individual tasks and generalization dimensions, and Section~\ref{sec:tool:diagnosis} then analyzes the episode-level failure patterns behind both.

\subsection{Capability Profile Across Task Groups and Generalization Dimensions}
\label{sec:tool:groups}

\begin{figure*}[t]
  \centering
  \includegraphics[width=\linewidth]{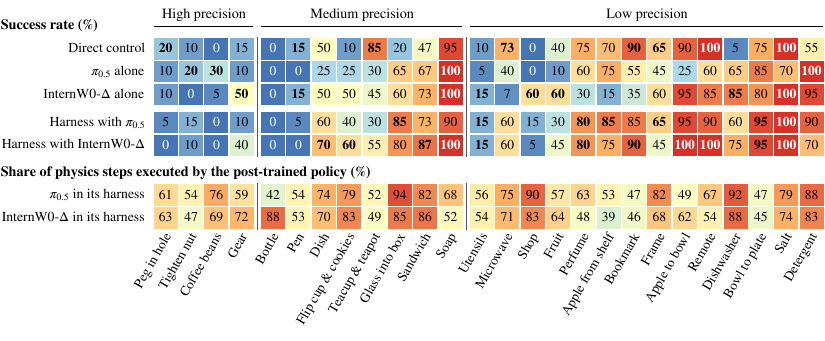}
  \caption{Per-task SR in percent across the 26 test tasks for the \astra-based agent under direct control, for each post-trained policy alone, and for the harness with $\pi_{0.5}$ and with InternW0-$\Delta$, together with the share in percent of physics steps that each post-trained policy executes in its harness configuration. Tasks are grouped by precision and ordered within each group by the mean SR of the two post-trained policies alone. One blue-to-red scale shades SR and step share alike on an absolute 0--100\% scale, values are rounded to whole percent, and bold marks the highest nonzero SR of each task. Table~\ref{tab:tool_per_task} lists the full task names.}
  \label{fig:tool_profile}
\end{figure*}

\paragraph{Task groups.} The harness with policy improves task success mainly on long-horizon and tabletop tasks, where the post-trained policy outperforms the agent under direct control, and it keeps the agent's lead on short-horizon tasks. With either policy, SR increases by roughly 13 percentage points on long-horizon tasks and by 11--14 percentage points on tabletop tasks in Table~\ref{tab:tool}. These are two of the three groups on which the agent ranks lowest under direct control in Section~\ref{sec:capability:profile}. On these groups, invoking the policy brings the agent to about the level of the policy alone. Specifically, the harness with InternW0-$\Delta$ matches InternW0-$\Delta$ alone on tabletop tasks at 33.6\% and stays close to it on long-horizon tasks at 40.4\% versus 42.0\%. The harness with $\pi_{0.5}$ exceeds $\pi_{0.5}$ alone on both groups, with 40.2\% versus 33.4\% on long-horizon tasks and 31.4\% versus 29.3\% on tabletop tasks. On short-horizon tasks, the agent already leads all compared post-trained policies under direct control. Both harness configurations preserve this strong performance and therefore outperform the corresponding policy alone by 18.2 percentage points. On Apple from shelf, for instance, InternW0-$\Delta$ alone achieves 15\% SR, whereas the harness with InternW0-$\Delta$ reaches 75\% SR. Appendix~\ref{app:stats} details the statistical significance of task-level improvements across all 26 tasks.

\paragraph{Task-level gains on bimanual and articulated stages.} At the task level, the two largest gains over direct control arise with either policy on Glass into box and Dishwasher, as shown in Figure~\ref{fig:tool_profile}. On Glass into box, SR increases from 20\% under direct control to 85\% with $\pi_{0.5}$ and 80\% with InternW0-$\Delta$. On Dishwasher, it increases from 5\% to 60\% and 75\%, respectively. On these two tasks, the invoked policy executes the bimanual transport, temple folding, lid closure, and rack manipulation. These include the stages where the agent under direct control ran out of steps or ended early, as Appendix Figure~\ref{fig:tool_rollouts}a,b illustrates for Glass into box.

\paragraph{Task-level shortfalls against either component.} Two sets of tasks depart from these gains. First, both harness configurations fall below direct control on four of the 11 tasks where neither policy alone exceeds the SR of direct control: Peg in hole, Pen, Microwave, and Teacup \& teapot. On Peg in hole, for example, SR drops from 20\% to 5\% with $\pi_{0.5}$ and 0\% with InternW0-$\Delta$, and on Pen from 15\% to 5\% and 0\%. Second, the harness with InternW0-$\Delta$ falls furthest below InternW0-$\Delta$ alone on Shop and Detergent, where SR drops from 60\% to 5\% and from 95\% to 70\%. On both tasks, InternW0-$\Delta$ executes more than 80\% of the physics steps in the harness, so the policy itself performs most of the motion. The shortfall arises from objects lost during the delegated segments or after the agent resumes control. Section~\ref{sec:tool:diagnosis} traces both sets to their failure patterns.

\paragraph{Generalization dimensions.} The choice of policy also determines whether the harness keeps the balanced generalization profile of the agent under direct control, as shown in Figure~\ref{fig:harness_radar}b and~d. With $\pi_{0.5}$, the harness stays above $\pi_{0.5}$ alone on all four dimensions. Over direct control, however, it improves only on Object, where SR rises from 44.2\% to 59.2\%, and on Instruction, where it rises from 44.6\% to 56.2\%. On Background and Mix, it remains at about the level of direct control. As a result, its range across dimensions widens from 6.6 to 13.0 points. The newly failed episodes explain this pattern. Resolved failures are distributed evenly at 21 to 25 episodes per dimension. Newly failed episodes, by contrast, concentrate on Background and Mix with 22 and 21 episodes, against only 7 on Object and 9 on Instruction. By failure cause, 25 of these 43 newly failed episodes begin when the agent's own grasp slips. By task, 21 of the 43 come from just four tasks. Teacup \& teapot leads them, as all five of its Background episodes newly fail. With InternW0-$\Delta$, newly failed episodes are fewer and spread more evenly at 9 to 17 per dimension, against 22 to 26 resolved failures on each. The harness therefore gains on every dimension and surpasses both of its components on each. Its Mix SR of 53.1\% and score of 0.718 are the highest of all compared policies and configurations, and its range narrows to 4.6 points, the smallest in Appendix Table~\ref{tab:generalization}.

\subsection{Failures Resolved and Introduced by the Policy}
\label{sec:tool:diagnosis}

\begin{table}[t]
  \centering
  \footnotesize
  \caption{Episode outcomes across the 510 test episodes under direct control and in the harness with $\pi_{0.5}$ or InternW0-$\Delta$, termination modes of unsuccessful episodes, and policy invocation statistics. A reported failure denotes an early termination in which the agent explicitly reports the task as incomplete. Recovery requests denote policy invocations whose stated reason is coded lexically as asking the policy to recover from a slip, missed grasp, or displaced object, as detailed in Appendix~\ref{app:tool}.}
  \label{tab:harness_outcomes}
  \begin{tabularx}{\linewidth}{l*{3}{>{\centering\arraybackslash}X}}
\toprule
 & \multirow{2}{*}{\hd{Direct control}} & \multicolumn{2}{c}{\hd{Harness with}} \\
\cmidrule(lr){3-4}
 &  & \hd{$\pi_{0.5}$} & \hd{InternW0-$\Delta$} \\
\specialrule{\lightrulewidth}{\aboverulesep}{0pt}
\rowcolor{tabband} \multicolumn{4}{c}{\textit{\textbf{Outcome of the 510 episodes}}} \\
\specialrule{\lightrulewidth}{0pt}{\belowrulesep}
Success & 237 & 269 & 284 \\
Partial credit & 188 & 162 & 168 \\
Score of zero & 85 & 79 & 58 \\
\specialrule{\lightrulewidth}{\aboverulesep}{0pt}
\rowcolor{tabband} \multicolumn{4}{c}{\textit{\textbf{Ending of the unsuccessful episodes}}} \\
\specialrule{\lightrulewidth}{0pt}{\belowrulesep}
Physics steps exhausted & 229 & 201 & 178 \\
Unconfirmed completion claim & 21 & 15 & 13 \\
Reported failure & 22 & 24 & 33 \\
Other & 1 & 1 & 2 \\
\specialrule{\lightrulewidth}{\aboverulesep}{0pt}
\rowcolor{tabband} \multicolumn{4}{c}{\textit{\textbf{Policy invocation}}} \\
\specialrule{\lightrulewidth}{0pt}{\belowrulesep}
Physics steps executed by the policy (\%) & -- & 66.9 & 65.4 \\
Policy invocations coded as recovery requests (\%) &  &  &  \\
\quad in successful episodes & -- & 6.3 & 5.8 \\
\quad in unsuccessful episodes & -- & 22.0 & 20.7 \\
\bottomrule
\end{tabularx}

\end{table}

Table~\ref{tab:harness_outcomes} follows how the 510 episodes end under direct control and in each harness configuration. Most unsuccessful episodes reach the physics-step limit in all three configurations, and nearly all of the rest end early with an unconfirmed completion claim or a reported failure of the agent. This section compares each ending between direct control and the harness to show which failures the policy resolves and which it introduces. Section~\ref{sec:tool:shared} then examines the episodes that fail with both policies.

\paragraph{Step-limit failures.} Invoking a policy mainly reduces failures at the step limit on long-horizon tasks but rarely resolves those on high-precision tasks, where both policies alone also remain weak. Relative to direct control, unsuccessful episodes that exhaust the physics-step limit decrease from 229 to 201 with $\pi_{0.5}$ and to 178 with InternW0-$\Delta$, and zero-score episodes decrease from 85 to 79 and 58, respectively. Of the 229 step-limit failures under direct control, 72 succeed with $\pi_{0.5}$ and 76 with InternW0-$\Delta$, and most of these conversions occur on long-horizon tasks. On high-precision tasks, only 2 and 8 of the 59 step-limit failures succeed, as detailed in Appendix Table~\ref{tab:tool_transitions}. In the newly successful episodes, the policy executes 78\% of the physics steps and typically completes the final stage itself, as it does in every newly successful episode of Dishwasher.

\paragraph{Unconfirmed completion claims.} Invoking a policy also reduces early terminations on an unconfirmed claim of completion from 21 under direct control to 15 with $\pi_{0.5}$ and 13 with InternW0-$\Delta$. Of these 21 episodes, 10 and 11 become full successes in Appendix Table~\ref{tab:tool_transitions}. The policy brings this gain through execution. In 7 of the 10 and 8 of the 11 converted episodes, the last motion is a policy segment that completes the task, and every converted Dishwasher episode ends this way. The remaining claims concentrate on two tasks. Dishwasher accounts for 6 of the 15 claims with $\pi_{0.5}$, and Gear accounts for 7 of the 13 with InternW0-$\Delta$, whereas direct control makes a single claim on Gear. In these Gear episodes, the agent claims completion while the gear rests unseated near the middle shaft.

\paragraph{Object losses in newly failed episodes.} Object losses, by contrast, end more episodes in the harness configurations. In a grip loss, an object slips from a gripper that remains commanded closed and the fingers close fully, as detected by the criteria of Appendix~\ref{app:events}. Early terminations in which the agent reports an unrecoverable failure rise from 22 under direct control to 24 with $\pi_{0.5}$ and 33 with InternW0-$\Delta$, and every one of them in either harness configuration names a lost object. Across the test set, 59 episodes that succeed under direct control newly fail in the harness with $\pi_{0.5}$, and 49 newly fail with InternW0-$\Delta$. A call-by-call reading of these episodes shows that the two configurations lose objects in different ways, as reported in Appendix Table~\ref{tab:tool_losses}. With $\pi_{0.5}$, 30 of the 59 begin with a slip of the agent's own grasp during a lift test or transport. In the Teacup \& teapot episode of Appendix Figure~\ref{fig:case_studies}c, for example, segments of $\pi_{0.5}$ end with an empty hand before the agent's own grasp of the teapot slips. With InternW0-$\Delta$, 30 of the 49 lose or misplace an object during a policy segment or just after control returns to the agent. Shop leaves little margin for such a loss, as InternW0-$\Delta$ alone needs 4{,}161 to 4{,}876 of the 5{,}000 physics steps in its successful episodes. In the harness, the agent delegates more than three quarters of the physics steps in 17 of the 19 unsuccessful episodes, and the delegated segments include failed bimanual handovers and their retries. All 19 episodes end at the step limit, and the harness completes a single Shop episode. Across the four high-precision tasks, none of their 6 and 7 newly failed episodes completes the final contact stage. In these episodes, pegs slip or are released before seating, the nut is released unseated, and gears are released off-axis or not fully seated.

\paragraph{Unmaintained grasps after policy segments.} Policy segments hold objects more securely than the agent's own commands. Within policy segments, grip losses occur 0.63 and 0.72 times per 1{,}000 held arm-steps with $\pi_{0.5}$ and InternW0-$\Delta$. Under the agent's commands, they occur 2.18 and 2.22 times overall and 0.87 and 1.00 times outside its grasp tests, as detailed in Appendix~\ref{app:tool:invocation}. The losses after a segment arise once control returns to the agent. The harness then holds each gripper at its measured finger width, and the agent receives fresh images together with these widths. Its next command, however, often operates one arm and leaves the holding arm unspecified instead of commanding it to stay closed. Counted per holding arm, such unmaintained grasps occur 338 times after a segment of $\pi_{0.5}$ and 316 times after one of InternW0-$\Delta$. A grip loss on that arm or a disruption reported in the agent's next stated reasons follows within two calls in 17\% and 19\% of these cases, respectively, compared with 6.5\% and 9.9\% when the agent leaves a holding arm unspecified after one of its own commands in the same harness configuration. Stratified by task, the odds of such a disruption after a policy segment are 2.95 and 1.29 times those after the agent's own commands, as detailed in Appendix~\ref{app:tool}. All four Detergent episodes that newly fail in the harness with InternW0-$\Delta$ are lost this way, as illustrated in Appendix~\ref{app:case_studies:irreversible}. The agent thus does not yet use the returned proprioceptive state to confirm and maintain a grasp that the policy has established.

\subsection{Policy Invocation Behavior}
\label{sec:tool:delegation}

\begin{figure}[t]
  \centering
  \includegraphics[width=\linewidth]{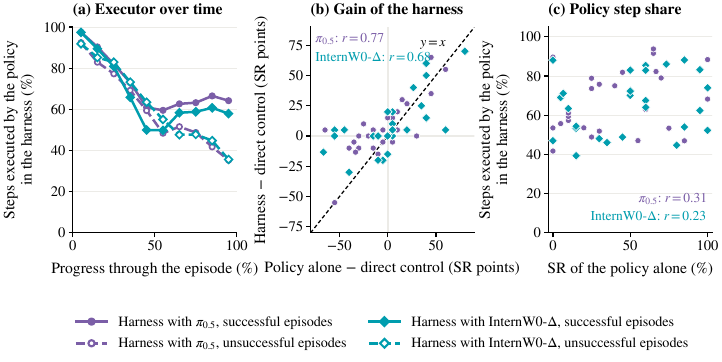}
  \caption{Policy invocation in the two harness configurations. (a) Share of the physics steps executed by the policy in each tenth of an episode, averaged over the successful and over the unsuccessful episodes. (b) Per task, the change in SR from direct control to the harness with policy against the SR difference between the policy alone and direct control, with the identity line. (c) Per task, the share of physics steps executed by the policy against the SR of the policy alone. Each marker in (b) and (c) is one of the 26 tasks, and $r$ is the Pearson correlation over tasks.}
  \label{fig:harness_dynamics}
\end{figure}

\paragraph{Delegation over an episode.} The agent delegates first and takes control back as the episode proceeds. Successful episodes then return control to the policy, whereas unsuccessful ones stay largely under the agent's own commands. In all 510 episodes with either policy, its first motion command is a policy invocation, and the policy executes 92--97\% of the physics steps in the first tenth of an episode, as shown in Figure~\ref{fig:harness_dynamics}a. This share declines over the first half as the agent steps in with grasp tests, corrections, and recoveries. In the second half of successful episodes, the policy's share recovers to 58--64\% in the final tenth. Within the same task, successful episodes also leave about 14 points more of their steps to $\pi_{0.5}$ than unsuccessful ones, and about 13 points more to InternW0-$\Delta$. In unsuccessful episodes, by contrast, the policy's share keeps falling to 35--36\%, and 68\% and 72\% of these episodes end with a motion command of the agent. Unsuccessful episodes are also longer and contain more transitions between the agent and the policy. They average 7.2 transitions against 4.2 in successful episodes with $\pi_{0.5}$ and 6.8 against 3.8 with InternW0-$\Delta$. Between transitions, however, the agent keeps control for longer stretches, with 0.26 and 0.24 transitions per motion command against 0.33 and 0.28 in successful episodes, as listed in Appendix Table~\ref{tab:invocation_stats}. The policy invocations of unsuccessful episodes also ask for recovery more often. With $\pi_{0.5}$ and InternW0-$\Delta$, 22.0\% and 20.7\% of these invocations are coded as asking the policy to recover a slipped, missed, or displaced object. These shares are more than three times the 6.3\% and 5.8\% in successful episodes in Table~\ref{tab:harness_outcomes}. Recovery requests thus mark the least successful episodes. Among the episodes with a reported disruption, those with a recovery request fail in 79.3\% and 79.0\% of cases, against 58.3\% and 53.8\% of those without one. The agent delegates 35\% and 34\% of the motion commands whose stated reasons report a slip, a lost object, or another disruption, and it executes the rest itself.

\paragraph{Competence and delegation across tasks.} Across tasks, the gain of the harness follows the competence of the policy, but the agent's delegation does not. On the one hand, the change in SR from direct control tracks the advantage of the policy alone over the agent under direct control, with Pearson correlations of $r=0.77$ for $\pi_{0.5}$ and $r=0.68$ for InternW0-$\Delta$, as shown in Figure~\ref{fig:harness_dynamics}b. On the other hand, the policy executes on average 66.9\% of the physics steps of a task with $\pi_{0.5}$ and 65.4\% with InternW0-$\Delta$. This share is not significantly correlated with the SR of the policy alone, with $r=0.31$ and $r=0.23$ in Figure~\ref{fig:harness_dynamics}c. For example, $\pi_{0.5}$ alone succeeds in at most 10\% of episodes on seven tasks, yet the agent still leaves it 60\% of the physics steps on these tasks on average and 42--90\% per task. The share is 71\% on the 11 tasks on which $\pi_{0.5}$ alone succeeds in at least half of the episodes. By the same thresholds, InternW0-$\Delta$ is weak on five tasks and strong on 15, and the agent leaves it 68\% and 71\% of the physics steps on them. The share differs instead by horizon. With $\pi_{0.5}$ and InternW0-$\Delta$, it is 58\% and 54\% on short-horizon tasks against 73\% and 74\% on long-horizon tasks.

\paragraph{Subgoal delegation under a fixed task instruction.}
\label{sec:tool:cases}
The agent's stated reasons show that it divides the work with the policy by subgoal, whereas the policy pursues the whole task. In a successful Flip cup \& cookies episode, for instance, the agent tests a grasp with its own waypoints under the stated reason ``Test cup grasp by raising 1 cm without rotation''. It then leaves the handover, the reorientation of the cup, and the pour to InternW0-$\Delta$, as listed in Appendix~\ref{app:case_studies:division}. In a Gear episode with $\pi_{0.5}$, the agent assigns each invocation a stage, from carrying the gear to the middle mounting position to its fine alignment and seating. At times it also assigns a verification such as ``Verify the released blue gear remains seated while withdrawing the hand''. The policy, however, is conditioned on the task instruction instead of that stage and reports nothing back. The agent then claims completion after checking only the overview image while the gear rests on the board unseated, as detailed in Appendix~\ref{app:case_studies:completion}. These episodes show that the agent's subgoal does not reach the policy upon invocation.

\subsection{Failure Modes Shared Across Policies}
\label{sec:tool:shared}

The two harness configurations fail largely on the same tasks. Of the 510 test episodes, 164 fail with both policies. This count is close to the 168 expected if the two configurations failed independently within each task. On the four high-precision tasks, 65 of the 80 episodes fail with both. Shared failures are therefore governed by task-level difficulty rather than by scene-specific factors.

Within these shared failures, however, where an episode breaks along its chain of grasping, transfer, and contact depends on the policy. The two runs of an episode end with the same category of decisive incident in only 58 of the 164 episodes, as broken down in Appendix Table~\ref{tab:tool_shared}. First, object losses decide 109 shared failures with $\pi_{0.5}$ and 105 with InternW0-$\Delta$, and their location shifts with the policy. With $\pi_{0.5}$, 58 of these losses occur during the agent's own lift tests and transport and 37 inside policy segments. With InternW0-$\Delta$, the order reverses, as 53 occur inside policy segments and 42 during the agent's own lift tests and transport. The remaining 14 and 10 follow a grasp left unmaintained after a policy segment, as analyzed in Section~\ref{sec:tool:diagnosis}. Second, 19 and 24 shared failures stall at the final insertion, threading, or seating stage, and 16 and 15 spend policy segments empty-handed or on the wrong object. Third, 19 and 20 shared failures end without a single decisive object loss. Of these, 16 and 11 reach the step limit as recovered slips, empty grasps, and displaced objects accumulate. Bottle accounts for 14 of these 27 step-limit runs, and in 11 of the 14, bottles slip, fall beside or behind the basket, or lie outside it. No run on Bottle places more than 9 of the 12 bottles. The other 3 and 9 end with a false completion claim.

\subsection{Shifted Boundary, Remaining Gaps, and Token Cost}
\label{sec:tool:summary}

The harness with policy moves the boundary of Section~\ref{sec:capability:boundary} where the policy supplies the closed-loop execution that direct control lacks, chiefly on the long-horizon and tabletop tasks. Two limits remain, and a final paragraph examines how the shift changes the agent's token cost.

\paragraph{Shared ceiling on high-precision contact.} First, the agent and both policies share an execution ceiling on high-precision contact, so combining them does not lift it. A per-task choice in hindsight between the agent alone and the policy alone reaches only 21.3\% SR on the high-precision tasks with either policy. The best of all three components reaches 30.0\%, which remains below the 32.5\% of OpenWAM-$\alpha$.

\paragraph{Gap to the better component from loose coordination.} Second, the combination loses part of what its components achieve alone. The per-task choice of the better component would attain 56.2\% with $\pi_{0.5}$ and 60.8\% with InternW0-$\Delta$. It would thus exceed the respective harness configuration by 3.1 and 4.8 points over all tasks and by 13.8 and 8.8 points on the high-precision tasks. The harness with $\pi_{0.5}$ exceeds both of its components on only ten tasks and falls below the better one on thirteen. The harness with InternW0-$\Delta$ exceeds both on eight tasks and falls below the better one on eleven. The preceding analyses trace this loss to three sources in how agent and policy coordinate. First, the agent allocates physics steps to the policy by task horizon rather than by the policy's competence on the task, as shown in Section~\ref{sec:tool:delegation}. Second, each policy segment conditions on the task instruction and the observations listed in Section~\ref{sec:tool:design} rather than on the agent's subgoal or the stages already completed, as shown in Section~\ref{sec:tool:cases}. Third, once a policy segment ends, the agent does not command the holding arm to stay closed at the returned finger width, as shown in Section~\ref{sec:tool:diagnosis}.

\begin{figure}[t]
  \centering
  \includegraphics[width=0.92\linewidth]{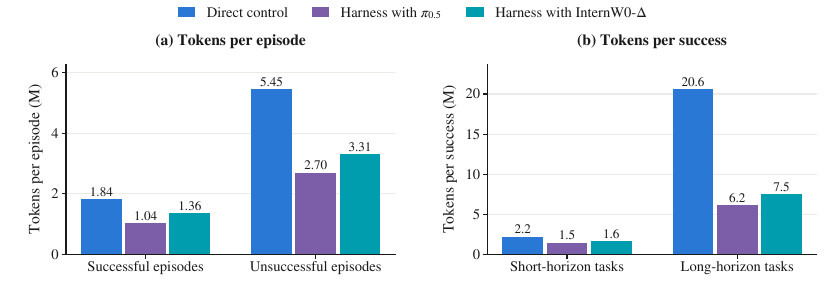}
  \caption{Token use of the \astra-based agent under direct control and in the harness with each policy on the 510 EBench test episodes. (a) Mean tokens per episode in successful and unsuccessful episodes. (b) Tokens per success on short-horizon and long-horizon tasks, dividing all tokens of each group, including those of its unsuccessful episodes, by its number of successful episodes.}
  \label{fig:token_cost}
\end{figure}

\paragraph{Token cost.} Invoking a policy also moves computation from the agent into the policy. The agent consumes 1.83M tokens per episode with $\pi_{0.5}$ and 2.22M with InternW0-$\Delta$, which is 52\% and 41\% less than the 3.77M under direct control. It also makes 23.7 and 22.7 tool calls per episode instead of 42.6. Per success, it needs 3.5M and 4.0M tokens, which is 57\% and 51\% fewer than the 8.1M under direct control. The savings concentrate where direct control is most expensive, as shown in Figure~\ref{fig:token_cost}. On long-horizon tasks, one success costs 6.2M and 7.5M tokens in the two harness configurations, which is less than two fifths of its cost under direct control. On short-horizon tasks, the cost per success falls by only a quarter to a third. Within each configuration, unsuccessful episodes still consume 2.6 and 2.4 times the tokens of successful ones, because tokens grow with the number of calls in the harness as well. At the published tariff of \astra, the agent's tokens cost USD 16.82 per success under direct control but only 4.99 and 5.49 in the two harness configurations, as detailed in Appendix~\ref{app:tool:cost}. In exchange, the policy adds 45 inferences of $\pi_{0.5}$ and 135 of InternW0-$\Delta$ per episode.

\begin{tcolorbox}[enhanced,colback=astralight!45!white,colframe=astrablue!75!black,boxrule=0pt,leftrule=3pt,arc=1pt,
  left=8pt,right=8pt,top=6pt,bottom=6pt,before skip=8pt,after skip=8pt]
With a post-trained policy, the agent reaches further on \textbf{long-horizon and tabletop tasks at a lower token cost per success}, but it remains bounded by \textbf{high-precision contact, which none of its components masters}, and by \textbf{loose coordination in delegation, subgoals, and control transfer}. Section~\ref{sec:conclusion} turns these limits into directions for future work.
\end{tcolorbox}

\FloatBarrier
\section{Conclusions and Future Directions}
\label{sec:conclusion}

This report profiles \astra on EBench as the core of an agent that controls a robot through an agent harness. The agent observes the scene under the same visual and proprioceptive protocol as post-trained policies and acts first under direct control and then in a harness with a post-trained VLA or WAM. Under direct control, the capability boundary of the agent lies between flexible task interpretation and reliable physical execution. Without robot-specific post-training of the MLLM, the direct-control agent draws on multimodal pretraining and one annotated demonstration for each task to compete with post-trained policies overall, with particular strengths in visual grounding, robustness to generalization shifts, adaptation to failures during an episode, and zero-shot composition of skills. Physical execution, by contrast, exposes a tension in how the agent controls the robot. The agent chooses its corrections between waypoint chunks, whereas grasps, contacts, and the alignment of parts evolve within each chunk, so precise contact, bimanual coordination, and long horizons limit its success. Because each correction also resends the accumulated context, the unsuccessful and long-horizon episodes that this tension produces account for most of the agent's token cost. The same unconstrained execution also drops held objects outside the agent's deliberate grasp tests and at times produces abrupt wrist rotations that risk damaging physical hardware.

Invoking a post-trained policy moves part of this boundary. With InternW0-$\Delta$, the harness with policy reaches the highest SR and score in Table~\ref{tab:overall}, and with either policy the agent needs far fewer tokens per success on long-horizon tasks, where direct control is costliest. However, two shortfalls of the combination indicate where further progress is needed. First, precise contact limits the agent and each policy alike, so combining them leaves this ceiling in place. Second, the combination falls short of its better component on many tasks, because the coordination between agent and policy remains loose. The agent delegates without regard to the policy's competence, the policy pursues the task instruction rather than the agent's subgoal, and when control returns from the policy, the agent often leaves the holding arm unmaintained, allowing grasped objects to slip. These limitations motivate us to explore several directions.

\paragraph{Designing subtask interfaces and training policies as learnable tools.} In the evaluated harness, the agent passes the policy only a step count, so it decides when the policy acts but can neither assign a subgoal nor receive a report of its completion. A finer interface would not only expose learned skills as tools that a planner selects~\citep{rpent} but also pass an explicit subgoal to the policy, verify the entry and exit conditions of each subtask~\citep{baton}, and return whether the subgoal was reached. Such an interface also requires a policy that can serve it. The evaluated policies are post-trained on complete EBench task trajectories under the task instruction, whereas the harness invokes them mid-episode from states that the agent has reached and for subgoals that the agent has chosen. The policy itself therefore has to be trained as a learnable tool of the harness. This training spans three levels and can build on the subtask instruction following of $\pi_{0.5}$~\citep{pi05} and on the subtask descriptions and execution feedback that the authors of InternW0-$\Delta$ list as future extensions of its interface~\citep{miao2026internw0deltaworldactionmodel}. First, the policy should follow the agent's subgoals rather than only the task instruction. Second, it should be trained on the mid-episode states in which the agent transfers control, including disturbed and displaced ones. Third, it should end each segment on subgoal completion or failure, returning structured feedback to the agent and leaving any held object in a stable grasp. The agent, in turn, can be trained to delegate a subtask according to the policy's competence on it and to use the proprioceptive state returned at every control transfer to keep a held object grasped.

\paragraph{Grounding exploratory experience into reusable memory and tools.} Within an episode, the agent diagnoses slips and displaced objects and revises its strategy, yet every new episode starts again from the same static prompt. Recent work turns such experience into reusable assets. Language-model programs compose perception and control APIs into robot behavior~\citep{codeaspolicies}, and recent harnesses synthesize task programs around post-trained policies~\citep{wang2026harnesspai}, expose embodiment-specific primitives~\citep{chen2026showharness}, or extend their control interfaces during deployment~\citep{li2026racap}. To admit only experience that generalizes, recent work also validates candidate skills on held-out episodes~\citep{lu2026aspire} and isolates the evaluation from the agent that revises its own prompt and tools~\citep{elmaaroufi2026rho}. For embodied agents, two questions follow: which recoveries generalize beyond a single trial, and whether experience accumulated with one MLLM serves other MLLMs as well. The stated reasons of the agent offer a natural starting point, as each names a diagnosed failure together with the revised action that follows it. Validated pairs of failure and revision can be retained across episodes as memory or compiled into tools that wrap closed-loop control.

\paragraph{Transferring the agent's exploratory and corrective strategies to smaller policies.} The recorded rollouts contain corrective strategies such as withdrawing for a wider view after a slip and lowering a sweep that missed the beans. The agent produces these strategies at the latency and token cost of a frontier MLLM. Distilling them into smaller models that run at deployment speed would pair failure states with the agent's revised strategies and their outcomes, including corrections that did not succeed. Such distillation would thereby extend behavior cloning from trajectories toward the intentions behind them. Policies already improve from their own experience and from corrective interventions~\citep{recap}, and an LLM agent can run the cycle of training and evaluating a VLA without a human in the loop~\citep{yu2026agenticrobotics}. Because agent rollouts carry the reason for every correction, they extend such training data to corrections that the agent discovers on its own.

\paragraph{Coupling geometric perception with active safety constraints in execution.} Precise contact and safe execution both depend on geometry that the agent infers from RGB views without depth. On insertion, threading, and seating, the agent has to judge the height of a surface, the clearance of a part, and the onset of contact from images, and its commands reach the robot without checks against the table, the workspace bounds, or the speed of the motion. The harness can therefore pair perception tools with a safety layer. Perception tools such as depth estimation, open-vocabulary detection and segmentation, and visual checks of the terminal state can estimate these quantities from the same camera views. The safety layer can check each command against workspace, reachability, and velocity limits and interrupt chunk execution on slip or collision. On real hardware, a queryable model of the robot's body already lets an agent convert functional targets into control targets~\citep{lou2026knowbody}, and such a model can supply the kinematic limits that this safety layer enforces. Reporting safety events beside SR and score would let benchmarks measure this progress.

\paragraph{Limitations.} This report mainly studies the capability profile of one frontier MLLM, \astra, as the core of an agent on one simulated benchmark with synchronous execution, both under direct control and in a harness with two post-trained policies. Natural extensions include measuring the agent's execution time, asynchronous execution in which the scene keeps evolving while the model deliberates, self-improvement of the agent with or without training the post-trained policy, a full evaluation of other frontier MLLMs as agent cores under the same protocol, and deployment on a physical robot. Appendix~\ref{app:caveats} lists the protocol of each run.

\bibliographystyle{iclr2027_conference}
\bibliography{main}

\clearpage
\appendix
\addtocontents{toc}{\protect\setcounter{tocdepth}{2}}
\renewcommand{\contentsname}{Appendix}
{\setlength{\parskip}{1pt}\tableofcontents}
\clearpage
\makeatletter
\setlength{\@fptop}{0pt}
\setlength{\@fpsep}{12pt}
\setlength{\@fpbot}{0pt plus 1fil}
\makeatother
\setlength{\floatsep}{12pt}
\setlength{\textfloatsep}{12pt}
\setlength{\intextsep}{10pt}
\setcounter{topnumber}{2}
\setcounter{bottomnumber}{1}
\setcounter{totalnumber}{3}
\renewcommand{\topfraction}{0.9}
\renewcommand{\bottomfraction}{0.6}
\renewcommand{\textfraction}{0.08}
\renewcommand{\floatpagefraction}{0.95}
\setcounter{section}{0}
\renewcommand{\thesection}{\Alph{section}}
\renewcommand{\thesubsection}{\thesection.\arabic{subsection}}
\renewcommand{\thesubsubsection}{\thesubsection.\arabic{subsubsection}}
\renewcommand{\thefigure}{\thesection.\arabic{figure}}
\renewcommand{\thetable}{\thesection.\arabic{table}}
\newcommand{\inputappendixsection}[1]{%
  \FloatBarrier
  \setcounter{figure}{0}%
  \setcounter{table}{0}%
  \input{#1}%
}

  \FloatBarrier
  \setcounter{figure}{0}%
  \setcounter{table}{0}%
  \section{Agent Interface and Prompts}
\label{app:prompts}

This appendix reproduces the inputs and outputs of the agent from the stored records of the three configurations, and the system instructions and tool declarations from the code that each run executed. Omissions are marked [\ldots], other bracketed text replaces images, identifiers, and file paths or labels an excerpt, non-ASCII characters are transliterated, and line breaks within a paragraph result from typesetting.

\subsection{Agent instructions and task prompt}

\paragraph{System instructions.} Every agent session of the direct-control evaluation begins with the following system instructions, followed by instructions of the model host on returning images from tool calls. In both harness configurations, the list of prohibited actions ends with ``or delegate to other agents'', and the \texttt{query\_icl} paragraph that closes the task prompt below is inserted between the two parts.

\begin{Verbatim}
You control one simulated EBench robot episode. Use only the supplied RGB and measured robot state and the declared robot tools. Do not read external files, use shell, browse, or delegate. The server judges success independently. Stop when it reports a terminal result or a budget is exhausted. Repeated observe at the same timestep shows the same scene. A closed gripper or finger contact alone does not prove a secure grasp. Choose continuous poses, motion timing and observation cadence yourself.

[... instructions of the model host on returning images from tool calls ...]
\end{Verbatim}

\paragraph{Task prompt.} The first user turn consists of the task prompt, the demonstration (Appendix~\ref{app:prompts-demo}), and the current observation. The excerpt below is the first turn of an episode of the harness with $\pi_{0.5}$ on the bowl-to-plate test task. Under direct control, the agent received the same turn without the paragraph on \texttt{execute\_pi05} and the paragraph on \texttt{query\_icl}.

\begin{Verbatim}
Complete this single task yourself using the tools: Take out one of the three bowls and stack it on the plate.
Robot: manip/lift2/R5a, two arms and a movable planar base. Left arm is left in the overlook view. EEF poses are local to each arm's planner base: +X forward, +Y robot-left, +Z up; quaternion order wxyz. [...] There is no supplied depth, object pose or camera calibration. Only RGB and measured robot EEF/gripper/base state are available. [...] Use execute_eef sequentially, not parallel calls. Use observations to judge motion and object manipulation. The server determines task success independently. Do not start another episode. Resource limits: unlimited tool calls, physics deadline enforced by the server; no hard token limit; [...] When finished, state the observed outcome concisely.
You may call execute_pi05 to delegate a manipulation segment to a learned policy. Choose when to delegate using the current scene. It replans every 30 steps and automatically returns control after the requested segment (at most 300 steps). The pi0.5 prompt is fixed for this task; always pass instruction=null and never invent a per-call prompt. Inspect its returned RGB/state and take over with EEF waypoints or delegate again. [...] The learned policy uses the same current three RGB cameras plus joint/gripper/base state; it receives no object ground truth or demonstration images.
Manipulation guidance: after visually aligning, close at a stationary pose and allow settling. A nonzero finger gap on the table is contact evidence, not proof of a secure grasp. Test each grasp by lifting only 0.01 m first, using motion_steps=16 or more, then hold 8..16 steps and request the active wrist plus overlook views. [...]
RGB inputs are downsampled: overlook 640x360, wrist 384x288; no crop or object annotation.
The saved ICL demonstration remains available through query_icl throughout this episode. [...] Ground actions in the current RGB/state.
[... the demonstration: annotation text and ten captioned keyframe images ...]
END OF HISTORICAL DEMONSTRATION. The following images and measured state are the CURRENT episode. [...]
Measured robot state (EEF poses in each arm's planner base): {"state.gripper": [0.044, 0.044, 0.044, 0.044], "state.base": [0.0, 0.0, -0.0], "state.ee_pose": [[[0.1002, 0.0, 0.16349], [1.0, 0.0, 2e-05, -0.0]], [[0.1002, 0.0, 0.16349], [1.0, -0.0, 2e-05, -0.0]]], "timestep": 0}
CURRENT RGB: overlook, 640x360
[image: overview camera]
[... the left and right wrist views in the same format ...]
\end{Verbatim}

\noindent The harness with InternW0-$\Delta$ received the same turn with the following paragraph in place of the one on \texttt{execute\_pi05}.

\begin{Verbatim}
You may call execute_wam to delegate a manipulation segment to a learned policy. Choose when to delegate using the current scene. It replans every 10 steps and automatically returns control after the requested segment (at most 300 steps). The WAM prompt is fixed to this task's instruction; always pass instruction=null and never invent a per-call prompt. [...] The learned policy uses the same current three RGB cameras plus robot state, and its own frames from earlier in this episode; it receives no object ground truth or demonstration images.
\end{Verbatim}

\subsection{Tool interface}

Table~\ref{tab:tools} lists the agent's tools with their argument schemas and their descriptions as declared to the model. Under direct control, the agent has the first three tools. The harness with $\pi_{0.5}$ adds \texttt{execute\_pi05} and \texttt{query\_icl}, the harness with InternW0-$\Delta$ adds \texttt{execute\_wam} and \texttt{query\_icl}, and the task prompt of each harness configuration adds the paragraphs on its two tools shown above. The declarations of the EEF tools and of \texttt{query\_icl} are identical in every configuration that has them, and no other tool is declared.

\begin{table}[htbp]
\caption{Agent tools, with descriptions reproduced from the tool declarations (omissions marked [\ldots]). All tools except \texttt{query\_icl} also take \texttt{cameras} (one to three of \texttt{overlook}, \texttt{left\_wrist}, and \texttt{right\_wrist}), and \texttt{execute\_eef}, \texttt{execute\_pi05}, and \texttt{execute\_wam} a free-text \texttt{reason}. Availability: \emph{all} denotes every configuration, \emph{VLA} the harness with $\pi_{0.5}$, \emph{WAM} the harness with InternW0-$\Delta$, and \emph{demo} both harness configurations, in which the demonstration of the first turn also remains available on request.}
\label{tab:tools}
\centering
\scriptsize
\renewcommand{\arraystretch}{1.25}
\setlength{\tabcolsep}{3pt}
\begin{tabularx}{\linewidth}{@{}>{\raggedright\arraybackslash}p{1.8cm}>{\raggedright\arraybackslash}p{2.5cm}>{\raggedright\arraybackslash}p{1.2cm}>{\raggedright\arraybackslash}X@{}}
\toprule
\hd{Tool} & \hd{Arguments} & \hd{Available} & \hd{Description as shown to the model} \\
\midrule
\texttt{execute\_eef} & \texttt{waypoints}: 1--6 of \{\texttt{left}, \texttt{right} (position, quaternion, and two finger targets, or null), base pose or null, \texttt{steps} 1--64, and optional \texttt{motion\_steps}\} & all & ``Execute 1..6 native GenManip absolute EEF waypoints in order. Each arm target has xyz(m), unit quaternion(wxyz) in that arm's planner-base frame, and two finger joint targets(m): [0.044,0.044] open, [0,0] close. [\ldots] The simulator performs IK; targets may remain unreachable, so inspect returned measured poses. No automatic grasp approval or task-stage rules are imposed. Returns final robot state and selected RGB cameras; sequences have no intermediate model observation. [\ldots]'' \\
\addlinespace
\texttt{set\_gripper} & \texttt{side}, \texttt{opening} in [0,1], \texttt{steps} 1--64 & all & ``Send a normalized gripper opening command: opening=0 fully closed, opening=1 fully open. [\ldots] Use this for closing at rest; measured opening may remain nonzero due to contact. Returns selected RGB and measured robot state.'' \\
\addlinespace
\texttt{observe} & none & all & ``Return selected RGB views and measured robot state from the latest acknowledged simulator step. Does not advance physics. [\ldots]'' \\
\addlinespace
\texttt{execute\_pi05} & \texttt{steps} 1--300, \texttt{instruction} (string or null) & VLA & ``Delegate a bounded manipulation segment to the trained EBench pi0.5 policy. It uses the CURRENT RGB images and robot state, predicts absolute joint/gripper targets, and replans every 30 steps. Choose this when learned dexterity may help with grasping, alignment, insertion or other fine manipulation. [\ldots] After the requested steps, control automatically returns to you with fresh RGB and state; [\ldots] pass instruction=null. [\ldots]'' \\
\addlinespace
\texttt{execute\_wam} & \texttt{steps} 1--300, \texttt{instruction} (string or null) & WAM & ``Delegate a bounded manipulation segment to the trained EBench InternW0-delta world-action-model policy. It uses the CURRENT RGB images and robot state together with its own frames from earlier in this episode, predicts absolute joint/gripper targets, and replans every 10 steps. [\ldots] After the requested steps, control automatically returns to you with fresh RGB and state; [\ldots] pass instruction=null. [\ldots]'' \\
\addlinespace
\texttt{query\_icl} & \texttt{view} (text, frames, or all), \texttt{frame\_indices} (keyframe indices) & demo & ``The saved ICL demonstration remains available through query\_icl throughout this episode. [\ldots] Querying does not advance simulation. Ground actions in the current RGB/state.'' \\
\bottomrule
\end{tabularx}
\end{table}

\FloatBarrier
\subsection{Demonstration format}
\label{app:prompts-demo}\label{app:demos}

The demonstration enters the first turn as one text block followed by alternating keyframe captions and images, and the three configurations received the same demonstration. Table~\ref{tab:demo_format} lists the parts of this text block for the bowl-to-plate task: the annotation of the source episode, a fixed paragraph on the interpretation of the records, and one numeric record per keyframe. Each record holds the measured EEF poses of the left arm \texttt{ee0} and the right arm \texttt{ee1} as position and wxyz quaternion, the summed finger gaps, and the base triplet, together with the exported action target.

\begin{table}[htbp]
\caption{Parts of the annotated bowl-to-plate demonstration in the first turn, with text reproduced from the prompt and omissions marked [\ldots]. The numeric record of one keyframe is shown with values rounded to three decimals.}
\label{tab:demo_format}
\centering
\scriptsize
\renewcommand{\arraystretch}{1.3}
\setlength{\tabcolsep}{4pt}
\begin{tabularx}{\linewidth}{@{}>{\raggedright\arraybackslash}p{2.4cm}>{\raggedright\arraybackslash}X@{}}
\toprule
\hd{Part} & \hd{Content as shown to the model} \\
\midrule
Header & ``HISTORICAL DEMONSTRATION -- reference from a separate recorded episode, not the current observation or the current episode's action history.'' \\
Source and instruction & ``Historical demonstration of bowl\_to\_plate, source episode 600. Reference instruction: Take out one of the three bowls and stack on the plate.'' \\
Arm roles & ``Right arm; base shifts from the sink toward the plate.'' \\
Observed sequence & ``Reach into the sink and grip the rim of the brown bowl. Lift it clear of the other bowls and sink rim, reposition the base toward the plate on the counter, place the bowl over the plate, lower it onto the plate, and open the fingers.'' \\
Observed outcome & ``The brown bowl is visibly supported on the white plate after the right fingers open; two patterned bowls remain in the sink.'' \\
Evidence limits & ``Final contact is partly occluded by the gripper. No official success label or force measurement is supplied.'' \\
Fixed guidance & ``This separate recorded episode is not the current observation or action history. [\ldots] Transfer the sequence and relative geometry, ground commands in live RGB and robot state, and obey the live task instruction.'' \\
\addlinespace
Numeric record, one per keyframe & \texttt{frame} 188, \texttt{source\_timestamp\_s} 12.533 \newline
\texttt{measured}: \texttt{ee0\_xyz\_wxyz} [0.100, 0.000, 0.163, 1.000, 0.000, 0.000, 0.000], \texttt{ee1\_xyz\_wxyz} [0.384, 0.015, 0.046, 0.701, 0.000, 0.713, 0.000], \texttt{finger\_gaps\_m\_ee0\_ee1} [0.088, 0.015], \texttt{recorded\_base\_triplet} [0.000, 0.152, 0.000] \newline
\texttt{exported\_action}: \texttt{ee0\_xyz\_wxyz} as measured, \texttt{ee1\_xyz\_wxyz} [0.010, 0.040, 0.204, 0.735, 0.246, $-$0.259, 0.576], \texttt{finger\_gaps\_m\_ee0\_ee1} [0.088, 0.000], \texttt{recorded\_base\_triplet} as measured \\
\addlinespace
Keyframe caption and image, one per keyframe & ``Historical demonstration: Historical episode 600; frame 188; 12.533s; overlook camera. Grasp and lift the brown bowl.'' followed by the keyframe image, as shown for all ten keyframes in Figure~\ref{fig:demo_keyframes}. \\
\bottomrule
\end{tabularx}
\end{table}

\begin{figure}[htbp]
  \centering
  \includegraphics[width=\linewidth]{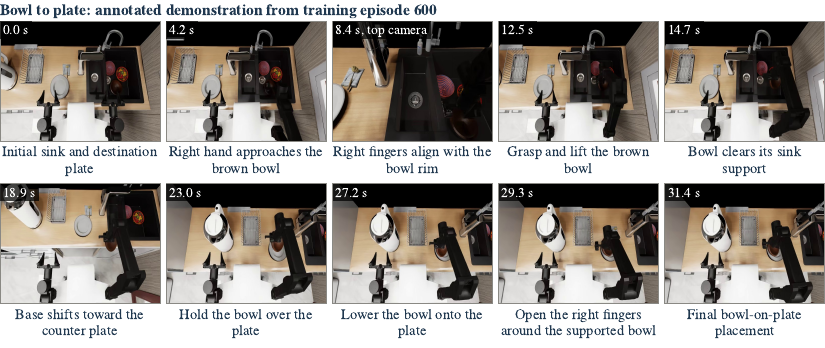}
  \caption{The ten keyframes of the annotated bowl-to-plate demonstration (training episode 600) with their captions, as they enter the first turn. Each frame shows the video time of the source recording and the camera when it is not the overview camera. The agent does not observe the top camera during execution.}
  \label{fig:demo_keyframes}
\end{figure}

\FloatBarrier
\subsection{Tool responses}
\label{app:prompts-feedback}

\paragraph{Direct control.} Every tool response consists of a JSON receipt, a text line with the measured robot state, and a caption and an image for each requested camera. The receipts under direct control have the same structure, and no receipt in the three configurations compares a command with the measured state. The exchange below is call 20 of the episode whose first turn is shown above. The right arm holds a bowl, and the commanded position lies 0.107\,m above the position reached. The receipt reports no error, so the shortfall is visible only from the measured state. The reason stated with the next call is appended.

\begin{Verbatim}
[Request, call 20: execute_eef]
{"cameras": ["overlook", "right_wrist"], "reason": "Lift the securely moving bowl slowly above sink.", "waypoints": [{"left": null, "right": {"xyz_m": [0.48996, -0.12997, 0.3], "quaternion_wxyz": [0.70722, 8e-05, 0.707, 0.00013], "gripper_m": [0, 0]}, "base_xy_yaw_deg": null, "motion_steps": 56, "steps": 64}]}
[Response]
{"executed_waypoints": [{"waypoint": 12, "executed_steps": 64, "measured_after": {"state.gripper": [0.044, 0.044, 0.00963, 0.0097], "state.base": [0.0003, 0.09162, -0.0], "state.ee_pose": [[[0.10034, -0.00058, 0.16485], [0.99998, 0.00116, 0.0035, -0.00427]], [[0.48621, -0.12846, 0.19264], [0.71538, -0.00048, 0.69874, 0.00032]]], "timestep": 1154}}], "remaining_tool_calls": null, "remaining_physics_steps": null, "reported_token_usage": {[...]}}
Measured robot state (EEF poses in each arm's planner base): [... the state of measured_after ...]
CURRENT RGB: overlook, 640x360
[image: overview camera]
CURRENT RGB: right_wrist, 384x288
[image: right wrist camera]
[Reason stated with the next call, call 21: execute_eef]
"Retract arm while raising to clear sink and reach a feasible carrying pose."
\end{Verbatim}

\paragraph{Post-trained policies.} A call to \texttt{execute\_pi05} returns the same structure, with a controller record in place of the waypoint receipts. The excerpt is call 1 of the same episode.

\begin{Verbatim}
[Request, call 1: execute_pi05]
{"cameras": ["overlook", "right_wrist"], "instruction": null, "reason": "Approach and align right hand with a bowl rim in the sink.", "steps": 120}
[Response]
{"executed_waypoints": [{"controller": "pi05", "requested_steps": 120, "executed_steps": 120, "inference_calls": 4, "control_returned_to": "agent", "episode_ended": false}], "remaining_tool_calls": null, "remaining_physics_steps": null, "reported_token_usage": {[...]}}
Measured robot state (EEF poses in each arm's planner base): {"state.gripper": [0.044, 0.044, 0.04398, 0.04398], "state.base": [-7e-05, 0.09306, -0.0], "state.ee_pose": [[[0.10035, 8e-05, 0.16352], [1.0, -0.00018, 0.00103, 0.00099]], [[0.35261, -0.03683, -0.00515], [0.70409, -0.00107, 0.71011, -0.00244]]], "timestep": 120}
[... captions and images of the overview and right wrist cameras ...]
\end{Verbatim}

A call to \texttt{execute\_wam} returns the same structure, and its controller record also states the fixed policy prompt and the number of inferences with a padded history frame. The excerpt is call 1 of the harness with InternW0-$\Delta$ on the same test scene.

\begin{Verbatim}
[Request, call 1: execute_wam]
{"cameras": ["overlook", "right_wrist"], "instruction": null, "reason": "Approach a bowl in the sink and align the right hand for a rim grasp.", "steps": 100}
[Response]
{"executed_waypoints": [{"controller": "wam", "requested_steps": 100, "prompt": {"prompt": "Take out one of the three bowls and stack it on the plate.", "source": "episode_instruction", "fixed_for": "episode", "prompt_sha256": "[hash]"}, "executed_steps": 100, "inference_calls": 10, "padded_recent_frames": 4, "control_returned_to": "agent", "episode_ended": false}], "remaining_tool_calls": null, "remaining_physics_steps": null, "reported_token_usage": {[...]}}
[... the measured state, and captions and images of the overview and right wrist cameras ...]
\end{Verbatim}

\paragraph{Fields.} Table~\ref{tab:fields} explains the fields of a response, which are the same in the three configurations apart from the controller records of the two policies.

\begin{table}[htbp]
\caption{Fields of a tool response.}
\label{tab:fields}
\centering
\scriptsize
\renewcommand{\arraystretch}{1.25}
\begin{tabularx}{\linewidth}{@{}>{\raggedright\arraybackslash}p{3.7cm}X@{}}
\toprule
\hd{Field} & \hd{Meaning} \\
\midrule
\texttt{executed\_waypoints} (EEF) & One record per executed waypoint: its running index within the episode, the physics steps executed, and the measured state after it. \\
\texttt{executed\_waypoints} ($\pi_{0.5}$) & The controller, the requested and executed physics steps, the number of policy inferences (one per 30 steps), and whether control returned to the agent or the episode ended. \\
\texttt{executed\_waypoints} (InternW0-$\Delta$) & As for $\pi_{0.5}$, with one inference per 10 steps, together with the fixed policy prompt and its source, scope, and SHA-256 hash, and the number of inferences in the call whose frame from 32 steps earlier had not been observed and was padded. \\
\texttt{state.gripper}, \texttt{state.base} & The four finger-joint positions in meters, left arm first (0.044 per finger is fully open), and the planar base pose ($x$, $y$ in meters, yaw in radians). \\
\texttt{state.ee\_pose}, \texttt{timestep} & For each arm, the EEF position in meters and its wxyz quaternion in that arm's planner-base frame, together with the physics steps executed since the start of the episode. \\
\texttt{remaining\_tool\_calls}, \texttt{remaining\_physics\_steps} & Always null, since no local limit applies and the benchmark's step limit is not disclosed. \\
\texttt{reported\_token\_usage} & Cumulative token counts of the session, omitted above. \\
\texttt{Measured robot state}, \texttt{CURRENT RGB} & The measured state after the call as a separate text line, and for each requested camera its name and resolution followed by its image. \\
\bottomrule
\end{tabularx}
\end{table}
  \FloatBarrier
  \setcounter{figure}{0}%
  \setcounter{table}{0}%
  \section{Additional Benchmark Results}
\label{app:results}

\subsection{Tasks, leaderboard, and per-task comparisons}

\begin{figure}[htbp]
  \centering
  \includegraphics[width=\linewidth]{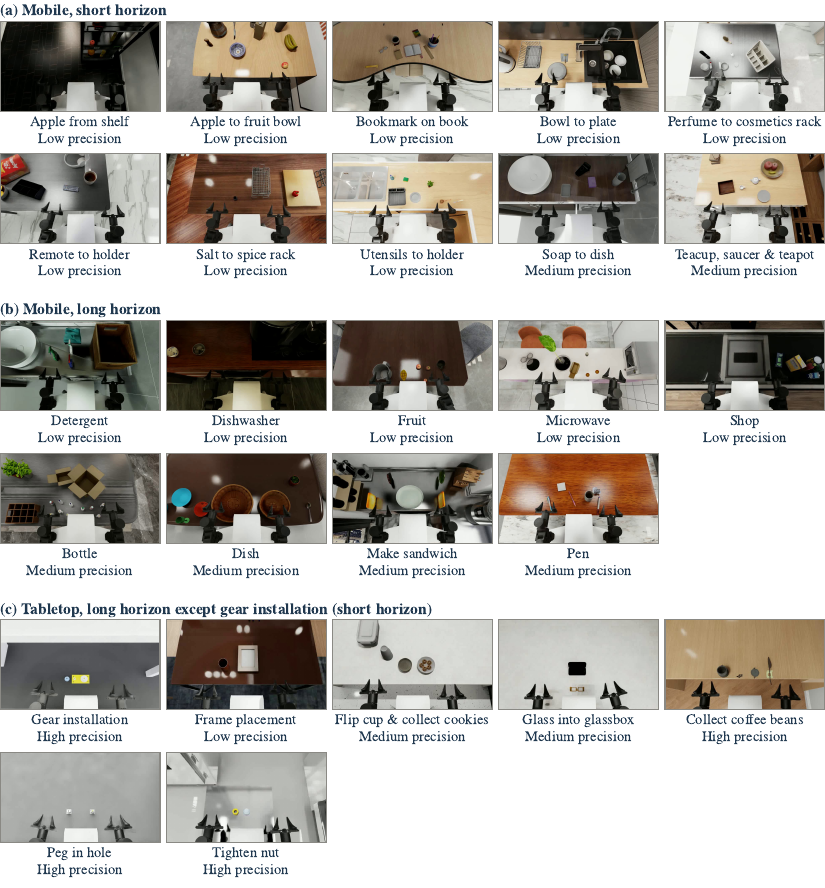}
  \caption{Initial overview frame of one direct-control test episode of each EBench task, by mobility and horizon, with the precision label of each task. Task names as in Table~\ref{tab:per_task_sr}. The tabletop frames show the upper central 80\% of the view.}
  \label{fig:task_gallery}
\end{figure}

\paragraph{Tasks.} Figure~\ref{fig:task_gallery} shows the initial overview frame of one test episode of each of the 26 tasks, grouped by mobility and horizon, with the precision label of each task.

\paragraph{Leaderboard.} Table~\ref{tab:leaderboard} lists all entries of the EBench leaderboard by mobility group, including the entries not shown in Table~\ref{tab:overall}: PICA-WAM~\citep{zou2026picawam}, which fine-tunes OpenWAM-$\alpha$ with an action-alignment objective under visual perturbations, Vela, AMapbot, Cosmos3-Edge, and X-VLA, as well as InternW0-$\Delta$ with the evaluation of Appendix~\ref{app:tool}. Among these 14 entries, the \astra-based agent under direct control ranks sixth in SR and fifth in score.

\begin{table}[htbp]
  \centering
  \fontsize{8}{9}\selectfont
  \setlength{\tabcolsep}{1.5pt}
  \caption{SR and score of all entries of the EBench leaderboard, overall and by mobility (19 mobile and 7 tabletop tasks, each averaged equally over tasks), with the best and second-best values in each column shown in bold and underlined. The results of the post-trained policies are taken from the leaderboard.}
  \label{tab:leaderboard}
  \begin{tabularx}{\linewidth}{>{\raggedright\arraybackslash}p{3.95cm}>{\centering\arraybackslash}p{1.15cm}*{6}{>{\centering\arraybackslash}X}}
\toprule
 & & \multicolumn{2}{c}{\hd{Overall}} & \multicolumn{2}{c}{\hd{Mobile}} & \multicolumn{2}{c}{\hd{Tabletop}} \\
\cmidrule(lr){3-4}\cmidrule(lr){5-6}\cmidrule(lr){7-8}
\hd{Policy} & \hd{Type} & \hd{SR (\%)} & \hd{Score} & \hd{SR (\%)} & \hd{Score} & \hd{SR (\%)} & \hd{Score} \\
\specialrule{\lightrulewidth}{\aboverulesep}{0pt}
\rowcolor{tabband} \multicolumn{8}{c}{\textit{\textbf{Post-trained policies}}} \\
\specialrule{\lightrulewidth}{0pt}{\belowrulesep}
PICA-WAM~\citep{zou2026picawam} & WAM & \best{61.4} & \best{0.749} & \best{67.5} & \best{0.783} & 45.0 & \second{0.663} \\
OpenWAM-$\alpha$~\citep{wang2026openwam} & WAM & \second{55.3} & \second{0.701} & \second{60.2} & \second{0.740} & 42.1 & 0.594 \\
Vela~\citep{li2026vela} & VLA & 49.7 & 0.655 & 51.7 & 0.661 & 44.3 & 0.641 \\
InternW0-$\Delta$~\citep{miao2026internw0deltaworldactionmodel} & WAM & 49.2 & 0.660 & 55.0 & 0.700 & 33.6 & 0.552 \\
AMapbot & -- & 48.9 & 0.639 & 49.3 & 0.642 & \second{47.9} & 0.630 \\
Qwen-RobotManip~\citep{yuan2026qwenrobotmanip} & VLA & 45.6 & 0.608 & 43.9 & 0.576 & \best{50.0} & \best{0.696} \\
$\pi_{0.5}$~\citep{pi05} & VLA & 41.4 & 0.544 & 45.9 & 0.578 & 29.3 & 0.452 \\
InternVLA-A1.5~\citep{ma2026internvla} & VLA & 34.2 & 0.468 & 42.3 & 0.562 & 12.1 & 0.215 \\
$\pi_0$~\citep{black2025pi0} & VLA & 33.7 & 0.475 & 34.0 & 0.469 & 32.9 & 0.491 \\
GigaBrain-0.7~\citep{gigabrainteam2026gigabrain} & VLA & 33.3 & 0.460 & 31.6 & 0.413 & 37.9 & 0.588 \\
Cosmos3-Edge~\citep{nvidia2026cosmos3} & WAM & 29.3 & 0.425 & 33.2 & 0.466 & 18.6 & 0.316 \\
Fast-WAM~\citep{yuan2026fastwam} & WAM & 25.6 & 0.371 & 34.0 & 0.460 & 2.9 & 0.129 \\
X-VLA~\citep{zheng2025xvla} & VLA & 23.7 & 0.358 & 29.3 & 0.402 & 8.6 & 0.241 \\
\specialrule{\lightrulewidth}{\aboverulesep}{0pt}
\rowcolor{tabband} \multicolumn{8}{c}{\textit{\textbf{\astra-based agent, no robot training}}} \\
\specialrule{\lightrulewidth}{0pt}{\belowrulesep}
\rowcolor{astralight} Direct control & MLLM & 46.7 & 0.654 & 56.6 & 0.730 & 20.0 & 0.447 \\
\bottomrule
\end{tabularx}

\end{table}

\paragraph{Per-task comparison.} Tables~\ref{tab:per_task_sr} and~\ref{tab:per_task_score} give the SR and score of the agent and the eight post-trained policies on every test task. Relative to the best post-trained policy on each task in Table~\ref{tab:per_task_sr}, the agent's SR is higher on three tasks, equal on five, and lower on eighteen. The three higher tasks are all mobile: Bookmark on book, Remote to holder, and Teacup, saucer \& teapot. The five equal tasks are Dish, Pen, Perfume to cosmetics rack, Salt to spice rack, and Bottle, which no policy solves. The largest deficits occur on Dishwasher, Flip cup \& collect cookies, and Glass into glassbox, where the agent reaches 5\%, 10\%, and 20\% against 90\% for the best post-trained policy. These three tasks involve an articulated object, coordinated use of both arms, and two-arm transfer with folding, respectively. The outcome breakdown separates two kinds of shortfall. On Tighten nut, Peg in hole, and Shop, no episode ends with a score of zero, so unsuccessful episodes receive partial credit for completed stages. On Gear installation and Utensils to holder, 15 and 13 of the 20 episodes end with a score of zero, so in these episodes the agent fails before completing the first scored stage.

\begin{table}[htbp]
  \centering
  \scriptsize
  \setlength{\tabcolsep}{2.8pt}
  \caption{SR (\%) per task on the EBench test set, with the best and second-best value of each task in bold and underlined. OWAM: OpenWAM-$\alpha$, IW0-$\Delta$: InternW0-$\Delta$, IVLA: InternVLA-A1.5, Giga: GigaBrain-0.7, FWAM: Fast-WAM. The direct-control run evaluated Bottle and Shop with a step limit of 3{,}000 steps instead of 5{,}000.}
  \label{tab:per_task_sr}
  \begin{tabularx}{\linewidth}{llll*{9}{>{\centering\arraybackslash}X}}
\toprule
\hd{Task} & \hd{Prec.} & \hd{Mob.} & \hd{Hor.} & \hd{OWAM} & \hd{IW0-$\Delta$} & \hd{Astra} & \hd{Qwen} & \hd{$\pi_{0.5}$} & \hd{IVLA} & \hd{$\pi_0$} & \hd{Giga} & \hd{FWAM} \\
\midrule
Apple from shelf & Low & Mob. & Short & \second{70.0} & 15.0 & \second{70.0} & 65.0 & \best{75.0} & 35.0 & 40.0 & 55.0 & 40.0 \\
\rowcolor{tabzebra} Apple to fruit bowl & Low & Mob. & Short & \best{95.0} & \best{95.0} & \second{90.0} & 50.0 & 25.0 & 20.0 & 10.0 & 20.0 & 70.0 \\
Bookmark on book & Low & Mob. & Short & 35.0 & 35.0 & \best{90.0} & 40.0 & \second{55.0} & 45.0 & 25.0 & 30.0 & 30.0 \\
\rowcolor{tabzebra} Bowl to plate & Low & Mob. & Short & \best{85.0} & \second{80.0} & 75.0 & 75.0 & \best{85.0} & 60.0 & 60.0 & 65.0 & 75.0 \\
Gear installation & High & Tab. & Short & \second{40.0} & \best{50.0} & 15.0 & \best{50.0} & 10.0 & 15.0 & 30.0 & 20.0 & 15.0 \\
\rowcolor{tabzebra} Perfume to cosmetics rack & Low & Mob. & Short & \best{75.0} & 30.0 & \best{75.0} & 55.0 & \second{60.0} & 45.0 & 25.0 & 25.0 & 50.0 \\
Remote to holder & Low & Mob. & Short & 65.0 & \second{85.0} & \best{100.0} & 65.0 & 60.0 & 60.0 & 70.0 & 50.0 & 50.0 \\
\rowcolor{tabzebra} Salt to spice rack & Low & Mob. & Short & \best{100.0} & \best{100.0} & \best{100.0} & 70.0 & 70.0 & \second{90.0} & 80.0 & 75.0 & 80.0 \\
Soap to dish & Med. & Mob. & Short & 85.0 & \best{100.0} & \second{95.0} & 85.0 & \best{100.0} & \second{95.0} & \second{95.0} & 70.0 & 90.0 \\
\rowcolor{tabzebra} Teacup, saucer \& teapot & Med. & Mob. & Short & 40.0 & \second{45.0} & \best{85.0} & \second{45.0} & 30.0 & 30.0 & 10.0 & 20.0 & 10.0 \\
Utensils to holder & Low & Mob. & Short & \best{35.0} & 15.0 & 10.0 & 15.0 & 5.0 & \second{20.0} & 0.0 & 10.0 & 0.0 \\
\rowcolor{tabzebra} Bottle & Med. & Mob. & Long & 0.0 & 0.0 & 0.0 & 0.0 & 0.0 & 0.0 & 0.0 & 0.0 & 0.0 \\
Collect coffee beans & High & Tab. & Long & 20.0 & 5.0 & 0.0 & 5.0 & \second{30.0} & 0.0 & 20.0 & \best{50.0} & 0.0 \\
\rowcolor{tabzebra} Detergent & Low & Mob. & Long & \second{95.0} & \second{95.0} & 55.0 & 85.0 & \best{100.0} & \second{95.0} & 75.0 & \second{95.0} & 35.0 \\
Dish & Med. & Mob. & Long & \second{40.0} & \best{50.0} & \best{50.0} & 20.0 & 25.0 & 20.0 & 0.0 & 5.0 & 0.0 \\
\rowcolor{tabzebra} Dishwasher & Low & Mob. & Long & \best{90.0} & \second{85.0} & 5.0 & 15.0 & 65.0 & \best{90.0} & \best{90.0} & 0.0 & 70.0 \\
Flip cup \& collect cookies & Med. & Tab. & Long & 40.0 & \second{50.0} & 10.0 & \best{90.0} & 25.0 & 30.0 & 30.0 & 40.0 & 0.0 \\
\rowcolor{tabzebra} Frame placement & Low & Tab. & Long & 50.0 & 60.0 & 65.0 & \best{95.0} & 45.0 & 0.0 & 25.0 & \second{70.0} & 0.0 \\
Fruit & Low & Mob. & Long & \second{55.0} & \best{60.0} & 40.0 & 45.0 & 10.0 & 5.0 & 0.0 & 0.0 & 0.0 \\
\rowcolor{tabzebra} Make sandwich & Med. & Mob. & Long & \best{80.0} & \second{73.3} & 46.7 & 66.7 & 66.7 & 33.3 & 20.0 & \best{80.0} & 46.7 \\
Microwave & Low & Mob. & Long & \best{93.3} & 6.7 & \second{73.3} & 33.3 & 40.0 & 60.0 & 46.7 & 0.0 & 0.0 \\
\rowcolor{tabzebra} Peg in hole & High & Tab. & Long & \best{40.0} & 10.0 & \second{20.0} & 10.0 & 10.0 & 0.0 & 15.0 & 15.0 & 0.0 \\
Pen & Med. & Mob. & Long & \second{5.0} & \best{15.0} & \best{15.0} & \second{5.0} & 0.0 & 0.0 & 0.0 & 0.0 & 0.0 \\
\rowcolor{tabzebra} Glass into glassbox & Med. & Tab. & Long & 75.0 & 60.0 & 20.0 & \best{90.0} & 65.0 & 0.0 & \second{85.0} & 55.0 & 5.0 \\
Shop & Low & Mob. & Long & \second{0.0} & \best{60.0} & \second{0.0} & \second{0.0} & \second{0.0} & \second{0.0} & \second{0.0} & \second{0.0} & \second{0.0} \\
\rowcolor{tabzebra} Tighten nut & High & Tab. & Long & \second{30.0} & 0.0 & 10.0 & 10.0 & 20.0 & \best{40.0} & 25.0 & 15.0 & 0.0 \\
\specialrule{\heavyrulewidth}{0pt}{\belowbottomsep}
\end{tabularx}

\end{table}

\begin{table}[htbp]
  \centering
  \scriptsize
  \setlength{\tabcolsep}{2.8pt}
  \caption{Score per task on the EBench test set, with the best and second-best value of each task in bold and underlined. Abbreviations and step limits as in Table~\ref{tab:per_task_sr}.}
  \label{tab:per_task_score}
  \begin{tabularx}{\linewidth}{llll*{9}{>{\centering\arraybackslash}X}}
\toprule
\hd{Task} & \hd{Prec.} & \hd{Mob.} & \hd{Hor.} & \hd{OWAM} & \hd{IW0-$\Delta$} & \hd{Astra} & \hd{Qwen} & \hd{$\pi_{0.5}$} & \hd{IVLA} & \hd{$\pi_0$} & \hd{Giga} & \hd{FWAM} \\
\midrule
Apple from shelf & Low & Mob. & Short & \second{0.70} & 0.15 & \second{0.70} & 0.65 & \best{0.75} & 0.35 & 0.40 & 0.55 & 0.40 \\
\rowcolor{tabzebra} Apple to fruit bowl & Low & Mob. & Short & \best{0.95} & \best{0.95} & \second{0.90} & 0.50 & 0.25 & 0.20 & 0.10 & 0.20 & 0.70 \\
Bookmark on book & Low & Mob. & Short & 0.35 & 0.35 & \best{0.90} & 0.40 & \second{0.55} & 0.45 & 0.25 & 0.30 & 0.30 \\
\rowcolor{tabzebra} Bowl to plate & Low & Mob. & Short & \best{0.85} & \second{0.80} & 0.75 & 0.75 & \best{0.85} & 0.60 & 0.60 & 0.65 & 0.75 \\
Gear installation & High & Tab. & Short & 0.50 & \best{0.60} & 0.20 & \second{0.58} & 0.13 & 0.20 & 0.38 & 0.28 & 0.15 \\
\rowcolor{tabzebra} Perfume to cosmetics rack & Low & Mob. & Short & \second{0.78} & 0.30 & \best{0.88} & 0.60 & 0.68 & 0.60 & 0.30 & 0.33 & 0.50 \\
Remote to holder & Low & Mob. & Short & 0.65 & \second{0.85} & \best{1.00} & 0.65 & 0.60 & 0.60 & 0.70 & 0.50 & 0.50 \\
\rowcolor{tabzebra} Salt to spice rack & Low & Mob. & Short & \best{1.00} & \best{1.00} & \best{1.00} & 0.70 & 0.70 & \second{0.90} & 0.80 & 0.75 & 0.80 \\
Soap to dish & Med. & Mob. & Short & 0.85 & \best{1.00} & \second{0.95} & 0.85 & \best{1.00} & \second{0.95} & \second{0.95} & 0.70 & 0.90 \\
\rowcolor{tabzebra} Teacup, saucer \& teapot & Med. & Mob. & Short & 0.53 & \second{0.58} & \best{0.93} & 0.50 & 0.40 & 0.33 & 0.15 & 0.28 & 0.23 \\
Utensils to holder & Low & Mob. & Short & \best{0.55} & \second{0.45} & 0.23 & 0.40 & 0.30 & 0.35 & 0.20 & 0.23 & 0.23 \\
\rowcolor{tabzebra} Bottle & Med. & Mob. & Long & \second{0.32} & \best{0.44} & 0.17 & 0.17 & 0.05 & 0.03 & 0.02 & 0.04 & 0.03 \\
Collect coffee beans & High & Tab. & Long & 0.40 & 0.11 & 0.24 & 0.51 & \second{0.55} & 0.16 & 0.35 & \best{0.71} & 0.00 \\
\rowcolor{tabzebra} Detergent & Low & Mob. & Long & \second{0.98} & \second{0.98} & 0.80 & 0.93 & \best{1.00} & 0.97 & 0.83 & \second{0.98} & 0.67 \\
Dish & Med. & Mob. & Long & \second{0.88} & \best{0.89} & 0.81 & 0.76 & 0.70 & 0.80 & 0.45 & 0.30 & 0.44 \\
\rowcolor{tabzebra} Dishwasher & Low & Mob. & Long & \best{0.95} & \second{0.93} & 0.53 & 0.27 & 0.73 & 0.90 & \best{0.95} & 0.50 & 0.87 \\
Flip cup \& collect cookies & Med. & Tab. & Long & 0.59 & \second{0.91} & 0.44 & \best{0.97} & 0.44 & 0.45 & 0.35 & 0.61 & 0.08 \\
\rowcolor{tabzebra} Frame placement & Low & Tab. & Long & 0.60 & 0.70 & \second{0.80} & \best{0.97} & 0.48 & 0.00 & 0.40 & 0.72 & 0.10 \\
Fruit & Low & Mob. & Long & \second{0.76} & \best{0.84} & 0.59 & 0.68 & 0.38 & 0.60 & 0.10 & 0.27 & 0.05 \\
\rowcolor{tabzebra} Make sandwich & Med. & Mob. & Long & \best{0.95} & \second{0.93} & 0.70 & 0.90 & 0.92 & 0.78 & 0.75 & \best{0.95} & 0.80 \\
Microwave & Low & Mob. & Long & \best{0.98} & 0.64 & \second{0.90} & 0.38 & 0.50 & 0.82 & 0.81 & 0.00 & 0.45 \\
\rowcolor{tabzebra} Peg in hole & High & Tab. & Long & \best{0.60} & 0.33 & \best{0.60} & 0.35 & 0.23 & 0.00 & 0.43 & \second{0.55} & 0.00 \\
Pen & Med. & Mob. & Long & \second{0.37} & 0.32 & \best{0.55} & 0.25 & 0.05 & 0.28 & 0.10 & 0.03 & 0.05 \\
\rowcolor{tabzebra} Glass into glassbox & Med. & Tab. & Long & 0.85 & 0.72 & 0.30 & \best{0.95} & 0.77 & 0.00 & \second{0.92} & 0.68 & 0.10 \\
Shop & Low & Mob. & Long & \second{0.67} & \best{0.90} & 0.60 & 0.60 & 0.58 & 0.17 & 0.44 & 0.30 & 0.09 \\
\rowcolor{tabzebra} Tighten nut & High & Tab. & Long & \second{0.63} & 0.50 & 0.55 & 0.55 & 0.58 & \best{0.70} & \second{0.63} & 0.58 & 0.48 \\
\specialrule{\heavyrulewidth}{0pt}{\belowbottomsep}
\end{tabularx}

\end{table}

\paragraph{Generalization dimensions.} Table~\ref{tab:generalization} gives SR and score in the four generalization dimensions. The Object dimension covers 24 tasks (120 episodes) and the other dimensions cover all 26 tasks (130 episodes), with five episodes per task and dimension.

\begin{table}[htbp]
  \centering
  \footnotesize
  \setlength{\tabcolsep}{4pt}
  \caption{SR (\%) and score of the \astra-based agent under direct control and the eight post-trained policies in the Object, Background, Instruction, and Mix generalization dimensions of the EBench test set, with the best and second-best value of each column in bold and underlined. Rows below the midrule are excluded from this ranking and report the agent in the harness with each policy, as analyzed in Section~\ref{sec:tool:groups}.}
  \label{tab:generalization}
  \begin{tabularx}{\linewidth}{l*{8}{>{\centering\arraybackslash}X}}
\toprule
 & \multicolumn{2}{c}{\hd{Object}} & \multicolumn{2}{c}{\hd{Background}} & \multicolumn{2}{c}{\hd{Instruction}} & \multicolumn{2}{c}{\hd{Mix}} \\
\cmidrule(lr){2-3}\cmidrule(lr){4-5}\cmidrule(lr){6-7}\cmidrule(lr){8-9}
\hd{Policy} & \hd{SR (\%)} & \hd{Score} & \hd{SR (\%)} & \hd{Score} & \hd{SR (\%)} & \hd{Score} & \hd{SR (\%)} & \hd{Score} \\
\midrule
\sysname{OpenWAM-$\alpha$} & \best{52.5} & \best{0.657} & \best{61.5} & \best{0.767} & \best{60.0} & \best{0.751} & \second{44.6} & 0.604 \\
\sysname{InternW0-$\Delta$} & \second{49.2} & 0.645 & 51.5 & 0.672 & \second{52.3} & \second{0.694} & \second{44.6} & \second{0.619} \\
\rowcolor{astralight} \sysname{\astra (direct control)} & 44.2 & \second{0.648} & 50.8 & 0.692 & 44.6 & 0.620 & \best{46.2} & \best{0.643} \\
\sysname{Qwen-RobotManip} & 40.8 & 0.565 & \second{55.4} & \second{0.722} & 51.5 & 0.655 & 33.8 & 0.485 \\
\sysname{$\pi_{0.5}$} & 41.7 & 0.534 & 42.3 & 0.575 & 46.2 & 0.597 & 34.6 & 0.457 \\
\sysname{InternVLA-A1.5} & 33.3 & 0.433 & 38.5 & 0.515 & 33.8 & 0.463 & 30.0 & 0.434 \\
\sysname{$\pi_0$} & 28.3 & 0.422 & 42.3 & 0.547 & 39.2 & 0.508 & 24.6 & 0.394 \\
\sysname{GigaBrain-0.7} & 27.5 & 0.405 & 42.3 & 0.541 & 36.9 & 0.502 & 25.4 & 0.389 \\
\sysname{Fast-WAM} & 21.7 & 0.295 & 32.3 & 0.440 & 26.9 & 0.394 & 21.5 & 0.330 \\
\midrule
\sysname{Harness with $\pi_{0.5}$} & 59.2 & 0.728 & 50.0 & 0.689 & 56.2 & 0.719 & 46.2 & 0.623 \\
\sysname{Harness with InternW0-$\Delta$} & 57.5 & 0.749 & 54.6 & 0.739 & 57.7 & 0.757 & 53.1 & 0.718 \\
\bottomrule
\end{tabularx}

\end{table}

\begin{figure}[htbp]
  \centering
  \includegraphics[width=0.84\linewidth]{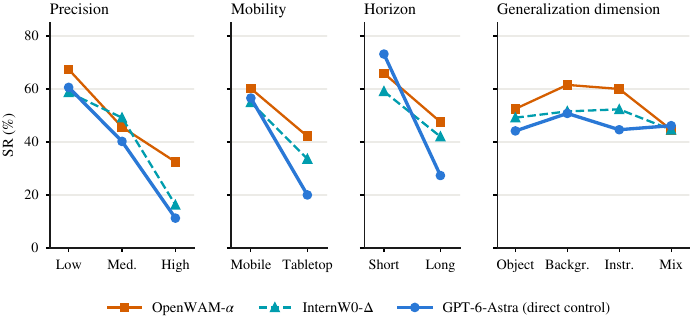}
  \caption{SR of OpenWAM-$\alpha$, InternW0-$\Delta$, and the \astra-based agent (direct control), the three highest-ranked systems in the per-task comparison, by precision, mobility, horizon, and generalization dimension on the 26 EBench test tasks.}
  \label{fig:profile}
\end{figure}

\paragraph{Capability profile.} Figure~\ref{fig:profile} compares the three highest-ranked systems among the agent and the eight post-trained policies by task group, as discussed in Section~\ref{sec:capability:profile}.

\FloatBarrier

\subsection{Evaluation protocol and event definitions}

\paragraph{Evaluation records.} The horizon labels follow the corrected annotation of the benchmark designers, which differs from the grouping displayed on the leaderboard. When the agent ends an episode, the robot is held in place until the evaluation server determines the outcome, and 6 of the 49 direct-control episodes that the agent ended early succeeded during this hold. In the 56 direct-control episodes interrupted by an infrastructure failure, execution resumed from the unchanged simulator state in a new model thread initialized with the recorded action history and the current observations. The direct-control run evaluated Bottle and Shop with a step limit of 3{,}000 steps, whereas the benchmark's current protocol, used by both harness configurations, sets a step limit of 5{,}000 steps for these two tasks. Within 3{,}000 steps, no episode of either task succeeds in any configuration. Under a step limit of 5{,}000 steps, the harness with policy solves 3 of 20 Shop episodes with $\pi_{0.5}$, 1 with InternW0-$\Delta$, and none of Bottle. These episodes raise the overall SR of the harness configurations from 52.4\% to 53.0\% and from 55.8\% to 56.0\%, respectively.

\paragraph{Comparison protocol.}
\label{app:caveats}
The results of the post-trained policies come from the public leaderboard, and those of InternW0-$\Delta$ alone come from the evaluation of Appendix~\ref{app:tool}, which is its leaderboard entry. Comparisons between configurations of the agent, and between InternW0-$\Delta$ alone and its harness configuration, pair test episodes that were run at different times, and each configuration was run once per episode. The two harness configurations differ in the policy's replanning interval and history as well as in the policy itself. The demonstrations include top-camera keyframes that the agent does not observe during execution, and for 31 of the 510 test episodes the demonstration is a recording of the test scene itself. The demonstration of Frame placement records the scene of four of its 20 test episodes. In the comparison with and without the demonstration (Appendix~\ref{app:icl}), the two conditions were rendered with partly different textures. The recordings shown in the figures were selected to illustrate each behavior.

\paragraph{Grasp events.}
\label{app:events}
The frequencies of disruptions and grip losses in Section~\ref{sec:capability} are computed from the finger gaps and commanded finger targets of both arms, recorded after every executed chunk of at most 8 physics steps, and from the stated reasons of all calls in the 510 direct-control episodes, including the segments resumed after an interruption. The finger gap of a hand is the summed opening of its two fingers, which reaches 88\,mm when the hand is fully open, or 0.044\,m per finger joint in \texttt{state.gripper}. A grip counts as closed on an object when, after a close command, the fingers come to rest between 2 and 80\,mm apart over two samples at different timesteps, so that fingertips closed for pushing do not count, and as lost when the gap then collapses while the gripper is still commanded closed. A stated reason counts as reporting a disruption when it names a slip, a drop, a fall, a displacement, or a loss outside a preventive phrase such as ``avoid'' or ``in case'', and words of an intended release such as ``released'' do not count. A loss counts as occurring in a grasp test when the reason of its call contains test, verify, confirm, or check as a whole word, or names a one-centimeter lift. In a random sample of 20 of the other losses, read together with the arm that loses the grip, 16 occur while an object is lifted or carried, 3 during door manipulation, and 1 during a placement.

\paragraph{Wrist and joint speed.}
\label{app:speed}
The speed measures of Section~\ref{sec:capability:safety} use the same chunk records, from all 510 episodes under direct control and from the segments of $\pi_{0.5}$ in all 510 episodes of the harness with $\pi_{0.5}$, as listed in Table~\ref{tab:speed}. Between consecutive chunks of one command, the wrist rotation is the angle between the measured EEF orientations of an arm, and the joint speed is the largest change of any arm joint, both divided by the executed physics steps. The two controllers differ in the tail of the rotation rates, as a single command of the agent can complete a large reorientation within a few chunks. In 38 commands, spread over 31 episodes, a wrist turns through 90$^\circ$ or more within 24 physics steps, whereas no segment of $\pi_{0.5}$ reaches this threshold. Most of these commands reorient or reposition the hand for a new approach. Fast joint motion shows the same contrast, as transitions above 0.08\,rad per physics step occur more than five times as often under direct control.

\begin{table}[htbp]
  \centering
  \small
  \caption{Wrist and joint speed between consecutive chunks of one command under direct control and in the segments of $\pi_{0.5}$ in the harness with $\pi_{0.5}$, over all 510 test episodes of each configuration. Parentheses give the number of episodes. A window spans three consecutive chunk transitions of one command.}
  \label{tab:speed}
  \begin{tabularx}{\linewidth}{l*{2}{>{\centering\arraybackslash}X}}
\toprule
 & \hd{Direct control} & \hd{$\pi_{0.5}$ in the harness} \\
\midrule
Chunk transitions measured & 109{,}079 & 88{,}573 \\
Median wrist rotation ($^\circ$ per step) & 0.00 & 0.08 \\
99th percentile ($^\circ$ per step) & 2.81 & 2.58 \\
99.9th percentile ($^\circ$ per step) & 4.06 & 3.81 \\
Transitions above 3$^\circ$ per step (episodes) & 643 (147) & 196 (94) \\
Commands turning a wrist by $90^\circ$ within a window of 24 steps (episodes) & 38 (31) & 0 (0) \\
Transitions with a joint above 0.08\,rad per step (episodes) & 606 (169) & 89 (59) \\
\bottomrule
\end{tabularx}

\end{table}

\subsection{Statistics}
\label{app:stats}

Aggregate differences between configurations are tested with a bootstrap over episodes within each task (5{,}000 resamples), which keeps the task composition fixed, while single-task differences use exact McNemar tests with Holm correction over the 26 tasks. For the harness with $\pi_{0.5}$ against direct control, the 95\% SR interval is $[+2.0, +10.5]$ points when episodes are resampled within each task and $[-2.5, +15.6]$ when tasks are resampled, and the score interval is $[+0.006, +0.071]$. By group, the SR intervals are $[+6.8, +19.0]$ for long-horizon, $[+3.6, +19.3]$ for tabletop, $[-8.6, +3.2]$ for short-horizon, and $[-11.3, +3.8]$ for high-precision tasks, and the score interval for high-precision tasks is $[-0.163, -0.029]$. The gain of the harness with InternW0-$\Delta$ over direct control is significant under both resampling schemes, and on the short-horizon tasks this configuration also exceeds InternW0-$\Delta$ alone under both. After Holm correction over the 26 tasks, Glass into glassbox and Dishwasher rise significantly over direct control with either tool, Teacup, saucer \& teapot falls significantly with $\pi_{0.5}$, and against InternW0-$\Delta$ alone, the corresponding harness configuration gains significantly on Apple from shelf and Perfume to cosmetics rack while losing significantly on Shop.
  \FloatBarrier
  \setcounter{figure}{0}%
  \setcounter{table}{0}%
  \section{Harness Details}
\label{app:tool}

\subsection{Post-trained policies}

\paragraph{$\pi_{0.5}$.} The VLA tool is the $\pi_{0.5}$ checkpoint fine-tuned on the EBench training set that appears as $\pi_{0.5}$ on the leaderboard. A call executes 1--300 physics steps. The policy predicts 50 absolute joint and gripper targets per inference and replans every 30 steps under the fixed task instruction, and control returns to the agent after the requested steps with the same observations as an EEF command. The agent's other tools and its prompt are those of Appendix~\ref{app:prompts}.

\paragraph{InternW0-$\Delta$.} The WAM tool is an InternW0-$\Delta$ checkpoint~\citep{miao2026internw0deltaworldactionmodel} post-trained on EBench data for 77{,}150 steps. It represents the robot by joint and EEF state and, in 10 denoising steps, predicts 32 absolute joint and gripper targets and base motions, without generating video at inference, of which the first 10 are executed before the next inference. Each inference is conditioned on the three camera views, the proprioceptive state, the first frame of the episode, and the frame 32 steps earlier, under the fixed task instruction. The harness opens one model session per episode with the episode's initial observation and executes each 10-action chunk as 8 and 2 steps, so that the frame 32 steps back is observed within a call. During the first inferences after the agent's own control, that frame is unavailable and is padded as at an episode start, which affects 4{,}276 of the 69{,}054 inferences. The agent invokes the policy 8.1 times per episode, with 135 inferences, against 45 inferences of $\pi_{0.5}$ in the configuration with $\pi_{0.5}$. Its standalone evaluation ran the same checkpoint with the same horizon, execution length, and history on all 510 test episodes with the original inference loop in which the earlier frame is always observed. This standalone evaluation is the entry of InternW0-$\Delta$ on the public leaderboard.

\subsection{Additional results}

\paragraph{Per-task results.} Table~\ref{tab:tool_per_task} lists every task, ordered by the change in SR with $\pi_{0.5}$. The harness with $\pi_{0.5}$ exceeds both the agent under direct control and $\pi_{0.5}$ alone on ten tasks and falls below the better of the two on thirteen. The ten tasks are Apple from shelf, Apple to fruit bowl, Bowl to plate, Dish, Flip cup \& collect cookies, Make sandwich, Perfume to cosmetics rack, Glass into glassbox, Utensils to holder, and Shop. On Shop, it solves 3 of 20 episodes at the 5{,}000-step limit, against none for $\pi_{0.5}$ alone and none for the direct-control run at 3{,}000 steps. The harness with $\pi_{0.5}$ loses 59 episodes that the agent solved under direct control. Of these, 11 come from Teacup, saucer \& teapot, 6 from Fruit, and 5 each from Dish, Frame placement, and Microwave. Of the 91 episodes that it gains, 14 come from Glass into glassbox, 11 from Dishwasher, and 7 each from Detergent, Dish, and Flip cup \& collect cookies. With InternW0-$\Delta$, the harness exceeds both the agent under direct control and InternW0-$\Delta$ alone on eight tasks and falls below the better of the two on eleven. These eight tasks are Apple from shelf, Apple to fruit bowl, Bowl to plate, Dish, Flip cup \& collect cookies, Make sandwich, Perfume to cosmetics rack, and Glass into glassbox. Of its 49 lost episodes, 8 come from Frame placement and 7 from Teacup, saucer \& teapot. Of its 96 gained episodes, 14 come from Dishwasher, 13 from Glass into glassbox, and 11 from Flip cup \& collect cookies.

\paragraph{Policy invocation measures.} The policy step share of an episode is the fraction of its executed physics steps that the policy executes, with EEF waypoints and gripper commands counted as the agent's steps, and the share of a task is the mean over its episodes. For Figure~\ref{fig:harness_dynamics}a, the executed steps of each episode are divided into tenths, and the policy's share is computed within each tenth. A policy invocation counts as a recovery request in Table~\ref{tab:harness_outcomes} when its stated reason contains a whole-word match for a term of recovery, retrying, slipping, dropping, missing, displacement, re-approach, regrasp, or failure, as in ``Recover slipped soap and realign a deeper centered grasp''. In a random sample of 40 calls coded in this way, 33 clearly ask the policy to recover from a disruption, 6 are ambiguous, and 1 is not a recovery request. Both harness configurations enter these measures with their Bottle and Shop episodes at 5{,}000 physics steps, as in Table~\ref{tab:tool}.

\begin{table}[htbp]
  \centering
  \scriptsize
  \setlength{\tabcolsep}{3pt}
  \caption{Per-task SR (\%) of \astra under direct control and in the harness with each policy, of each policy alone, and the share (\%) of physics steps executed by the policy in each harness configuration, with the best and second-best SR of each task in bold and underlined. The direct-control run evaluated Bottle and Shop with a step limit of 3{,}000 steps instead of 5{,}000.}
  \label{tab:tool_per_task}
  \begin{tabularx}{\linewidth}{l*{7}{>{\centering\arraybackslash}X}}
\toprule
\multirow{2}{*}{\hd{Task}} & \multirow{2}{=}{\centering\hd{Direct control}} & \multicolumn{2}{c}{\hd{Policy alone}} & \multicolumn{2}{c}{\hd{Harness with}} & \multicolumn{2}{c}{\hd{Policy step share}} \\
\cmidrule(lr){3-4}\cmidrule(lr){5-6}\cmidrule(lr){7-8}
 &  & \hd{$\pi_{0.5}$} & \hd{InternW0-$\Delta$} & \hd{$\pi_{0.5}$} & \hd{InternW0-$\Delta$} & \hd{$\pi_{0.5}$} & \hd{InternW0-$\Delta$} \\
\midrule
Teacup, saucer \& teapot & \best{85.0} & 30.0 & 45.0 & 30.0 & \second{55.0} & 51.9 & 48.9 \\
\rowcolor{tabzebra} Peg in hole & \best{20.0} & \second{10.0} & \second{10.0} & 5.0 & 0.0 & 61.1 & 63.4 \\
Microwave & \best{73.3} & 40.0 & 6.7 & \second{60.0} & \second{60.0} & 75.0 & 71.2 \\
\rowcolor{tabzebra} Fruit & 40.0 & 10.0 & \best{60.0} & 30.0 & \second{45.0} & 57.3 & 64.0 \\
Pen & \best{15.0} & 0.0 & \best{15.0} & \second{5.0} & 0.0 & 53.5 & 53.4 \\
\rowcolor{tabzebra} Remote to holder & \best{100.0} & 60.0 & 85.0 & \second{90.0} & \best{100.0} & 67.4 & 54.0 \\
Bookmark on book & \best{90.0} & 55.0 & 35.0 & \second{85.0} & \best{90.0} & 47.0 & 46.0 \\
\rowcolor{tabzebra} Gear installation & 15.0 & 10.0 & \best{50.0} & 10.0 & \second{40.0} & 59.4 & 72.3 \\
Soap to dish & \second{95.0} & \best{100.0} & \best{100.0} & 90.0 & \best{100.0} & 68.3 & 52.3 \\
\rowcolor{tabzebra} Bottle & 0.0 & 0.0 & 0.0 & 0.0 & 0.0 & 41.8 & 88.1 \\
Collect coffee beans & 0.0 & \best{30.0} & \second{5.0} & 0.0 & 0.0 & 75.8 & 69.1 \\
\rowcolor{tabzebra} Frame placement & \best{65.0} & 45.0 & \second{60.0} & \best{65.0} & 45.0 & 81.9 & 67.5 \\
Salt to spice rack & \best{100.0} & \second{70.0} & \best{100.0} & \best{100.0} & \best{100.0} & 79.0 & 74.0 \\
\rowcolor{tabzebra} Apple to fruit bowl & 90.0 & 25.0 & \second{95.0} & \second{95.0} & \best{100.0} & 48.9 & 62.4 \\
Tighten nut & 10.0 & \best{20.0} & 0.0 & \second{15.0} & 10.0 & 53.6 & 47.0 \\
\rowcolor{tabzebra} Utensils to holder & \second{10.0} & 5.0 & \best{15.0} & \best{15.0} & \best{15.0} & 55.7 & 54.4 \\
Perfume to cosmetics rack & \second{75.0} & 60.0 & 30.0 & \best{80.0} & \best{80.0} & 62.9 & 48.1 \\
\rowcolor{tabzebra} Dish & 50.0 & 25.0 & 50.0 & \second{60.0} & \best{70.0} & 73.6 & 70.1 \\
Shop & 0.0 & 0.0 & \best{60.0} & \second{15.0} & 5.0 & 89.7 & 82.9 \\
\rowcolor{tabzebra} Apple from shelf & 70.0 & \second{75.0} & 15.0 & \best{85.0} & \second{75.0} & 53.4 & 39.3 \\
Bowl to plate & 75.0 & \second{85.0} & 80.0 & \best{95.0} & \best{95.0} & 47.0 & 44.6 \\
\rowcolor{tabzebra} Make sandwich & 46.7 & 66.7 & \second{73.3} & \second{73.3} & \best{86.7} & 82.3 & 86.1 \\
Flip cup \& collect cookies & 10.0 & 25.0 & \second{50.0} & 40.0 & \best{60.0} & 78.9 & 83.0 \\
\rowcolor{tabzebra} Detergent & 55.0 & \best{100.0} & \second{95.0} & 90.0 & 70.0 & 88.4 & 83.2 \\
Dishwasher & 5.0 & 65.0 & \best{85.0} & 60.0 & \second{75.0} & 91.6 & 88.2 \\
\rowcolor{tabzebra} Glass into glassbox & 20.0 & 65.0 & 60.0 & \best{85.0} & \second{80.0} & 93.8 & 85.4 \\
\specialrule{\heavyrulewidth}{0pt}{\belowbottomsep}
\end{tabularx}

\end{table}

\paragraph{Generalization dimensions.} Of the direct-control successes, the harness with $\pi_{0.5}$ loses 43 of 126 on the Background and Mix dimensions against 16 of 111 on Object and Instruction (Fisher's exact test, $p=0.0005$). With InternW0-$\Delta$, the corresponding counts are 31 of 126 and 18 of 111 ($p=0.15$).

\begin{figure}[htbp]
  \centering
  \includegraphics[width=0.62\linewidth]{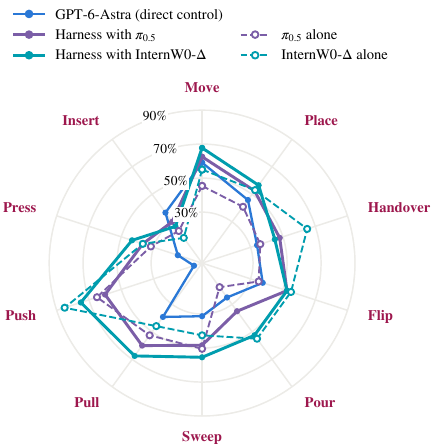}
  \caption{SR by atomic skill of the \astra-based agent under direct control, in the harness with $\pi_{0.5}$ and InternW0-$\Delta$, and of each policy standalone. Each skill averages over the tasks that require it: placing (23 tasks), moving (17), handover (5), sweeping (4), inserting (3), pressing (3), flipping (2), pouring (2), pulling (2), and pushing (1). Grasping, which is required by all 26 tasks, is omitted. The radial axis ends at the largest SR, rounded up to a multiple of 10\%.}
  \label{fig:harness_skills}
\end{figure}

\paragraph{Atomic skills.}\label{app:skills} Figure~\ref{fig:harness_skills} groups the per-task SRs of Table~\ref{tab:tool_per_task} by the atomic skills that EBench annotates for each task. With $\pi_{0.5}$ and InternW0-$\Delta$ alike, mean SR rises most on pushing, pressing, and pulling, by 21 to 70 percentage points, and sweeping improves by 17 and 24 points, respectively. Several tasks require more than one of these skills, as Dishwasher counts toward pushing, pulling, and sweeping, and Glass into glassbox accounts for most of the increase on pressing. Insertion is the only skill whose mean SR drops in both harness configurations, from 36.7\% under direct control to 30.0\% with $\pi_{0.5}$ and 26.7\% with InternW0-$\Delta$, while the two policies alone reach 23.3\% and 18.3\%. On handover, the harness with InternW0-$\Delta$ reaches 45.0\% against 65.0\% for InternW0-$\Delta$ alone, its largest shortfall against the standalone policy among the skills.

\paragraph{Diagnosis.} An episode is classified as agent-terminated when the agent stops before the step limit without succeeding. Phrase rules assign its final message to either an unconfirmed completion claim or a reported failure, and a manual check of ten episodes per run confirms this assignment. Table~\ref{tab:tool_transitions} follows every episode of the agent under direct control to the same scene in each harness configuration. The two harness configurations newly fail 108 episodes that the agent solved under direct control: 59 with $\pi_{0.5}$ and 49 with InternW0-$\Delta$, on 85 distinct test episodes. Each of these episodes was read call by call, with the executor, the finger widths before and after, the base motion, and the agent's stated reason of every call. Each was then assigned the incident after which the unfinished stage was never completed, as Table~\ref{tab:tool_losses} lists. Each agent-terminated episode is assigned the decisive incident preceding termination: a slip of the agent's grasp, a loss around a policy segment, or an unmet stage behind a completion claim. A grip counts as held when the fingers come to rest between 2 and 80\,mm apart over two samples at different timesteps while commanded closed, and as lost when it collapses while still commanded closed, and a reported slip is a stated reason coded as in Appendix~\ref{app:events}. An unmaintained grasp is an agent command that follows a policy segment and leaves an arm holding an object unspecified rather than closing it, so that the arm holds its measured finger width. Unmaintained grasps are counted per arm, so the 338 and 316 of Section~\ref{sec:tool:diagnosis} fall in 337 and 313 agent commands. The comparison in Section~\ref{sec:tool:diagnosis} counts unspecified holding arms after the agent's own commands within the same configuration, and its stratified odds ratios use the Mantel--Haenszel estimator over the tasks on which both kinds occur.

\begin{table}[h]
  \centering
  \footnotesize
  \setlength{\tabcolsep}{3pt}
  \caption{How the 510 episodes under direct control end and how the same scenes end in each harness configuration, by success, step limit, a completion claim, or a failure report of the agent, with the few episodes of other kinds omitted from the columns of the harness configurations.}
  \label{tab:tool_transitions}
  \begin{tabularx}{\linewidth}{l*{9}{>{\centering\arraybackslash}X}}
\toprule
 & & \multicolumn{4}{c}{\hd{Harness with $\pi_{0.5}$}} & \multicolumn{4}{c}{\hd{Harness with InternW0-$\Delta$}} \\
\cmidrule(lr){3-6}\cmidrule(lr){7-10}
\hd{Direct control} & \hd{$n$} & \hd{Succ.} & \hd{Limit} & \hd{Claim} & \hd{Fail} & \hd{Succ.} & \hd{Limit} & \hd{Claim} & \hd{Fail} \\
\midrule
Success & 237 & 178 & 52 & 3 & 4 & 188 & 33 & 2 & 13 \\
\rowcolor{tabzebra} Step limit & 229 & 72 & 140 & 6 & 10 & 76 & 135 & 8 & 9 \\
Ended, claims completion & 21 & 10 & 3 & 6 & 2 & 11 & 5 & 3 & 2 \\
\rowcolor{tabzebra} Ended, reports failure & 22 & 8 & 6 & 0 & 8 & 9 & 4 & 0 & 9 \\
Other & 1 & 1 & 0 & 0 & 0 & 0 & 1 & 0 & 0 \\
\bottomrule
\end{tabularx}

\end{table}

\begin{table}[h]
  \centering
  \footnotesize
  \caption{Decisive incident in the episodes that the agent solved under direct control and each harness configuration lost.}
  \label{tab:tool_losses}
  \begin{tabularx}{\linewidth}{X*{2}{>{\centering\arraybackslash}p{1.9cm}}}
\toprule
\hd{Decisive incident} & \hd{With $\pi_{0.5}$} & \hd{With InternW0-$\Delta$} \\
\midrule
Object lost or misplaced during a policy segment or when the agent resumes control & 16 & 30 \\
\rowcolor{tabzebra} Policy without the object, on the wrong object or subgoal, or undoing a stage & 5 & 1 \\
Agent's own grasp slips & 30 & 12 \\
\rowcolor{tabzebra} Completion claimed while a stage remains unmet & 3 & 2 \\
Other & 5 & 4 \\
\midrule
Total & 59 & 49 \\
\bottomrule
\end{tabularx}

\end{table}

\begin{table}[htbp]
  \centering
  \footnotesize
  \setlength{\tabcolsep}{3.5pt}
  \caption{Decisive incidents of the 164 test episodes that fail with both harness configurations, read call by call, per configuration, with the number of episodes whose two runs show the same incident and the tasks that contribute most.}
  \label{tab:tool_shared}
  \begin{tabularx}{\linewidth}{@{}>{\raggedright\arraybackslash}p{5.2cm}ccc>{\raggedright\arraybackslash}X@{}}
    \toprule
    \hd{Decisive incident} & \hd{$\pi_{0.5}$} & \hd{InternW0-$\Delta$} & \hd{Same in both} & \hd{Main tasks} \\
    \midrule
    Agent's own grasp slips (lift test or transport) & 58 & 42 & 19 & Utensils to holder, Peg in hole, Pen \\
    Object lost or misplaced in a policy segment & 37 & 53 & 10 & Peg in hole, Pen, Shop \\
    Contact stage never completed & 19 & 24 & 12 & Tighten nut, Gear installation, Peg in hole \\
    Policy segment empty-handed or on the wrong object & 16 & 15 & 4 & Shop, Bottle \\
    Step limit without a decisive loss & 16 & 11 & 4 & Bottle \\
    Slip after resuming control without a maintained grasp & 14 & 10 & 7 & Collect coffee beans \\
    False completion claim & 3 & 9 & 2 & Gear installation, Dishwasher \\
    Other & 1 & 0 & 0 & \\
    \midrule
    Total & 164 & 164 & 58 & \\
    \bottomrule
  \end{tabularx}
\end{table}

\paragraph{Invocation and outcome.}
\label{app:tool:invocation}
Table~\ref{tab:invocation_stats} compares successful and unsuccessful episodes of the two harness configurations. A motion command is a call of the EEF interface, the gripper command, or the policy, and a transition is a change of executor between two consecutive motion commands. Pooled over episodes, the number of transitions correlates with failure, with point-biserial correlations of $r=0.42$ and $r=0.44$, and within the 23 and 18 tasks with both outcomes, unsuccessful episodes make more transitions on 22 and 17 tasks (Wilcoxon signed-rank $p=2.1\times10^{-4}$ and $1.5\times10^{-5}$). They also issue more motion commands on every one of these tasks, with fewer transitions per motion command on 17 of the 23 tasks ($p=0.015$) and 15 of the 18 ($p=0.06$). Episodes with a recovery request to the policy fail in 75.5\% and 72.0\% of cases, against 15.8\% and 16.8\% without one. Of the 251 and 239 episodes with a disruption report, 227 and 200 contain a recovery request.

\begin{table}[htbp]
\centering
\footnotesize
\caption{Policy invocation in the successful and unsuccessful episodes of the two harness configurations. Transitions are changes of executor between consecutive motion commands, and disruption reports are motion commands whose stated reasons report a slip, a drop, or a displaced object.}
\label{tab:invocation_stats}
\begin{tabularx}{\linewidth}{>{\raggedright\arraybackslash}p{4.4cm}*{4}{>{\centering\arraybackslash}X}}
\toprule
 & \multicolumn{2}{c}{\hd{Harness with $\pi_{0.5}$}} & \multicolumn{2}{c}{\hd{Harness with InternW0-$\Delta$}} \\
\cmidrule(lr){2-3}\cmidrule(lr){4-5}
 & \hd{Successful} & \hd{Unsuccessful} & \hd{Successful} & \hd{Unsuccessful} \\
\midrule
Episodes & 269 & 241 & 284 & 226 \\
Motion commands per episode & 14.7 & 32.0 & 15.2 & 30.4 \\
Transitions per episode & 4.2 & 7.2 & 3.8 & 6.8 \\
Transitions per motion command & 0.33 & 0.26 & 0.28 & 0.24 \\
Episodes ending with an agent command (\%) & 42.8 & 68.0 & 47.2 & 71.7 \\
Episodes with a recovery request to the policy (\%) & 24.5 & 84.2 & 25.0 & 81.0 \\
Disruption reports executed by the policy (\%) & 50.5 & 32.4 & 39.6 & 32.7 \\
\bottomrule
\end{tabularx}

\end{table}

\paragraph{Grip losses by executor.} Using the held state and the losses defined in Appendix~\ref{app:events}, every physics step between two samples of a held arm counts as a held arm-step of the executor of its call, and rates are given per 1{,}000 held arm-steps. The harness with $\pi_{0.5}$ holds objects for 278{,}496 arm-steps under the policy and 146{,}922 under the agent, with 176 and 321 losses, corresponding to rates of 0.63 and 2.18 per 1{,}000 held arm-steps. With InternW0-$\Delta$, the corresponding values are 262{,}037 and 142{,}024 arm-steps, with 189 and 315 losses, or 0.72 and 2.22 per 1{,}000. Of the agent's losses, 193 and 173 fall in grasp-test calls, and outside these calls the agent's rates are 0.87 and 1.00. Under direct control, 811 losses over 447{,}352 held arm-steps give 1.81 per 1{,}000, and 1.05 outside grasp tests. Among the tasks with more than 2{,}000 held arm-steps under both executors, the agent's rate exceeds the policy's on 13 of 16 tasks with $\pi_{0.5}$ and 15 of 18 with InternW0-$\Delta$.

\subsection{API-equivalent cost}
\label{app:tool:cost}

Prompt caching dominates the API-equivalent cost of every configuration (Table~\ref{tab:api_cost}). On each request, the agent resends the growing episode context, including the instructions, the demonstration, earlier observations, and tool results, so most of the input repeats an earlier prefix from the same episode and is billed at the cached-input rate. At the Standard tariff of \astra (USD 10, 1, 12.5, and 50 per million uncached input, cached input, cache-write, and output tokens), cached input covers 91.5--97.0\% of the input tokens and still accounts for 44.0--70.5\% of the cost per episode despite its tenfold discount. The recorded tokens cost USD 7.82 per episode under direct control and 2.63 and 3.06 in the harness with $\pi_{0.5}$ and with InternW0-$\Delta$, corresponding to 16.82, 4.99, and 5.49 per success. For 44 direct-control episodes the final token record is missing, and the direct-control values here, in Figure~\ref{fig:token_cost}, and in Section~\ref{sec:capability:failures} count the recorded tokens. In all three configurations, token use rises with the number of tool calls, with Pearson correlations of 0.88 under direct control and 0.94 and 0.87 in the harness with $\pi_{0.5}$ and with InternW0-$\Delta$. No request exceeds the 272k input tokens above which the tariff raises input and cache rates twofold and output rates by half. Only the direct-control run records cache writes, because part of its episodes ran through the API rather than the subscription route, which reports none.

\begin{table}[htbp]
\centering
\footnotesize
\setlength{\tabcolsep}{3pt}
\caption{Mean token composition and API-equivalent cost per episode of \astra on the 510 EBench test episodes, with input split into cached input, cache writes, and other uncached input, and the cost per success dividing the total cost by the number of successes.}
\label{tab:api_cost}
\resizebox{\linewidth}{!}{%
\begin{tabular}{@{}lccccccc@{}}
\toprule
\hd{Configuration} & \hd{Tokens (M)} & \hd{Cached (\%)} & \hd{Uncached (k)} & \hd{Writes (k)} & \hd{Output (k)} & \hd{USD/episode} & \hd{USD/success} \\
\midrule
Direct control & 3.77 & 91.5 & 75.6 & 244.8 & 11.2 & 7.82 & 16.82 \\
Harness with $\pi_{0.5}$ & 1.83 & 96.5 & 64.2 & 0 & 4.6 & 2.63 & 4.99 \\
Harness with InternW0-$\Delta$ & 2.22 & 97.0 & 65.7 & 0 & 4.9 & 3.06 & 5.49 \\
\bottomrule
\end{tabular}%
}
\end{table}

  \FloatBarrier
  \setcounter{figure}{0}%
  \setcounter{table}{0}%
  \section{Additional Qualitative Results}
\label{app:samples}

\begin{figure}[htbp]
  \centering
  \includegraphics[width=\linewidth]{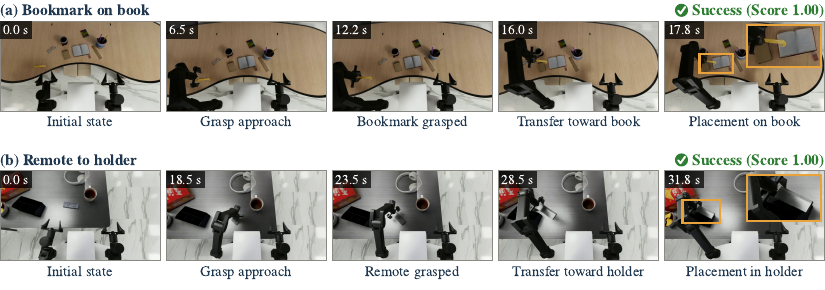}
  \caption{Two successful direct-control episodes of the \astra-based agent: Bookmark on book and Remote to holder. Overview frames with video time, with insets enlarging the outlined grasp regions.}
  \label{fig:successful}
\end{figure}

\begin{figure}[htbp]
  \centering
  \includegraphics[width=\linewidth]{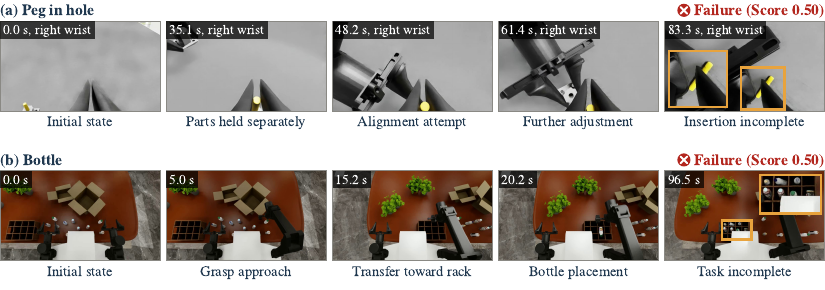}
  \caption{Two unsuccessful direct-control episodes of the \astra-based agent, each ending with score 0.50: Peg in hole (right-wrist view) and Bottle (overview). Video time is shown on each frame.}
  \label{fig:incomplete}
\end{figure}

\begin{figure}[htbp]
  \centering
  \includegraphics[width=\linewidth]{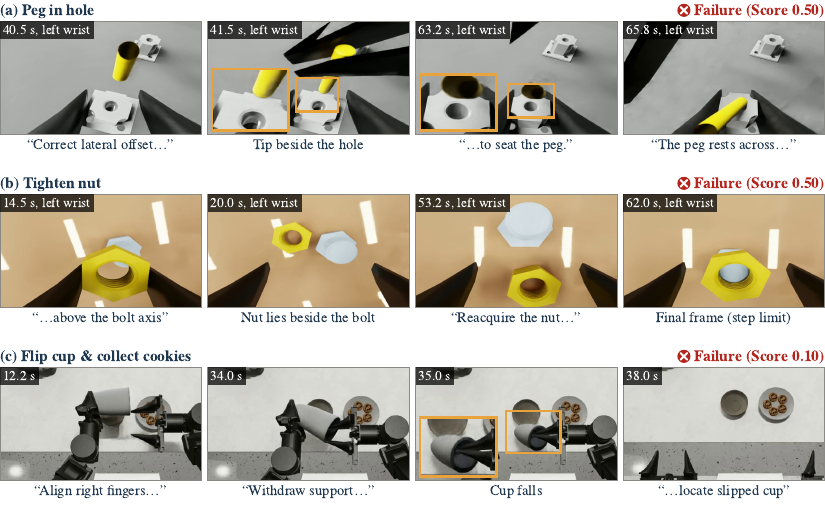}
  \caption{Three unsuccessful direct-control episodes of the \astra-based agent, with the agent's stated reasons below the frames. Orange boxes are enlarged in the insets. (a) Peg in hole: the peg tip lands beside the hole, and the peg ends lying across the receiver at the step limit (score 0.50). (b) Tighten nut: the nut falls beside the bolt, is re-acquired, and rests unthreaded on the bolt tip at the step limit (score 0.50). (c) Flip cup \& collect cookies: during the second handover the supporting hand withdraws, and the cup falls off the table (score 0.10).}
  \label{fig:failure_modes}
\end{figure}

This appendix collects keyframes and prompt excerpts from the recordings of the direct-control evaluation. Appendix~\ref{app:cases} gives prompt-grounded cases, Appendix~\ref{app:policy_rollouts} compares rollouts of the agent with those of two post-trained policies, and Appendix~\ref{app:icl} shows the episodes with and without the demonstration. Figure~\ref{fig:successful} shows two successful episodes on tasks where the agent leads, as discussed in Section~\ref{sec:capability:strengths}. Figure~\ref{fig:incomplete} shows two unsuccessful episodes that end with partial credit, as discussed in Section~\ref{sec:capability:deficits}. Figure~\ref{fig:failure_modes} shows three unsuccessful episodes of the kind analyzed in Section~\ref{sec:capability:failures}: two that stop at the final stage of a high-precision task and one that fails in a two-arm handover.

\begin{figure}[htbp]
  \centering
  \includegraphics[width=\linewidth]{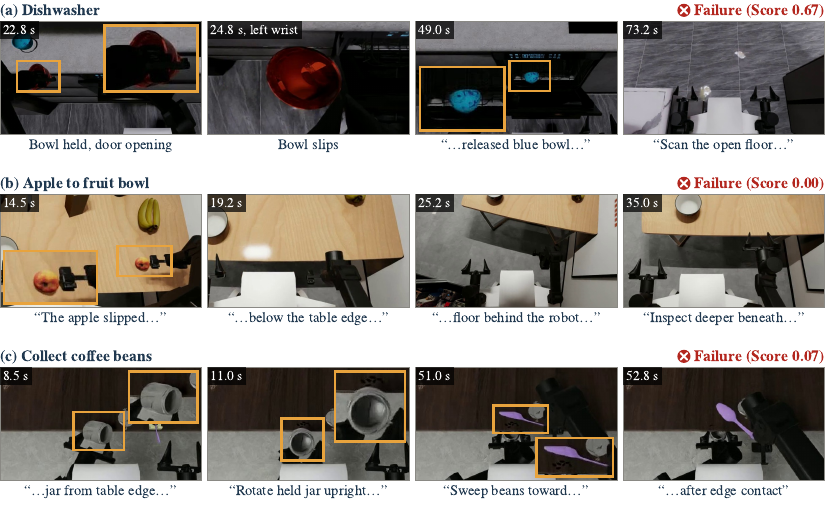}
  \caption{Three direct-control episodes of the \astra-based agent with safety-relevant behavior, with the agent's stated reasons below the frames. Orange boxes are enlarged in the insets. (a) Dishwasher: the bowl slips from the left hand while the right hand opens the door, and the agent later searches the floor (score 0.67). (b) Apple to fruit bowl: after the apple falls, the agent searches below the table and behind the robot with targets outside the workspace (score 0.00). (c) Collect coffee beans: the held jar and later the spoon contact the table edge (score 0.07).}
  \label{fig:safety}
\end{figure}

\begin{figure}[htbp]
  \centering
  \includegraphics[width=\linewidth]{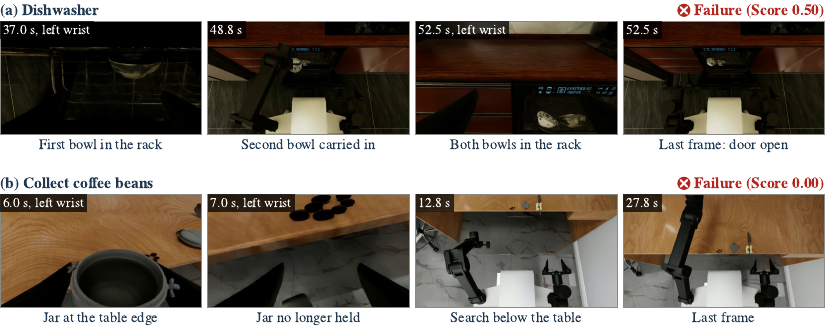}
  \caption{Two unsuccessful direct-control episodes that the \astra-based agent ended itself, with the video time and camera of each frame. (a) Dishwasher: the final message reports both bowls in the rack while the door is open. (b) Collect coffee beans: the held jar is lost at the table edge.}
  \label{fig:self_ended}
\end{figure}

Figure~\ref{fig:safety} shows three episodes with the safety-relevant behavior discussed in Section~\ref{sec:capability:safety}, and Figure~\ref{fig:self_ended} shows the two ways in which the agent ends an unsuccessful episode itself, as discussed in Section~\ref{sec:capability:failures}.

\subsection{Prompt-grounded cases}
\label{app:cases}

The stored prompts and stated reasons of unsuccessful episodes show how supplied guidance and its execution diverge. The first box below belongs to panel b of Figure~\ref{fig:adaptive}, and the second shows a Fruit episode in which the agent follows the live instruction over the demonstration. Both list the guidance and the agent's response.

\begin{promptcase}{Contact geometry: Collect coffee beans, score 0.50}
\textbf{Execution prompt.} ``Measured local EEF z is not height above the tabletop.''\\
\textbf{Demonstration.} ``With the right hand, grasp the spoon handle, lift it out of its holder, and sweep the beans toward the jar.''\\
\textbf{Agent, call 21.} ``Angle spoon edge down to tabletop behind beans''\\
\textbf{Outcome.} The lowered and tilted sweeps do not complete the collection, and the episode ends at score 0.50.
\end{promptcase}

\begin{promptcase}{Instruction against demonstration: Fruit, score 0.60}
\textbf{Task instruction.} ``Make a fruit milkshake directly in the cup.''\\
\textbf{Demonstration.} ``The left hand then carries and releases each fruit into the large pale jug.''\\
\textbf{Agent, call 8.} ``Carry the secured fruit above the small cup, not the large jug.''\\
\textbf{Outcome.} The agent carries the fruit toward the right container, and the episode ends at score 0.60 without success.
\end{promptcase}

\subsection{Rollouts of the agent and post-trained policies}
\label{app:policy_rollouts}

\begin{figure}[htbp]
  \centering
  \includegraphics[width=\linewidth]{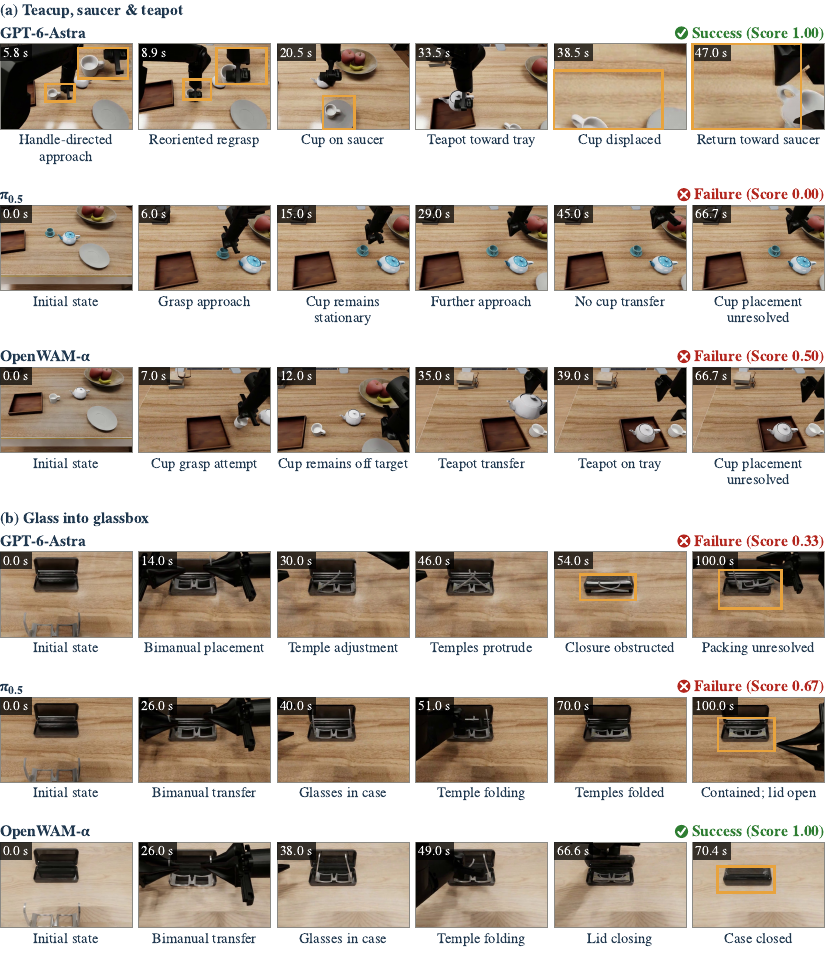}
  \caption{The \astra-based agent (direct control), $\pi_{0.5}$, and OpenWAM-$\alpha$ on two tasks: (a) Teacup, saucer \& teapot and (b) Glass into glassbox. Keyframes are in chronological order with video time. The outcome and score of each rollout are given at the right, and insets show enlarged views of the outlined manipulation regions. The rollouts differ in object variants and viewpoints.}
  \label{fig:policies}
\end{figure}

\paragraph{Recovery from subgoal disruptions.} Figure~\ref{fig:policies} compares rollouts of the \astra-based agent, $\pi_{0.5}$, and OpenWAM-$\alpha$ on the two tasks summarized in Section~\ref{sec:capability:cases}. In Teacup, saucer \& teapot (Figure~\ref{fig:policies}a), the agent first approaches the cup handle and refines its wrist and gripper configuration over successive attempts before grasping and placing the cup onto the saucer. When subsequent manipulation of the teapot inadvertently displaces the cup from the saucer, the agent detects that the completed subgoal has been invalidated, re-grasps the cup, and restores it to the saucer rather than continuing the remaining action sequence. By contrast, $\pi_{0.5}$ repeatedly approaches and retracts without completing the cup transfer, whereas OpenWAM-$\alpha$ fails to grasp the cup and proceeds to place the teapot onto the tray while leaving the cup off the saucer.

\paragraph{Precise contact in object manipulation.} In Glass into glassbox (Figure~\ref{fig:policies}b), although the agent uses both arms to place the glasses into the open case, its subsequent adjustments fail to fold the temples fully inward, leaving a protruding temple that prevents the lid from closing despite repeated corrective pushes. By comparison, $\pi_{0.5}$ places the glasses into the case and folds both temples compactly (though leaving the lid open), while OpenWAM-$\alpha$ completes both temple folding and lid closure.

\subsection{In-context learning}
\label{app:icl}

\begin{figure}[htbp]
  \centering
  \includegraphics[width=\linewidth]{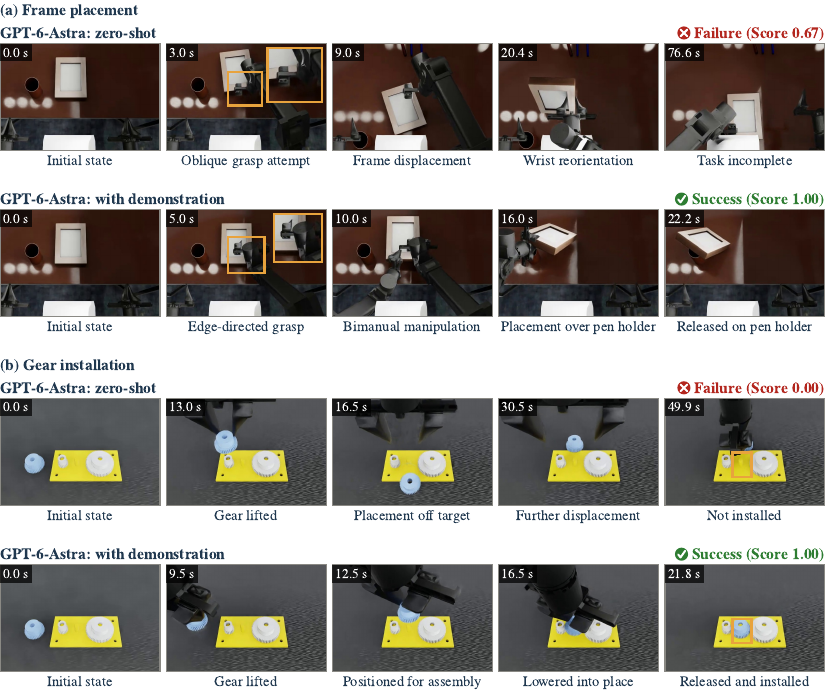}
  \caption{The \astra-based agent (direct control) without (zero-shot) and with the demonstration on the same test episodes: (a) Frame placement and (b) Gear installation. Keyframes are in chronological order with video time, and columns do not correspond to matched elapsed time.}
  \label{fig:icl}
\end{figure}

Figure~\ref{fig:icl} shows keyframes from the paired evaluation with and without the annotated demonstration across all 20 test episodes of Frame placement and Gear installation (Section~\ref{sec:capability:cases}). On Frame placement (1/20 successes zero-shot vs.\ 13/20 with the demonstration, mean score 0.283 vs.\ 0.800), under zero-shot control the agent reaches and displaces the frame, but repeatedly adjusts its approach trajectory and wrist orientation without achieving stable placement. When conditioned on the demonstration, the agent coordinates both grippers to hold and transfer the frame in 17 of 20 episodes (including all 13 successful episodes, compared with 1 of 20 under zero-shot control) and positions the frame upright against the pen holder (Figure~\ref{fig:icl}a). On Gear installation (3/20 successes under both conditions), without the demonstration the agent grasps and lifts the gear but leaves it outside the assembly region, whereas with the demonstration it brings the gear between the two mounted gears and lowers it, but it scores no stage in 15 of the 20 episodes (Figure~\ref{fig:icl}b).
  \FloatBarrier
  \setcounter{figure}{0}%
  \setcounter{table}{0}%
  \section{Case Studies in the Harness with Policy}
\label{app:case_studies}

\begin{figure}[htbp]
  \centering
  \includegraphics[width=\linewidth]{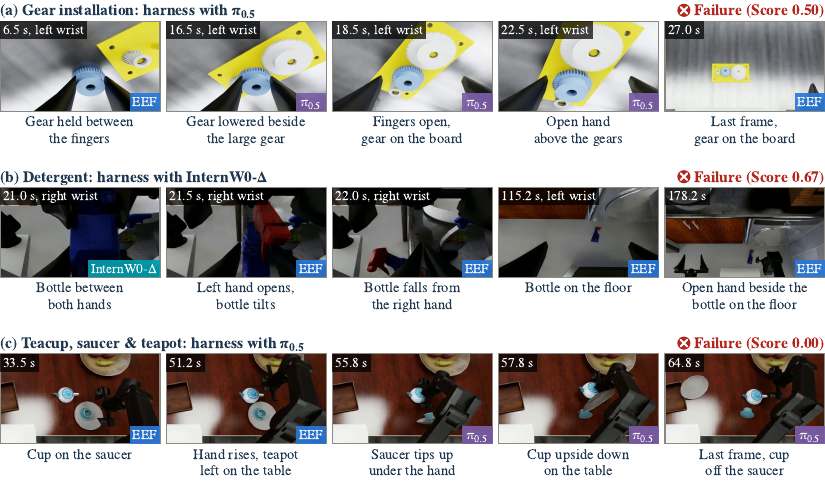}
  \caption{Three unsuccessful episodes in the harness with policy on scenes that the agent solved under direct control: (a) Gear installation with $\pi_{0.5}$, (b) Detergent with InternW0-$\Delta$, and (c) Teacup, saucer \& teapot with $\pi_{0.5}$. Each frame shows the video time, the camera when it is a wrist camera, and the executor, with EEF marking the agent's own commands.}
  \label{fig:case_studies}
\end{figure}

The direct-control episodes of Appendix~\ref{app:samples} illustrate three failure categories: an unconfirmed completion claim, an object lost beyond the reachable workspace, and a long-horizon task that ends with partial credit. Figure~\ref{fig:case_studies} presents three episodes in the harness with policy on scenes that the agent solved under direct control, covering an unconfirmed completion claim, an unrecovered object loss, and a completed stage undone by a subsequent policy segment. The call-by-call records report the executor, the physics steps, the finger opening as defined in Appendix~\ref{app:events}, and the agent's stated reason.

\begin{figure}[htbp]
  \centering
  \includegraphics[width=\linewidth]{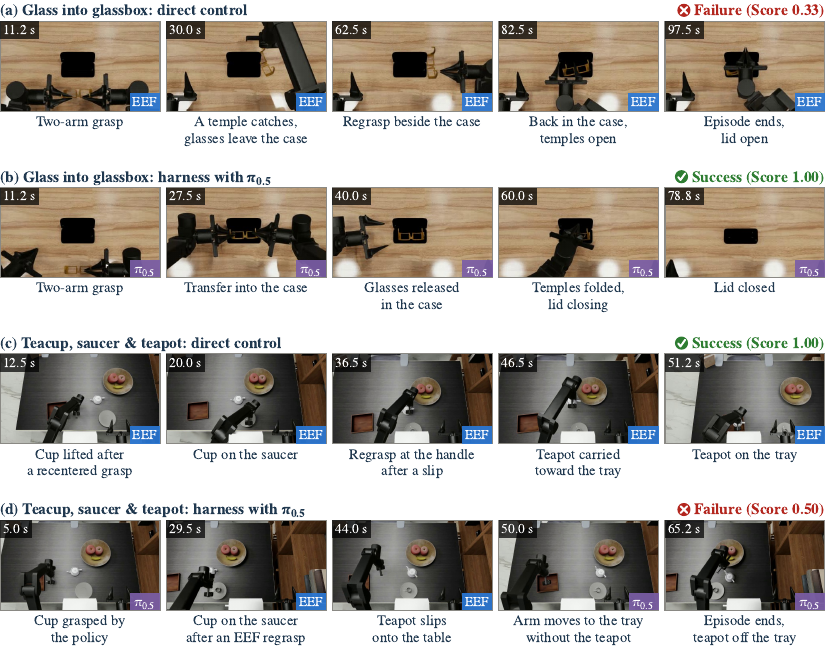}
  \caption{Keyframes of the \astra-based agent under direct control and in the harness with $\pi_{0.5}$ as a tool, with the video time, the executor of each frame, and the outcome of each episode. (a, b) Glass into glassbox on one test scene, under direct control and with $\pi_{0.5}$. (c, d) Teacup, saucer \& teapot on one test scene, under direct control and with $\pi_{0.5}$.}
  \label{fig:tool_rollouts}
\end{figure}

Figure~\ref{fig:tool_rollouts} compares, on one scene each of Glass into glassbox and Teacup, saucer \& teapot, the agent under direct control with the harness with $\pi_{0.5}$ (Sections~\ref{sec:tool:groups} and~\ref{sec:tool:diagnosis}).

\subsection{Division of labor}
\label{app:case_studies:division}
The stated reasons below come from the successful Flip cup \& collect cookies episode discussed in Section~\ref{sec:tool:cases}.

\begin{promptcase}{Division of labor: Flip cup \& collect cookies, harness with InternW0-$\Delta$, score 1.00}
\textbf{Agent, call 3 (EEF).} ``Test cup grasp by raising 1 cm without rotation.''\\
\textbf{Agent, call 4 (InternW0-$\Delta$).} ``Cup rose with the left gripper; continue controlled reorientation and handover.''\\
\textbf{Agent, call 12 (InternW0-$\Delta$).} ``Plate rises with the right hand; transport toward receiving bowl and begin controlled tilt.''\\
\textbf{Outcome.} InternW0-$\Delta$ performs the handover, the reorientation of the cup, and the pour, and the episode succeeds after 14 calls and 2{,}301 physics steps. On the same scene, the agent under direct control reaches a score of 0.60 after 100 calls.
\end{promptcase}

\subsection{Verifying physical completion}
\label{app:case_studies:completion}

Like the direct-control Dishwasher episode of Figure~\ref{fig:self_ended}a, the Gear installation episode of Figure~\ref{fig:case_studies}a with $\pi_{0.5}$ ends with an unconfirmed completion claim. The agent closes the left hand on the gear hub and lifts the gear by 1\,cm (calls 3 and 4, finger opening 29.9 and 30.0\,mm), then invokes $\pi_{0.5}$ for transport and seating in three segments of 180, 180, and 120 physics steps. In the second segment (call 6, stated reason ``Continue fine alignment and seating of the blue gear on the middle axis''), $\pi_{0.5}$ opens the left fingers from 29.9 to 86.2\,mm beside the large gear. The agent then withdraws the open left hand and closes it (calls 8 and 9), requesting only the overview image for both commands, in which the board occupies a small part of the frame. It ends the episode at 828 physics steps with the final message ``The blue gear is in the middle position between the white gears and remained there after the hand withdrew'', and the episode scores 0.5. On the same scene, the agent under direct control releases the gear, inspects the assembly from a more vertical wrist view (call 12), re-approaches the hub to correct a forward offset from the spindle (call 13), and withdraws only after the spindle becomes visible through the gear bore (call 16), solving the scene in 821 physics steps.

\subsection{Unrecovered object losses}
\label{app:case_studies:irreversible}

In the recorded episodes, an object that leaves the reachable workspace is not recovered for the rest of the episode, as when the apple of Figure~\ref{fig:safety}b falls and the agent under direct control searches outside the workspace. In the Detergent episode of Figure~\ref{fig:case_studies}b, the agent drops the first bottle to the floor when it resumes control during a bimanual handover. InternW0-$\Delta$ closes the right hand on the first bottle while the left hand still holds it (call 4, 250 physics steps), ending with a fully closed finger command and a measured finger opening of 32.9\,mm. The agent then opens the left hand (call 5) without commanding the right hand to stay closed, so the right fingers remain at their measured opening of 32.9\,mm (Section~\ref{sec:tool:diagnosis}), and the bottle tilts in the right hand while the left hand opens. In the 1\,cm lift test that follows (call 6, stated reason ``Check whether the receiving hand retained the bottle using a small lift''), the right fingers close from 32.8 to 2.2\,mm within the first 8 physics steps, and the bottle falls from the counter to the floor. InternW0-$\Delta$ places the two other bottles in the basket by physics step 2{,}348, but the next 20 calls (calls 19 to 38) and 2{,}020 physics steps fail to recover the bottle on the floor: seven EEF waypoints directed toward the floor end 11 to 20\,cm from their targets, and three InternW0-$\Delta$ segments of 300 physics steps each end with the fingers closed at 0.0 to 0.1\,mm. The agent ends the episode at 4{,}368 physics steps with the final message ``The blue/red bottle fell to the floor, and recovery attempts failed'', and the episode scores 0.67. Under direct control, the agent solves the same scene in 1{,}681 physics steps, carrying each bottle in the left hand without a transfer between the hands.

\subsection{Preserving completed stages}
\label{app:case_studies:progress}

Figure~\ref{fig:incomplete}b shows a Bottle episode under direct control that ends with partial credit. In the harness with policy, a policy segment can also undo a stage that is already complete, because the post-trained policy receives the fixed task instruction without a record of the completed stages, as analyzed in Section~\ref{sec:tool:summary}. In the Teacup, saucer \& teapot episode of Figure~\ref{fig:case_studies}c, five segments of $\pi_{0.5}$ at the cup (calls 2, 4, and 5) and at the teapot (calls 22 and 24) end with the right hand empty and take 540 of the first 1{,}602 physics steps. Between those segments, the agent completes the cup placement itself, grasping the cup, releasing it on the saucer at physics step 994 (call 20), and verifying the placement (call 21). Its own handle grasp of the teapot holds at a 9.6\,mm finger opening in the lift test (call 31), but the fingers close to 0.0\,mm as the teapot is lifted clear (call 32). The agent invokes $\pi_{0.5}$ for recovery with the stated reason ``reapproach shifted handle with learned alignment'' (call 34, 180 physics steps). In this segment, the hand descends beside the teapot, the saucer tips up under it, and the cup falls and comes to rest upside down on the table, undoing the earlier placement. The agent then tests the new contact as a teapot grasp (call 36) and releases it (call 37), and the episode runs until the step limit of 2{,}000 physics steps inside a final segment of $\pi_{0.5}$ with the stated reason ``Recover displaced cup and saucer'' (call 38). The episode scores 0. Under direct control, the agent solves the same scene in 1{,}631 physics steps.

\subsection{Recovery after delegated handovers}
\label{app:case_studies:shop}

\begin{figure}[htbp]
  \centering
  \includegraphics[width=\linewidth]{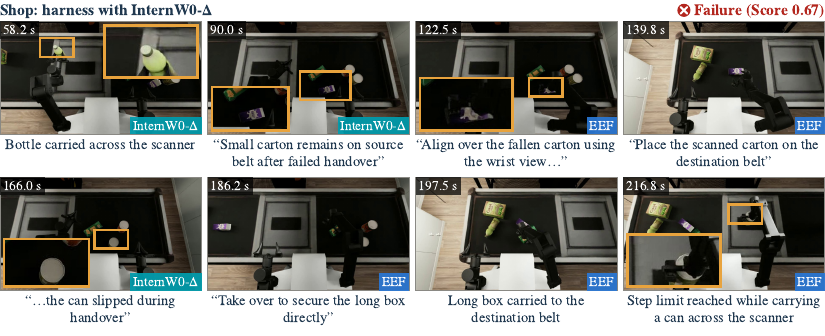}
  \caption{A Shop episode in the harness with InternW0-$\Delta$, with the video time, the executor of each frame, and the agent's stated reasons in quotation marks. Orange boxes are enlarged in the insets.}
  \label{fig:shop_case}
\end{figure}

Figure~\ref{fig:shop_case} shows a Shop episode in the harness with InternW0-$\Delta$ that fails although InternW0-$\Delta$ alone completes the same scene in 4{,}166 physics steps. The failure begins when the small carton drops in a handover between the hands that the agent delegates to the policy. The agent first lets the policy retry and then takes over, placing the carton on the destination belt with its own waypoints. After a can also slips in a later delegated handover, the agent again takes over, this time to secure the long box. Both recoveries succeed, but together with the failed handovers they use up the budget, so the episode reaches the 5{,}000-step limit while the agent carries a can across the scanner.

\FloatBarrier

\end{document}